\documentclass{article}

\usepackage[main, final]{neurips_2026}

\usepackage[utf8]{inputenc} %
\usepackage[T1]{fontenc}    %
\usepackage{hyperref}       %
\usepackage{url}            %
\usepackage{booktabs}       %
\usepackage{amsfonts}       %
\usepackage{nicefrac}       %
\usepackage{microtype}      %
\usepackage{xcolor}         %

\usepackage{graphicx}
\usepackage{subcaption}

\usepackage{amsmath}
\usepackage{amssymb}
\usepackage{mathtools}
\usepackage{amsthm}

\def\E{{\mathbb{E}}}
\def\P{{\mathbb{P}}}
\def\R{{\mathbb{R}}}
\def\qed{{$\hfill\blacksquare$}}

\DeclareMathOperator{\Tr}{Tr}
\DeclareMathOperator{\Cov}{Cov}
\DeclareMathOperator{\Var}{Var}

\usepackage{enumitem} %

\usepackage{csquotes}
\MakeOuterQuote{"}

\theoremstyle{plain}
\newtheorem{theorem}{Theorem}[section]
\newtheorem{proposition}[theorem]{Proposition}
\newtheorem{lemma}[theorem]{Lemma}

\theoremstyle{definition}

\theoremstyle{remark}
\newtheorem{remark}[theorem]{Remark}

\title{Decomposing SGD Dynamics in Neural Networks: Teacher-Induced Spikes and Variance Inflation}

\author{Samet Demir\textsuperscript{1}\thanks{Current affiliation: Oden Institute for Computational Engineering and Sciences, The University of Texas at Austin.}\;, \; Zafer Doğan\textsuperscript{1,2}\thanks{Corresponding author}\\
\textsuperscript{1}MLIP Research Group, KUIS AI Center, Koç University \; \textsuperscript{2}Department of EEE, Koç University \\
\texttt{\{sdemir20,zdogan\}@ku.edu.tr}
}

\begin{document}

\setlength{\abovedisplayskip}{4pt plus 1pt minus 2pt}
\setlength{\belowdisplayskip}{4pt plus 1pt minus 2pt}
\setlength{\abovedisplayshortskip}{0pt plus 1pt}
\setlength{\belowdisplayshortskip}{2pt plus 1pt minus 1pt}

\maketitle

\begin{abstract}%
  We study two-layer neural networks trained by stochastic gradient descent (SGD) on a multi-index teacher model, focusing on how first-layer parameters orthogonal to the teacher subspace affect generalization. We decompose the first-layer weights into a signal component aligned with the teacher and an orthogonal complement, and analyze their joint SGD dynamics together with the second layer in the proportional limit. We show that although SGD updates to the complement are initially isotropic, the training dynamics induce teacher-dependent spectral spikes in the complement Gram matrix. This emergent anisotropy provides a concrete mechanism by which representation variance is amplified, and generalization degrades. To isolate this effect, we introduce Decomposed Dynamics, an analytically tractable surrogate whose asymptotic test-error dynamics coincide with those of SGD. In simulations with a ReLU student, freezing the directions of the orthogonal complement improves generalization relative to training them. Finally, we show that the signal component rapidly aligns with the second-layer weights, revealing strong cross-layer coupling even in this shallow setting. Together, our results provide a dynamical explanation for how feature learning outside the teacher subspace harms generalization.
\end{abstract}

\section{Introduction}

Understanding what stochastic gradient descent (SGD) \emph{actually does} in neural networks—beyond reducing training loss—remains a central open problem in learning theory. Even for two-layer networks, which are simple enough to admit sharp asymptotic analysis, our theoretical understanding is fragmented across regimes and modeling assumptions. In the ``lazy'' or kernel regime, features remain nearly fixed during training, leading to kernel regression with random features or the neural tangent kernel (NTK) \cite{RahimiRecht2007,JacotHG2018}. At the opposite extreme, mean-field and particle-based approaches describe feature learning at infinite width via PDE or interacting particle dynamics \cite{MeiMontanariNguyen2018,ChizatBach2018OT,RotskoffVandenEijnden2018,SirignanoSpiliopoulos2020}. While both viewpoints have delivered major insights, they leave unresolved a basic structural question that arises repeatedly in teacher--student analyses of representation learning:

\emph{When the target depends on a low-dimensional teacher subspace, what role do the many first-layer parameters outside that subspace play in generalization?}

Most existing teacher--student analyses focus on the alignment of learned features with teacher directions \cite{GoldtEtAl2020TeacherStudentSGD,DandiEtAl2024GiantStep,CuiEtAl2024OneStep}. However, in high-dimensional overparameterized settings, the majority of first-layer degrees of freedom lie in the teacher-orthogonal complement. A common simplifying assumption—often justified by concentration or Gaussian equivalence arguments—is that this complement can be summarized by a few low-dimensional quantities or treated as isotropic noise. Yet SGD is not passive: through coupling with the data and the second layer, it can reshape covariance structure, induce anisotropy, and affect test-time behavior in ways invisible if one tracks only teacher-aligned components.

In this work, we make the teacher-orthogonal complement explicit and study its dynamics directly. We decompose the first-layer weights into (i) a \emph{signal matrix} supported on the teacher subspace and (ii) an orthogonal \emph{complement matrix}. We analyze their joint SGD training with the second layer in the high-dimensional proportional limit. A central finding is that the Gram matrix of the complement does \emph{not} remain bulk-like: SGD induces \emph{spectral spikes}, revealing emergent anisotropic structure in the complement covariance. While reminiscent of finite-rank deformations and spiked random matrix models \cite{Johnstone2001,BaikBenArousPeche2005,BenaychGeorgesNadakuditi2009}, here the spikes are \emph{generated endogenously by training dynamics} rather than imposed by the data.

This characterization enables a tractable surrogate description of SGD. Combining our analysis of the complement Gram evolution with the unchanged SGD updates of the second layer and the signal matrix, we introduce \textbf{Decomposed Dynamics}, a simplified algorithm whose asymptotic test-error dynamics coincide with those of SGD. This surrogate isolates the effect of complement anisotropy. The decomposition also lets us address whether training the complement is beneficial for generalization. Counterintuitively, our experiments show that \emph{freezing the complement directions} can outperform training them, consistent with avoiding variance inflation induced by complement anisotropy. We also identify a rapid alignment phenomenon in which the signal matrix becomes strongly coupled to the second-layer weights, underscoring that feature learning and last-layer adaptation are tightly intertwined even in shallow networks.

Beyond its theoretical implications, this mechanism suggests an explanation for why constraining or partially freezing feature representations can improve test performance in overparameterized models. Finally, our perspective connects naturally to stochastic-process approximations of SGD. Classical diffusion and stationary-distribution viewpoints explain aspects of implicit regularization \cite{MandtHoffmanBlei2017,LiTaiE2017SME}, while broader generalization phenomena—such as interpolation and double descent—highlight the importance of variance control \cite{HastieEtAl2019Ridgeless,BelkinEtAl2018DoubleDescent}. Our results sharpen this message for two-layer feature learning: the teacher-orthogonal complement provides a concrete mechanism by which SGD inflates variance through emergent anisotropy, and explicitly controlling—or freezing—this component can improve generalization.

\noindent{Our contributions are summarized as follows:}
\begin{enumerate}[topsep=0pt,itemsep=-1ex,partopsep=1ex,parsep=1ex, leftmargin=*]
    \item Using the standard decomposition of the first-layer weights into a \emph{signal matrix} and a teacher-orthogonal \emph{complement matrix}, we analyze their \emph{joint} SGD dynamics with the second layer in the high-dimensional proportional limit at the level of the complement Gram matrix.
    \item We show that SGD injects coherent teacher-induced structure into the complement, which produces \emph{spectral spikes} in its Gram matrix (Figure~\ref{fig:geometry_and_gram}), revealing emergent anisotropic structure in teacher-orthogonal covariance and identifying a concrete mechanism for variance inflation.
    \item Leveraging this characterization, we propose \textbf{Decomposed Dynamics}, a surrogate whose asymptotic generalization-error dynamics match those of SGD, and show in simulations that \emph{fixing the complement directions} can outperform training them, while also uncovering fast alignment between the signal component and second-layer weights.
\end{enumerate}

\paragraph{Related work ---} Prior scaling-limit analyses of SGD include the teacher-orthogonal complement, but typically through reduced state variables: the overlap matrices of a fixed- or growing-width student in the order-parameter ODEs of online learning \cite{SaadSolla1995,GoldtEtAl2020TeacherStudentSGD,VeigaStephanLoureiroKrzakalaZdeborova2022}, summary statistics such as overlaps, norms and the risk in high-dimensional limit theorems \cite{BenArousGheissariJagannath2024,ArnaboldiStephanKrzakalaLoureiro2023,CollinsWoodfinEtAl2024}, two-time correlation and response functions in DMFT \cite{MignaccoKrzakalaUrbaniZdeborova2020,Gerbelot2024DMFT,BordelonPehlevan2022DMFT}, and the distribution of neurons in infinite-width mean-field limits \cite{MeiMontanariNguyen2018,ChizatBach2018OT,RotskoffVandenEijnden2018,SirignanoSpiliopoulos2020}. These descriptions characterize alignment, risk and marginal dynamics, but do not resolve the teacher-induced structure of the covariance that SGD injects into the Gram matrix of the complement at proportional width, which is the state we track. Spikes in trained first-layer weights are known from one or a few large gradient steps, where they typically lie in teacher-relevant directions and help learning \cite{ba2022highdimensional,moniri2023theory,CuiEtAl2024OneStep,DandiEtAl2024GiantStep}. Here, online SGD injects coherent structure into the complement itself, which carries no target signal and receives isotropic inputs, and Section~\ref{sec:variance-inflation} relates this structure to the test error. Appendix~\ref{app:related-work} gives a detailed discussion of the related work.

\vspace{-0.5em}
\paragraph{Notation ---} We write $\E_t[\cdot] := \E[\cdot \mid \mathcal F_t]$, where $\mathcal F_t$ denotes the $\sigma$-field generated by all randomness up to time~$t$, including the initialization and samples $\{x_s\}_{s<t}$. In particular, $(F_t, a_t)$ is $\mathcal F_t$-measurable, so $\E_t$ averages over the new sample $(x_t,y_t)$ only. For a matrix $A$, we use $\|A\|_2$, $\|A\|_F$, and $\Tr(A)$ to denote its operator norm, Frobenius norm, and trace, respectively. For an integer $m$, let $[m]:=\{1,\dots,m\}$. We use standard big-$O$ and little-$o$ notation; write $A\lesssim B$ if $A\le CB$ for some constant $C$ independent of the diverging dimensions; and write $\asymp$ to denote same order in the diverging dimensions.
\vspace{-0.5em}

\section{Problem setting}
\label{sec:setting}

We study the dynamics of stochastic gradient descent (SGD) in two-layer neural networks trained on data generated by a multi-index teacher. We describe the setting in detail below.

\paragraph{Teacher model (multi-index) ---}
Let $x \sim \mathcal N(0, I_n)$ and consider a \emph{multi-index} teacher of the form
\begin{equation}
    y \;=\; g\!\big(\Xi^\top x\big), 
    \; \textit{ with } \; \Xi \in \mathbb R^{n\times k_\star},\ \ \Xi^\top \Xi = I_{k_\star},
    \label{eq:teacher-multi-index}
\end{equation}
where $k_\star=O(1)$ is fixed and $g:\mathbb R^{k_\star}\to \mathbb R$ is bounded and Lipschitz.
Let $z_\star := \Xi^\top x \sim \mathcal N(0, I_{k_\star})$.

\paragraph{Student (two-layer neural network) ---}
We consider the two-layer neural network model
\begin{align}
    \hat y(x;\,F,a)
    &\;=\; \frac{1}{k}\sum_{i=1}^k a_i\,\sigma\!\big(f_i^\top x\big),
    \label{eq:student}
\end{align}
where $F=[f_1, \dots, f_k]^\top\in\mathbb R^{k\times n}$ and $a\in\mathbb R^k$ are the first and second layer weights. These parameters are trained on the squared loss $\ell(\hat{y}, y):=\frac12(\hat{y}(x;\,F, a)-y)^2+\frac{\lambda}{2k}\|a\|_2^2$ with regularization constant $\lambda$, and we evaluate performance using the generalization error (true risk) $\mathcal L_t:=\E_{x,y}[\frac12(\hat{y}(x;\,F_t, a_t)-y)^2]$.

\paragraph{Training} We study the training dynamics of stochastic gradient descent (SGD) that uses a single sample $(x_t, y_t)$ to update the parameters $(F_t,a_t)$ at each step $t$ (with step sizes $\eta_f, \eta_a > 0$) as follows:
\begin{align}
    f_{i,t+1}
    &=
    f_{i,t} - \eta_f \nabla_{f_i}\ell_t
    \;=\;
    f_{i,t} - \frac{\eta_f}{k}\, e_t\,a_{i,t}\,\sigma'(f_{i,t}^\top x_t)\,x_t,
    \label{eq:sgd-update-f}\\
    a_{t+1}
    &=
    a_t - \eta_a \nabla_a \ell_t
    \;=\;
    a_t - \frac{\eta_a}{k}\Big(e_t\,\sigma(F_t x_t) + \lambda a_t\Big),
    \label{eq:sgd-update-a}
\end{align}
for all $i\in[k]$, where $\ell_t:=\ell(\hat y(x_t;F_t,a_t),y_t)$, $e_t=\hat y(x_t;F_t,a_t)-y_t$ and $\sigma(F_t x_t):=(\sigma(f_{1,t}^\top x_t),\dots,\sigma(f_{k,t}^\top x_t))^\top$. Note that we will drop the time step $t$ from our notation whenever it is not necessary. Each sample is fresh, i.e.,\ independent of $\mathcal F_t$, so our analysis covers online one-pass SGD with batch size one. Extending it to sample reuse is an open direction.

\paragraph{Subspace decomposition ---}
To explicitly identify the degrees of freedom that govern feature learning and generalization in the overparameterized regime, we decompose the first-layer weights into components aligned with the teacher subspace and its orthogonal complement. Prior work \cite{mousavi-hosseini2023neural, MousaviHosseiniWuErdogdu2025MultiIndex, ba2022highdimensional, moniri2023theory, demir2025asymptotic, sonthalia2025low} indicates that feature learning in the first layer is closely related to the teacher directions $\Xi$ in \eqref{eq:teacher-multi-index}. Motivated by this, we study the SGD dynamics of $F_t$ through a decomposition into parts living in $\mathrm{span}(\Xi)$ and its orthogonal complement. Let $P_\parallel := \Xi\Xi^\top$ and $P_\perp := I - P_\parallel$ denote the orthogonal projectors onto the teacher subspace and its orthogonal complement, respectively. Each row of $F_t$ admits the decomposition
\begin{equation}
    f_{i,t} = \Xi m_{i,t} + f_{i,t}^{\perp}, 
    \qquad
    m_{i,t} := \Xi^\top f_{i,t} \in \mathbb R^{k_\star}, 
    \quad
    f_{i,t}^{\perp} := P_\perp f_{i,t} \in \mathbb R^n .
    \label{eq:row-decomp}
\end{equation}
Define $r_{i,t} := \|f_{i,t}^{\perp}\|_2$ and $u_{i,t} := f_{i,t}^{\perp}/\|f_{i,t}^{\perp}\|_2 \in \mathrm{span}(\Xi)^\perp$ (with the convention $u_{i,t}\equiv 0$ when $r_{i,t} = 0$, an event of probability zero throughout under our assumptions). Let $M_t \in \R^{k \times k_\star}$ and $U_t \in \R^{k \times n}$ be the matrices with rows $m_{i,t}^\top$ and $u_{i,t}^\top$, respectively, and define $r_t := (r_{1,t}, \dots, r_{k,t})^\top$. In the remainder of this work, we focus on the joint dynamics of $(a_t, M_t, F_t^\perp)$, where the \emph{complement matrix} is
\begin{equation}
    F^{\perp}_t \;:=\; F_t P_\perp \;=\; 
    \begin{bmatrix} f_{1,t}^\perp, \cdots, f_{k,t}^\perp \end{bmatrix}^\top
    \;=\;\mathrm{diag}(r_t)\,U_t \ \in \R^{k\times n}.
    \label{eq:complement-matrix-def}
\end{equation}

\subsection{Assumptions}

\label{sec:assumptions}
Here, we list the standing assumptions of our theoretical development.

\begin{enumerate}
    \item[\textbf{A1.}] Data samples $(x_t, y_t)$ distributed as $x_t\overset{iid}{\sim}\mathcal N(0,I_n)$ and $y_t = g(\Xi^\top x_t)$.
    \item[\textbf{A2.}] Teacher function $g:\mathbb{R}^{k_\star}\to\mathbb{R}$ is bounded, $k_\star=O(1)$, and $g$ is $L_g$-Lipschitz with respect to $\|\cdot\|_2$ on $\mathbb{R}^{k_\star}$.

    \item[\textbf{A3.}] The activation function $\sigma:\R\to\R$ lies in $W^{3,\infty}(\R)$; that is,
    $\sigma$, $\sigma'$, $\sigma''$, and $\sigma'''$ are bounded.

    \item[\textbf{A4.}] The input dimension $n$ and width of the neural network $k$ jointly diverge with finite ratio, over training horizons proportional to $n$: $n,k \to \infty$ with $k/n\to\alpha\in(0,\infty)$ and $t\le\tau n$ for fixed $\tau \in (0,\infty)$, with step sizes $\eta_a, \eta_f = O(1)$ and regularization $\lambda \ge 0$ fixed constants.
    \item[\textbf{A5.}] The neural network initialization satisfies $f_{i,0}\overset{iid}{\sim} \mathcal N(0, I_n/n)$ and $a_{i,0}$ i.i.d. sub-Gaussian with parameter $K_{\mathrm{sg}}<\infty$ and $\E[a_{i,0}^2]>0$, independent of $F_0$.
\end{enumerate}

\paragraph{Discussion of assumptions ---} Boundedness of $\sigma''$ is needed for the Gaussian-interpolation and Stein's identity steps in the proofs of Lemmas~\ref{lem:complement-geometry}--\ref{lem:gram-onestep} and Theorem~\ref{thm:equivalence-spike} (which also uses bounded $\sigma'''$), the Gaussian--Sobolev bound for the conditional drift in Appendix~\ref{app:teacher-finite-rank}, and the linearization in Proposition~\ref{prop:fast-alignment}. Extensions to non-smooth activations (e.g.,\ ReLU) are not covered by the present analysis; see Appendix~\ref{app:extensions}.

\paragraph{A derived event ---} The following is a \emph{consequence} of \textbf{A1}--\textbf{A5} together with the SGD updates \eqref{eq:sgd-update-f}--\eqref{eq:sgd-update-a}, established as Lemma~\ref{lem:stability-sufficient} in Appendix~\ref{app:stability}; we name it here as a labeled high-probability event so that downstream theoretical results can work on it.

\begin{enumerate}
    \item[\textbf{A6.}]
    \emph{Moment stability of the SGD trajectory.} For each fixed $\tau>0$ there exists $B_\tau<\infty$ such that, with probability $1-o(1)$ as $n\to\infty$,
    $\sup_{t\le \tau n}\frac{1}{k}\sum_{i=1}^k(a_{i,t}^2+\|m_{i,t}\|_2^2+r_{i,t}^2)\le B_\tau$.
\end{enumerate}

\section{Theoretical results}
\label{sec:results}

In this section, we present our main theoretical results. We begin by characterizing the covariance (diffusion) of the row-wise SGD updates of the \emph{complement matrix} $F_t^\perp$. Although the update direction $g_t=P_\perp x_t\sim\mathcal N(0,P_\perp)$ and the initial $F_0^\perp$ are isotropic in the complement, the $k\times k$ covariance of the per-neuron drivers is not. We show that training thereby induces \emph{teacher-dependent structure}, which leads to the emergence of spectral spikes (Figure~\ref{fig:geometry_and_gram}). In particular, Lemmas~\ref{lem:complement-geometry}--\ref{lem:gram-onestep} show that the covariance injected at every step contains a coherent, teacher-conditioned part $S_t$. 

Building on this covariance characterization together with the mean (drift) of the complement updates, we introduce an analytically tractable surrogate algorithm, termed \emph{Decomposed Dynamics}, whose test-error dynamics are asymptotically equivalent to those of SGD under mild conditions. We then quantify the resulting variance inflation caused by teacher-induced anisotropy in the complement. Finally, in an idealized single-index setting, we show that the signal vector $m_t$ rapidly aligns with the second-layer weights $a_t$, highlighting strong cross-layer coupling in the joint training of the two layers.

For clarity of exposition, we state the main objects and results in the single-index case ($k_\star=1$), with the extension to the multi-index setting deferred to Appendix~\ref{app:multi-index}; all results except Proposition~\ref{prop:fast-alignment} extend to any fixed $k_\star$. In the single-index reduction, we write $\Xi=\xi\in\R^n$ and decompose the input as
$x_t = z_{0,t}\xi + g_t$, where $z_{0,t}\sim\mathcal N(0,1)$ and $g_t\sim\mathcal N(0,P_\perp)$ are independent. The teacher output then takes the form $y_t = g(z_{0,t})$.

\subsection{Geometry of updates to the complement matrix}

Define the per-neuron driver $v_{i,t}:=e_t\,a_{i,t}\,\sigma'(f_{i,t}^\top x_t)$, and let $v_t=(v_{i,t})_{i=1}^k\in\R^k$. Writing $g_t:=P_\perp x_t$, the complement update takes the rank-one form
\begin{equation}
    F^{\perp}_{t+1}
    \;=\;
    F^{\perp}_t - \frac{\eta_f}{k}\, v_t g_t^\top ,
    \label{eq:complement-update-main}
\end{equation}
showing that the geometry of the complement updates is governed by the accumulation of $v_t v_t^\top$.
We now characterize this accumulation and show how it decomposes into a centered bulk and teacher-induced structure.

\begin{lemma}[Complement geometry: normalized bulk isotropy and teacher-induced structure]
\label{lem:complement-geometry}
Assume \textbf{A1}--\textbf{A5} and work on the event \textbf{A6}. Fix $\tau>0$ and let $t\le \tau n$.
Decompose $v_s=\mu_s+\tilde v_s$ with $\mu_s:=\E_s[v_s\mid z_{0,s}]$ (an average over $g_s$ only) and $\E_s[\tilde v_s\mid z_{0,s}]=0$, and define the teacher-induced structure $S_s:=\E_s[\mu_s\mu_s^\top]\in\R^{k\times k}$.
Let
\begin{align}
    \widetilde C_t:=\frac1n I_k+\Big(\frac{\eta_f}{k}\Big)^2\sum_{s<t}\tilde v_s\tilde v_s^\top,
\qquad
C_t:=\frac1n I_k+\Big(\frac{\eta_f}{k}\Big)^2\sum_{s<t} v_s v_s^\top . \label{eq:C-accum}
\end{align}
Then, there exists $\varepsilon_n\to0$ such that, with probability $1-o(1)$,
\begin{align}
\sup_{s\le t}\frac{1}{k^2}\|\widetilde C_s-c_s I_k\|_F^2 \le \varepsilon_n,
\qquad
c_s:=\frac1k\Tr(\widetilde C_s), \label{eq:residual-isotropy}\\
C_t=\widetilde C_t+\Big(\frac{\eta_f}{k}\Big)^2\sum_{s<t}S_s+R_t,
\qquad
\sup_{s\le\tau n}\|R_s\|_F\le \varepsilon_n .
\label{eq:C-decomp}
\end{align}
\end{lemma}

\begin{proof}[Proof sketch]
Lemma~\ref{lem:app-decouple-centered} implies negligible normalized off-diagonal drift in $\widetilde C_t$ over $t\le\tau n$.
Writing $\widetilde C_t$ as drift plus martingale, stability (\textbf{A6}) and bounded $\sigma,\sigma'$
yield uniform control and \eqref{eq:residual-isotropy}.
Expanding $v_s=\mu_s+\tilde v_s$ gives \eqref{eq:C-decomp}, with $R_t$ controlled similarly
(Appendix~\ref{app:complement-geometry-proof}).
\end{proof}

Term by term, in $v_sv_s^\top=\tilde v_s\tilde v_s^\top+\mu_s\mu_s^\top+\mu_s\tilde v_s^\top+\tilde v_s\mu_s^\top$ the first term builds $\widetilde C_t$, the conditional mean $S_s$ of the second builds the structured term, and the cross terms together with $\mu_s\mu_s^\top-S_s$ form $R_t$. With fresh samples these are martingale differences, so $R_t$ is a martingale of vanishing Frobenius norm, whereas $S_s\succeq0$ accumulates coherently. ``Teacher-induced'' refers to the conditioning on $z_{0,s}$ in $\mu_s$; the neuron-space directions of $S_s$ are set by the SGD state.
The accumulated structure $(\eta_f/k)^2\sum_{s<t}S_s$ admits a uniform-in-time finite-mode approximation (Appendix~\ref{app:teacher-finite-rank}).

We next analyze how this accumulated structure propagates to the Gram matrix of the complement $F_t^\perp$ and
gives rise to spectral spikes.

\begin{lemma}[One-step update of the complement Gram]
\label{lem:gram-onestep}
Assume \textbf{A1}--\textbf{A5} and work on the event \textbf{A6}. Fix $\tau>0$ and let $t\le \tau n$, with $n_\perp:=\Tr(P_\perp)=n-1$.
Define the normalized complement Gram
\begin{equation}
    G_t:=\frac1{n_\perp}F_t^\perp(F_t^\perp)^\top\in\R^{k\times k},
    \label{eq:complement_gram}
\end{equation}
the complement preactivations $h_t^\perp:=F_t^\perp g_t\in\R^k$ with $g_t:=P_\perp x_t\sim\mathcal N(0,P_\perp)$, the driver covariance $K_t:=\E_t[v_t v_t^\top]=S_t+\E_t[\tilde v_t\tilde v_t^\top]$, and the Stein operator $J_t:=\E_t[\partial v_t/\partial h_t^\perp]\in\R^{k\times k}$, i.e.,\ $(J_t)_{ij}=\frac1k\E_t[a_{i,t}a_{j,t}\sigma'(h_{i,t})\sigma'(h_{j,t})]+\delta_{ij}\,a_{i,t}\E_t[e_t\sigma''(h_{i,t})]$, with $h_{i,t}:=f_{i,t}^\top x_t$.
Then, with probability $1-o(1)$ uniformly over $t\le\tau n$,
\begin{equation}
\E_t[G_{t+1}]
=
G_t-\frac{\eta_f}{k}\big(J_tG_t+G_tJ_t\big)
+\Big(\frac{\eta_f}{k}\Big)^2 K_t
+E_t,
\label{eq:gram-onestep}
\end{equation}
where
\begin{equation}
\sup_{t\le\tau n}\|E_t\|_F=O\big(n^{-3/2}\big).
\label{eq:gram-onestep-rem}
\end{equation}
\end{lemma}

\begin{proof}[Proof sketch]
Expanding $G_{t+1}=\frac1{n_\perp}F_{t+1}^\perp(F_{t+1}^\perp)^\top$ with $F_{t+1}^\perp=F_t^\perp-(\eta_f/k)v_t g_t^\top$ yields $G_t$, linear terms in $\E_t[v_th_t^{\perp\top}]$ and its transpose, and a quadratic term $(\eta_f/k)^2\E_t[(\|g_t\|_2^2/n_\perp)v_tv_t^\top]$. Given $(\mathcal F_t,z_{0,t})$, $h_t^\perp\sim\mathcal N(0,n_\perp G_t)$, so Gaussian integration by parts gives exactly $\E_t[v_th_t^{\perp\top}]=n_\perp J_tG_t$. Gaussian norm concentration gives the quadratic term as $(\eta_f/k)^2K_t+O(n^{-3/2})$; see Appendix~\ref{app:gram-onestep-proof}.
\end{proof}

\begin{figure}[t]
    \centering
    \begin{subfigure}[b]{0.32\textwidth}
         \centering
         \includegraphics[width=0.99\linewidth]{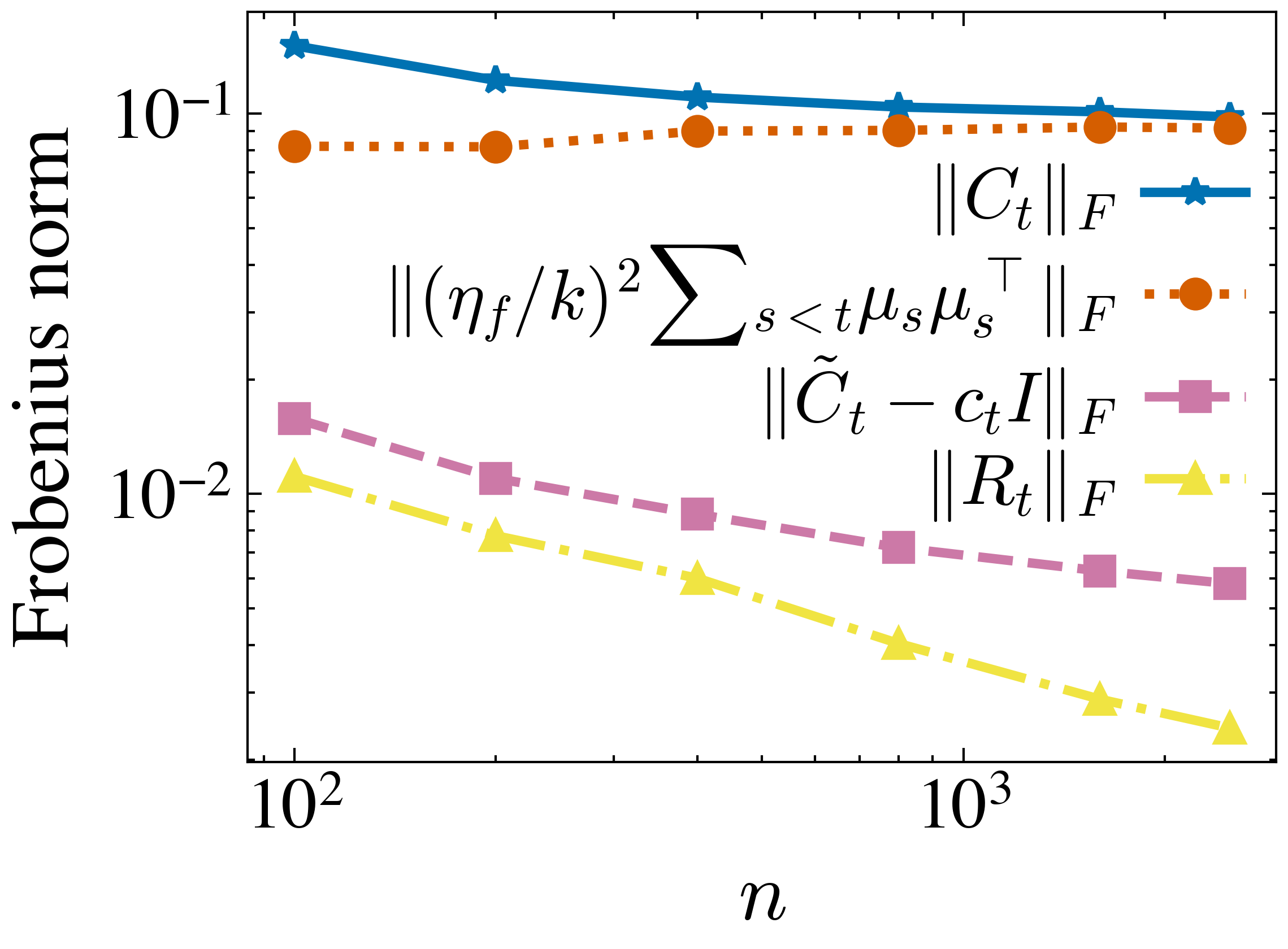}
         \caption{\centering Frobenius norm of $C_t$}
     \end{subfigure}
     \hfill
    \begin{subfigure}[b]{0.32\textwidth}
         \centering
         \includegraphics[width=0.99\linewidth]{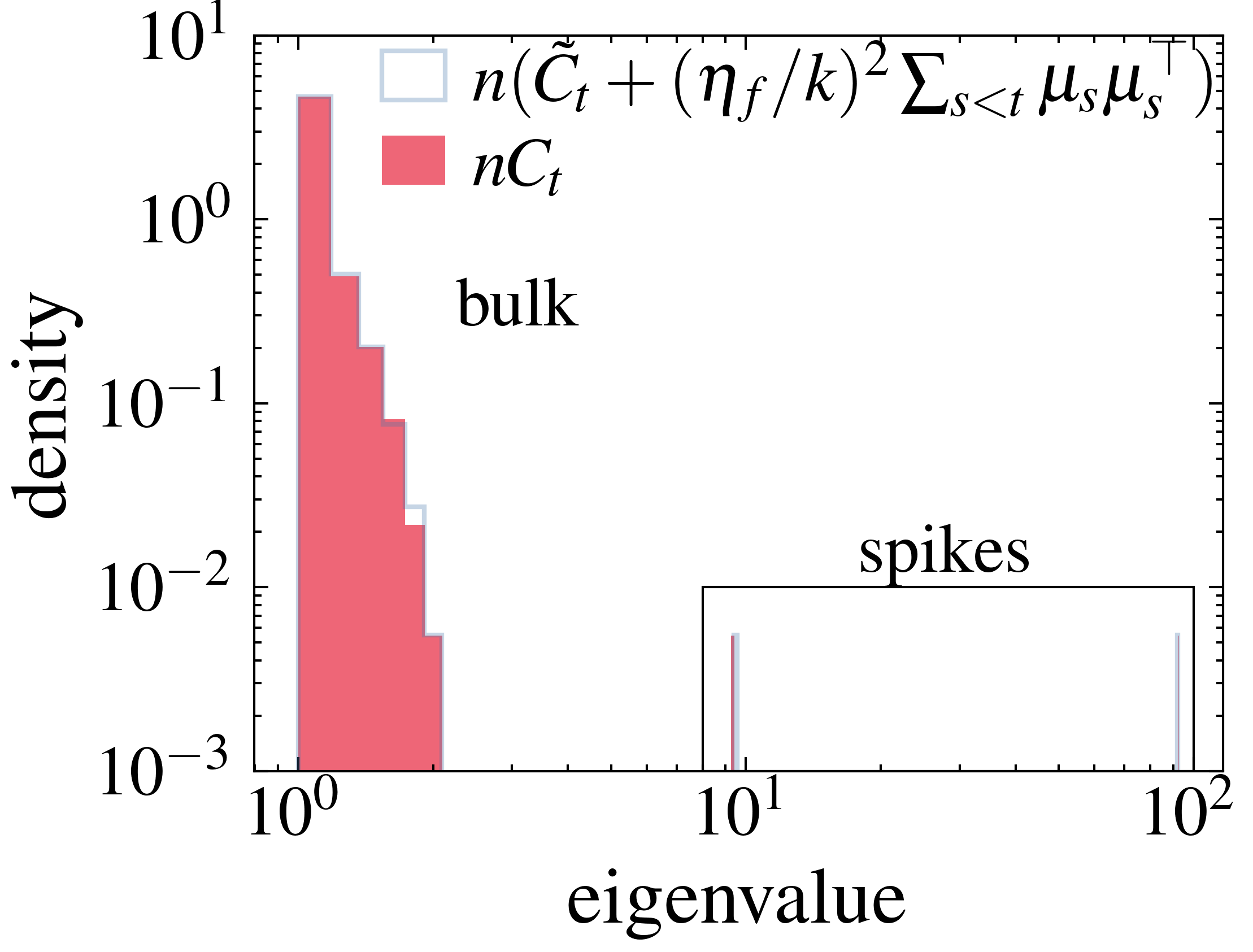}
         \caption{\centering Spectral density of $nC_t$}
     \end{subfigure}
    \hfill
    \begin{subfigure}[b]{0.32\textwidth}
         \centering
         \includegraphics[width=0.99\linewidth]{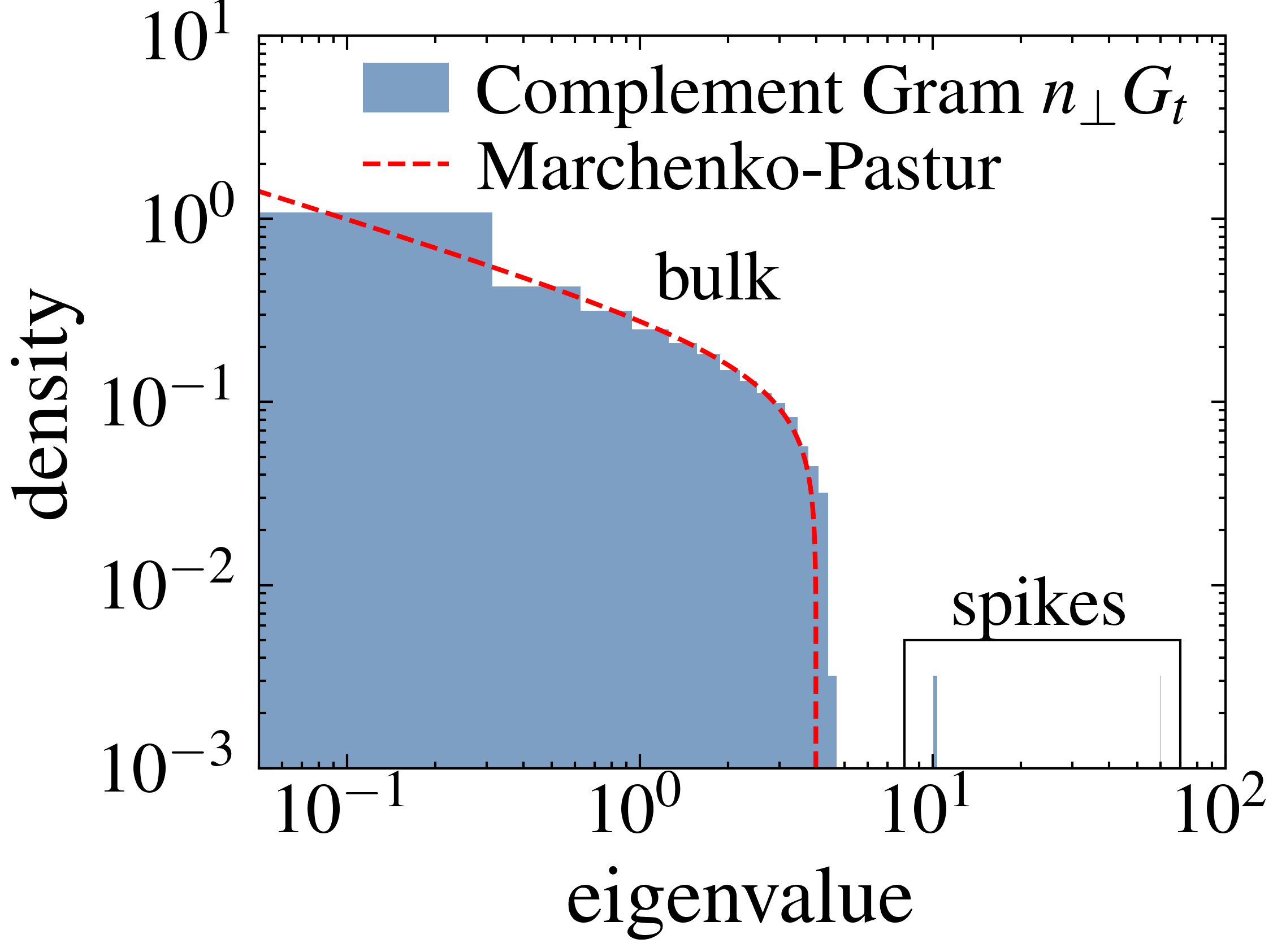}
         \caption{\centering Spectral density of $n_\perp G_t$}
     \end{subfigure}
     \caption{Visualization of Lemma~\ref{lem:complement-geometry} and Lemma~\ref{lem:gram-onestep}. Panels (a)--(b) illustrate how the accumulation of teacher-induced structure, captured by the driver accumulator $C_t$ in \eqref{eq:C-accum}, leads to bulk-plus-spike behavior in its spectral density, in line with Lemma~\ref{lem:complement-geometry}. Panel (c) shows the resulting bulk-plus-spike structure in the spectrum of $n_\perp G_t$, in line with Remark~\ref{rem:bulk-plus-spikes}. Norms in (a) are unnormalized, and panels (a)--(b) use the realized $\mu_s\mu_s^\top$ in place of $S_s$. Parameters are $\alpha = 1.0$, $\tau = 2.0$, $\lambda = 0.01$, $k_\star = 1$, $g(\cdot) = \tanh(\cdot)$, $\sigma=\mathrm{ReLU}$, and $\eta_a = \eta_f = 0.8$ throughout, with input dimension $n=1000$ for panels (b)--(c).}
     \label{fig:geometry_and_gram}
\end{figure}

\begin{remark}[Bulk and teacher-induced spikes in the complement Gram]
\label{rem:bulk-plus-spikes}
Lemmas~\ref{lem:complement-geometry} and~\ref{lem:gram-onestep} separate the mechanisms shaping the spectrum of $G_t$.
First, Lemma~\ref{lem:complement-geometry} shows that, up to the centered accumulator $\widetilde C_t$ and a vanishing remainder, $C_t$ is the accumulated teacher-induced component $(\eta_f/k)^2\sum_{s<t}S_s$, which admits a uniform-in-time finite-mode approximation (Appendix~\ref{app:teacher-finite-rank}).
Second, Lemma~\ref{lem:gram-onestep} yields the recursion $\E_t[G_{t+1}] = G_t-\frac{\eta_f}{k}(J_tG_t+G_tJ_t) + (\eta_f/k)^2 K_t$, up to $O(n^{-3/2})$ error.
The $J_t$ term damps the current Gram through the first (Gauss--Newton) summand of $J_t$, while $(\eta_f/k)^2K_t$ injects a positive semidefinite covariance at each step. The coherent part of $K_t$, the approximately low-rank teacher-induced structure $S_t$ (Appendix~\ref{app:teacher-finite-rank}), creates outliers above a Marchenko--Pastur bulk in the spectrum of $G_t$; the lemmas establish this mechanism, and Figure~\ref{fig:geometry_and_gram}c shows the resulting spectrum.
\end{remark}

Figure~\ref{fig:geometry_and_gram} illustrates this bulk-plus-spikes phenomenon. As the input dimension $n$ increases, $\|\widetilde C_t-c_tI_k\|_F$ and $\|R_t\|_F$ decrease, while the accumulated teacher-induced term $(\eta_f/k)^2\sum_{s<t}S_s$ remains non-negligible and dominates the norm of $C_t$. The resulting spectrum of $n_\perp G_t=F_t^\perp(F_t^\perp)^\top$ exhibits a Marchenko--Pastur bulk together with additional spikes, in agreement with Remark~\ref{rem:bulk-plus-spikes}.

\paragraph{How large are the spikes? ---} A natural question is whether the outliers in Figure~\ref{fig:geometry_and_gram}c are large enough to matter or merely a finite-size effect. In the setting of Figure~\ref{fig:geometry_and_gram}c, over five seeds (mean $\pm$ standard deviation), the top eigenvalue of $n_\perp G_t$ is $\lambda_1=63.3\pm1.6$, far above the Marchenko--Pastur edge $4$ of the initial spectrum. Its per-neuron share $\lambda_1/k\approx0.063$ is about $5.5\%$ of the mean complement variance $\frac1k\sum_i r_{i,t}^2$ but $40\pm3\%$ of the mean signal variance $\frac1k\sum_i m_{i,t}^2$: the spike is small compared with the complement as a whole, yet comparable to the teacher-aligned signal. Halving $n$ roughly halves $\lambda_1$ (ratio $0.52\pm0.02$), so $\lambda_1/k$ stays nearly constant as $n$ grows.

\subsection{Equivalence of SGD and Decomposed Dynamics in test-error dynamics}

Next, we focus on the characterization of the test-error (generalization) dynamics of the SGD by studying the updates of the complement $F_t^\perp$. Specifically, in our setting, if two algorithms have the same drift (the conditional mean of the update) and diffusion (its neuron-space second moment), then, given the stability estimates of Lemma~\ref{lem:S}, they perform equivalently on the test metrics of interest (mainly generalization error), in the spirit of propagation-of-chaos for proportional-regime SGD \cite{sznitman1991topics, celentano2021highdimensional}. In the following theorem, we use this property to replace the original update of the complement matrix $F_t^\perp$ with a simpler surrogate update rule while preserving the test-error dynamics in the asymptotic limit.

\begin{theorem}[Equivalence of SGD and Decomposed Dynamics in learning curves]
\label{thm:equivalence-spike}
Assume \textbf{A1}--\textbf{A5} and $|a_{i,0}|\le A$ almost surely for a constant $A$. Consider two training procedures initialized from the same $(F_0,a_0)$:
\begin{enumerate}
    \item \textbf{(SGD)}: stochastic gradient descent updates \eqref{eq:sgd-update-f}--\eqref{eq:sgd-update-a}.
    \item \textbf{(Decomposed Dynamics)}: a surrogate that (i) updates $(a_t^{\mathrm{DD}},M_t^{\mathrm{DD}})$ by the SGD rule (Appendix~\ref{app:eq-step2}),
    and (ii) evolves its complement matrix $(F^{\perp}_t)^{\mathrm{DD}}$ by the following update, which has the same
    conditional mean as the SGD complement update:
\begin{equation}
    (F^{\perp}_{t+1})^{\mathrm{DD}}
    \;=\;
    (I-\widetilde\Lambda_t)\,(F^{\perp}_t)^{\mathrm{DD}} \;+\; \zeta_t,
    \label{eq:dd-complement-update}
\end{equation}
where $\mathrm{vec}(\zeta_t^\top)\sim\mathcal N(0,\Gamma_t\otimes P_\perp)$ is generated from fresh standard Gaussian innovations,
independent across time and of the fresh sample driving the $(a_t,M_t)$ updates. 

Furthermore,
$\widetilde\Lambda_t$ and $\Gamma_t$ are defined as
\begin{equation}
    \Gamma_t
    \;:=\;
    \Big(\frac{\eta_f}{k}\Big)^2
    \E_t\!\left[v_t v_t^\top \right]\in\R^{k\times k},
    \qquad
    \widetilde\Lambda_t \;:=\; \frac{\eta_f}{k}\,J_t,
    \label{eq:gamma-lambdatotal-def}
\end{equation}
with $J_t$ the Stein operator of Lemma~\ref{lem:gram-onestep}; both are evaluated at the current DD state.
\end{enumerate}
Then, for any fixed $\tau>0$, $\sup_{0\le t\le \tau n}\big|\mathcal L_t^{\mathrm{SGD}}-\mathcal L_t^{\mathrm{DD}}\big|
    \xrightarrow[]{\mathbb P} 0$.
\begin{proof}
Decomposed Dynamics matches (i) the conditional drift and diffusion of $(a_t,M_t)$, (ii) the
complement second moment via $\mathrm{vec}(\zeta_t^\top)\sim\mathcal N(0,\Gamma_t\otimes P_\perp)$ with
\eqref{eq:gamma-lambdatotal-def}, and (iii) the conditional mean of the SGD complement increment, which by Stein's identity is $-\widetilde\Lambda_tF_t^\perp$. Hence the conditional mean increments of the reduced state $(a_t,M_t,\Sigma_t)$ with $\Sigma_t = F_t^\perp(F_t^\perp)^\top$, which determines the test loss, agree up to $O(n^{-3/2})$ per step in the norm $\|\cdot\|_\star$ of Appendix~\ref{app:eq-step4}. By Lemma~\ref{lem:S}, a discrete Gr\"onwall argument transports this to the test loss. The full proof is provided in Appendix~\ref{app:equivalence-proof}.
\end{proof}
\end{theorem}

Figure \ref{fig:equivalence} illustrates our numerical results comparing the SGD with the equivalent surrogate (Decomposed Dynamics) for different learning rates $\eta = \eta_f = \eta_a$, various $\alpha = k/n$ ratios, and changing activation functions $\sigma(\cdot)$. The results affirm that the Decomposed Dynamics closely match the learning curves of the SGD in terms of the generalization (test) error under various conditions, even in moderate dimensions. This surrogate is useful for characterizing the behavior of the complement matrix $F_t^\perp$, as it separates the injected covariance $\Gamma_t$, which contains the teacher-induced part $(\eta_f/k)^2S_t$, from the drift $-\widetilde\Lambda_tF_t^\perp$.

\subsection{Variance inflation by the teacher-induced structure}\label{sec:variance-inflation}

Next, we study the impact of the teacher-induced complement structure (spikes) on the generalization error in the following proposition. To begin, we observe that $h_t^\perp := F_t^\perp g_t \sim \mathcal{N}(0, F_t^\perp (F_t^\perp)^\top)$ for a given $F_t$. Remark~\ref{rem:bulk-plus-spikes} and Figure~\ref{fig:geometry_and_gram} indicate that the complement Gram admits an approximate bulk-plus-structure decomposition: $n_\perp G_t = F_t^\perp (F_t^\perp)^\top \approx \hat{c}_t I_k + B_t^{\mathrm{acc}}(B_t^{\mathrm{acc}})^\top$ where $\hat{c}_t I_k$ captures the isotropic (bulk) part while $B_t^{\mathrm{acc}}(B_t^{\mathrm{acc}})^\top$ captures the accumulated teacher-induced structure (spikes), with $B_t^{\mathrm{acc}}$ the accumulated loading matrix defined in \eqref{eq:app-Bacc-def} of Appendix~\ref{app:teacher-finite-rank}, and $\hat c_t$ the bulk-eigenvalue level of $n_\perp G_t$ (with $\hat c_0 \approx 1$ from the i.i.d. Gaussian initialization). Replacing the covariance of $h_t^\perp$ by this approximation, we define an effective predictor as follows:
\begin{equation}
    \hat y_t^{\mathrm{eff}}
    \;:=\;
    \frac{1}{k}\sum_{i=1}^k a_{i,t}\,
    \sigma\!\Big(m_{i,t}^\top z_\star + \big(\sqrt{\hat c_t}\varepsilon + B_t^{\mathrm{acc}}\,s\big)_i\Big),
    \label{eq:yhat-eff}
\end{equation}
where $z_\star\sim\mathcal N(0,I_{k_\star})$, $\varepsilon\sim\mathcal N(0, I_k)$, $s\sim\mathcal N(0,I)$ are independent random vectors, independent of $\mathcal F_t$ (the dimension of $s$ is the column count of $B_t^{\mathrm{acc}}$).

\begin{proposition}[Variance Inflation by Teacher-Induced Complement Structure]
\label{prop:variance-inflation}
Assume \textbf{A1}--\textbf{A5} and work on the event \textbf{A6}; fix $\tau>0$, $t\le \tau n$.
Let $\mathcal F_t$ be the $\sigma$-field generated by the randomness up to time $t$ so that $(a_t, M_t, \hat c_t, B_t^{\mathrm{acc}})$ are measurable (deterministic). Define the corresponding effective test loss
\begin{equation}
    \mathcal L_t^{\mathrm{eff}}
    \;:=\;
    \frac12\,\E\!\left[\big(\hat y_t^{\mathrm{eff}}-g(z_\star)\big)^2 \,\middle|\, \mathcal F_t\right].
    \label{eq:Leff-def}
\end{equation}
Then $\mathcal L_t^{\mathrm{eff}}$ admits the decomposition
\begin{equation}
    2\mathcal L_t^{\mathrm{eff}}
    \;=\;
    \mathcal L_{\mathrm{bias},t}
    \;+\;
    \mathcal L_{\mathrm{res},t}
    \;+\;
    \mathcal L_{\mathrm{str},t},
    \label{eq:Leff-decomp}
\end{equation}
where
\begin{align}
    \mathcal L_{\mathrm{bias},t}
    &:=\E_{z_\star}\!\left[\Big(\E[\hat y_t^{\mathrm{eff}}\mid z_\star,\mathcal F_t]-g(z_\star)\Big)^2\right], \quad \mathcal L_{\mathrm{res},t} :=\E_{z_\star}\E_{s}\!\left[\Var_{\varepsilon}\!\big(\hat y_t^{\mathrm{eff}}\mid z_\star,s,\mathcal F_t\big)\right], \label{eq:Leff-bias-res}\\
    \mathcal L_{\mathrm{str},t}
    &:=\E_{z_\star}\!\left[\Var_{s}\!\Big(\E_{\varepsilon}[\hat y_t^{\mathrm{eff}}\mid z_\star,s,\mathcal F_t]\Big)\right]
    \;\ge\;0. \label{eq:Leff-str}
\end{align}
Moreover, $\mathcal L_{\mathrm{str},t}=0$ whenever $B_t^{\mathrm{acc}}=0$ (no teacher-induced structure in the complement channel). Let $\omega_{i,t}(z_\star):=a_{i,t}\,\E_\varepsilon[\sigma'(m_{i,t}^\top z_\star+\sqrt{\hat c_t}\,\varepsilon_i)]$ be the slope-weighted readout. Then $\mathcal L_{\mathrm{str},t}>0$ whenever $(B_t^{\mathrm{acc}})^\top\omega_t(z_\star)\neq0$ for some $z_\star\in\R^{k_\star}$.
\begin{proof}
This is an iterated law-of-total-variance identity: first condition on $z_\star$, then condition on $(z_\star,s)$ and integrate over
$\varepsilon$. Full proof is given in Appendix~\ref{app:variance-inflation}.
\end{proof}
\end{proposition}

\begin{figure*}[t]
    \centering
    \begin{subfigure}[b]{0.32\textwidth}
         \centering
         \includegraphics[width=0.99\linewidth]{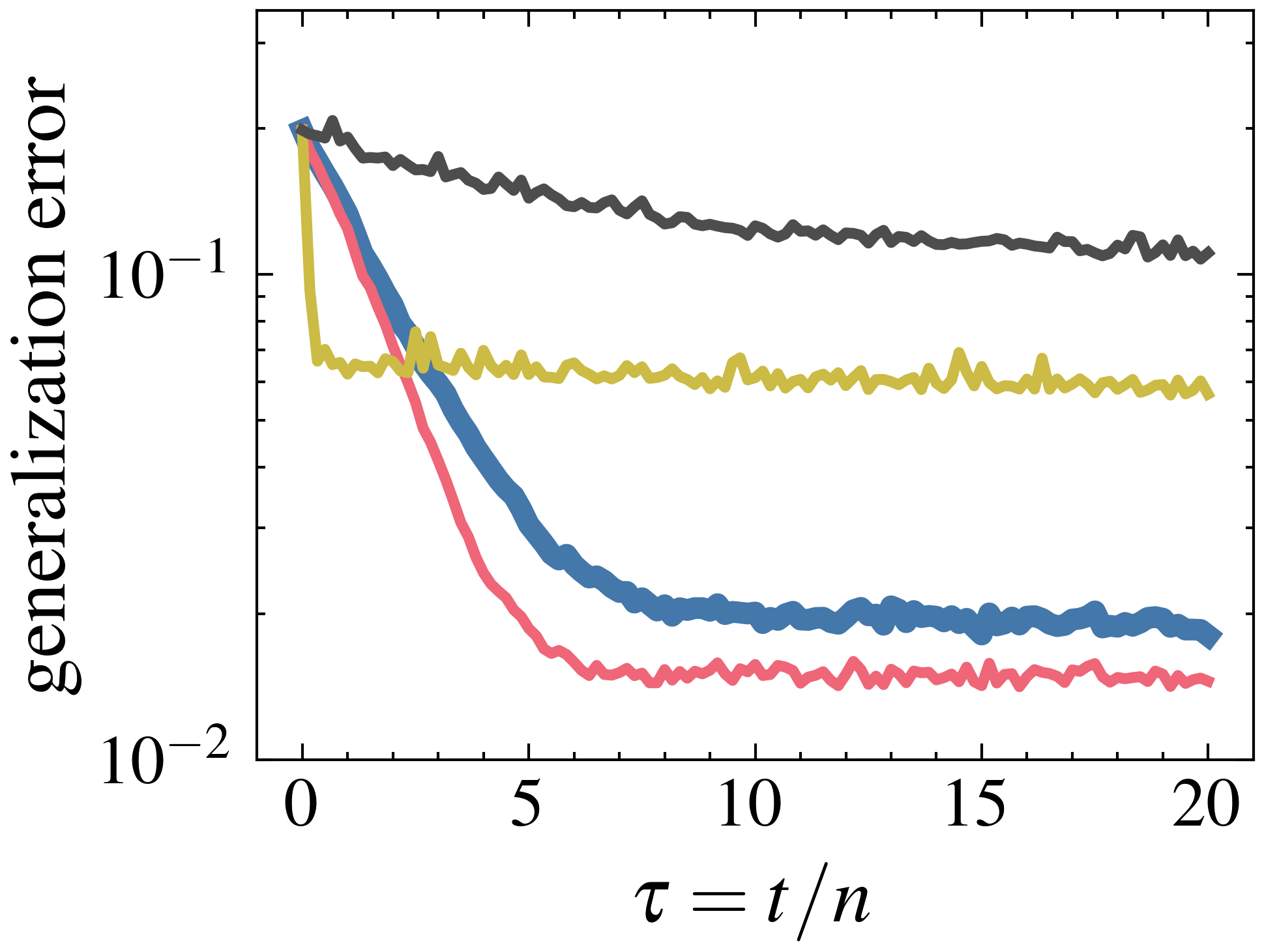}
         \caption{generalization}
     \end{subfigure}
     \hfill
    \begin{subfigure}[b]{0.32\textwidth}
         \centering
         \includegraphics[width=0.99\linewidth]{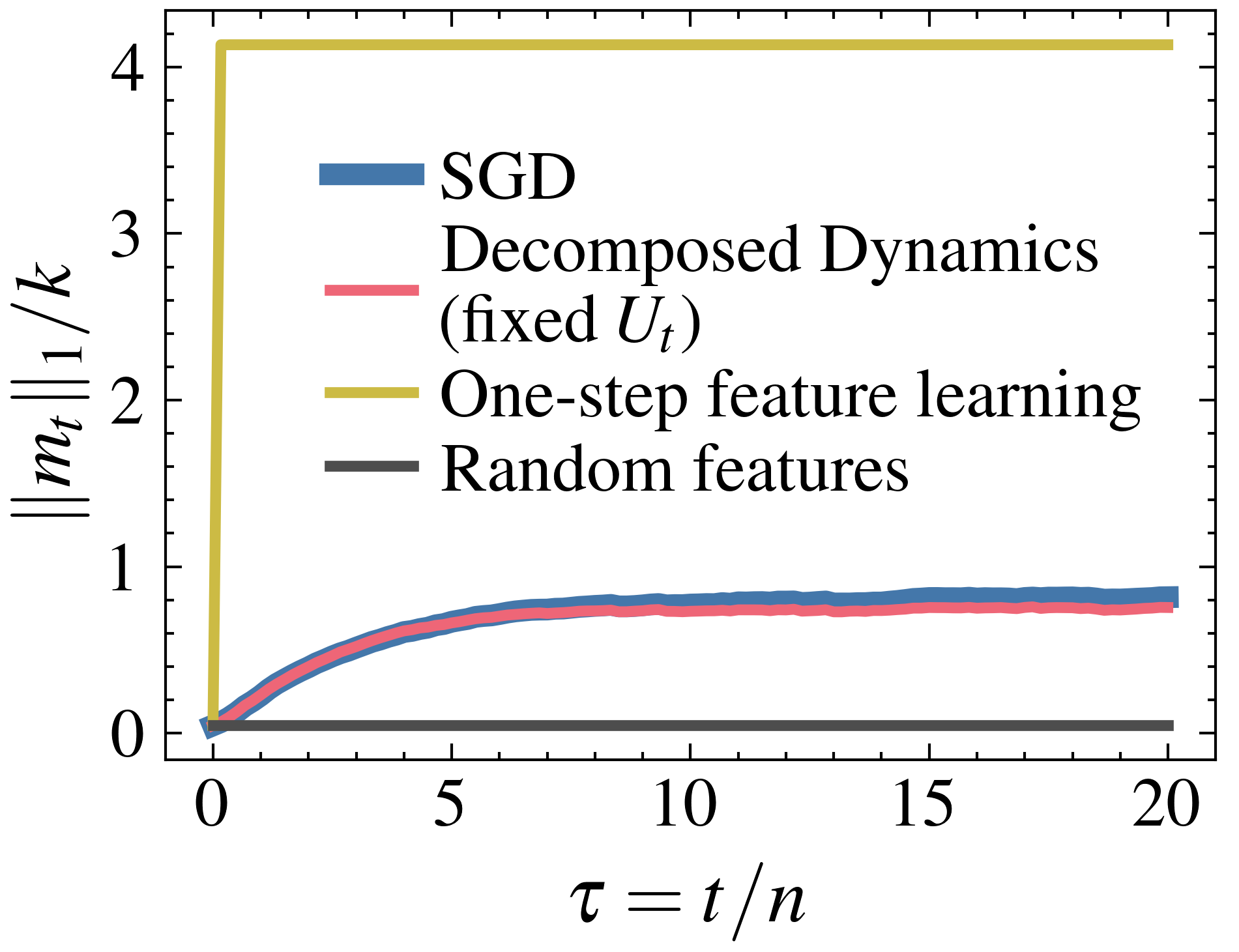}
         \caption{mean $|m|$ (subspace strength)}
     \end{subfigure}
     \hfill
    \begin{subfigure}[b]{0.32\textwidth}
         \centering
         \includegraphics[width=0.99\linewidth]{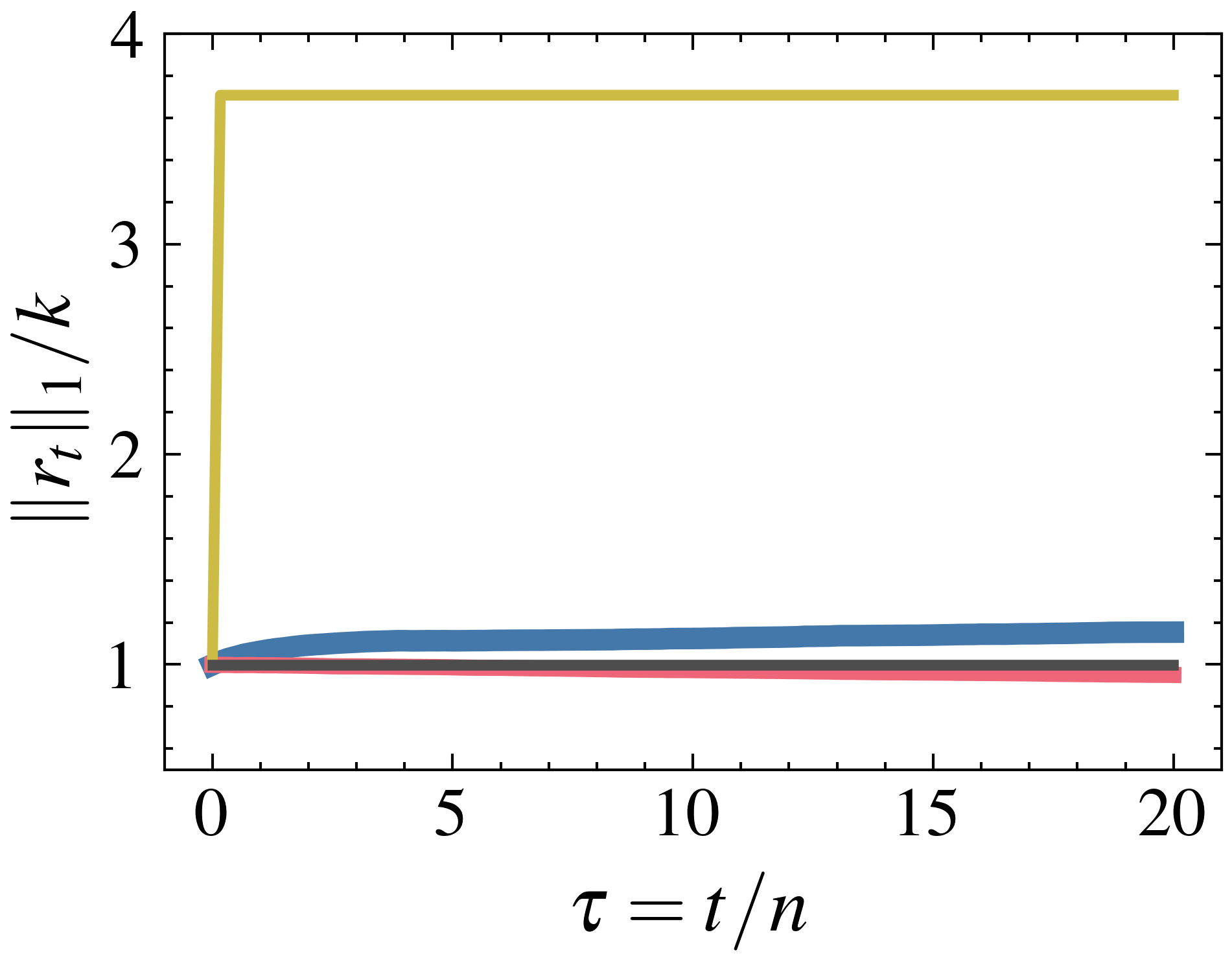}
         \caption{mean $r$ (complement norm)}
     \end{subfigure}

     \caption{Comparison of learning algorithms in the decomposition variables. Panels show (a) test error, (b) mean teacher-subspace strength $\|m_t\|_1/k$, and (c) mean complement norm $\|r_t\|_1/k$ for full SGD, Decomposed Dynamics with fixed complement directions $U_t$, one-step feature learning, and random features. Parameters are $n=300$, $\alpha=1.0$ ($k=300$), $\tau=20.0$, $k_\star=1$, $g(\cdot)=\tanh(\cdot)$, $\sigma(\cdot)=\operatorname{ReLU}(\cdot)$, $\eta_a=\eta_f=1.0$ for the SGD/DD baselines, ridge $\lambda=10^{-4}$, and single-sample training; the one-step baseline uses one initial first-layer update estimated with $1000$ samples, and all methods use the same initialization; random features use $\eta_a=60$ and one-step learning $\eta_a=1.5$ with a first-layer step $\eta_f/k=\sqrt n$.}
     \label{fig:comparison_of_algorithms}
\end{figure*}

Proposition \ref{prop:variance-inflation} isolates the variance of the predictor along the structure (spike) directions, $\mathcal L_{\mathrm{str},t}\ge0$, which we call ``teacher-induced variance inflation''. In Section~\ref{sec:sims} (Figure~\ref{fig:comparison_of_algorithms}a), we illustrate that the Decomposed Dynamics with fixed $U_t$ (fixed complement directions dropping the teacher-induced spikes from the learning dynamics similar to setting $B_t^{\mathrm{acc}} = 0$) achieves better (lower) test-error curve than the SGD, consistent with the prediction that teacher-induced structure can increase the test error by variance inflation.

\subsection{Coupling of the first and second layer weights}

Finally, we investigate the coupling of the first layer $F_t$ and the second layer $a_t$ in an idealized (oracle) version of the dynamics in the following proposition. Focusing on the single-index case $k_\star = 1$, where $F_t$ is decomposed as $F_t = m_t \xi^\top + F_t^\perp$ for some $m_t \in \R^k$, we show that $m_t$ and $a_t$ get aligned quickly during the early steps of the training. Let $\kappa_g:=\E[g(z)z]$ and $\kappa_\sigma:=\E[\sigma'(z)]$, $z\sim\mathcal N(0,1)$, be the first Hermite coefficients of $g$ and $\sigma$, and let $\cos_t:=a_t^\top m_t/(\|a_t\|_2\|m_t\|_2)$.

\begin{proposition}[Fast alignment of $m_t$ and $a_t$]
\label{prop:fast-alignment}
Let $k_\star=1$ and assume \textbf{A1}--\textbf{A5} with $\eta_a,\eta_f>0$ and $|a_{i,0}|\le A$ almost surely for a constant $A$. Suppose that $g$ and $\sigma$ are odd with $\kappa_g\kappa_\sigma\neq0$, and let $(a_t,m_t)$ follow the oracle no-spike dynamics of Appendix~\ref{app:fast-alignment}, in which the complement channel is replaced by synthetic independent standard Gaussians. Then for every $\varepsilon>0$ there is $\tau_0(\varepsilon)>0$ such that, with probability $1-o(1)$ over initialization,
\begin{equation}
    1-\cos_t^2\;\le\;\varepsilon \qquad \text{for all } k^{2/3}\le t\le \tau_0(\varepsilon)\,k.
    \label{eq:fast-alignment-bound-main}
\end{equation}
In particular, $m_t$ aligns with $a_t$ within $o(k)$ steps.

\begin{proof}
See Appendix~\ref{app:fast-alignment}.
\end{proof}
\end{proposition}

\begin{figure*}[t]
    \centering
    \begin{subfigure}[b]{0.32\textwidth}
         \centering
         \includegraphics[width=0.99\linewidth]{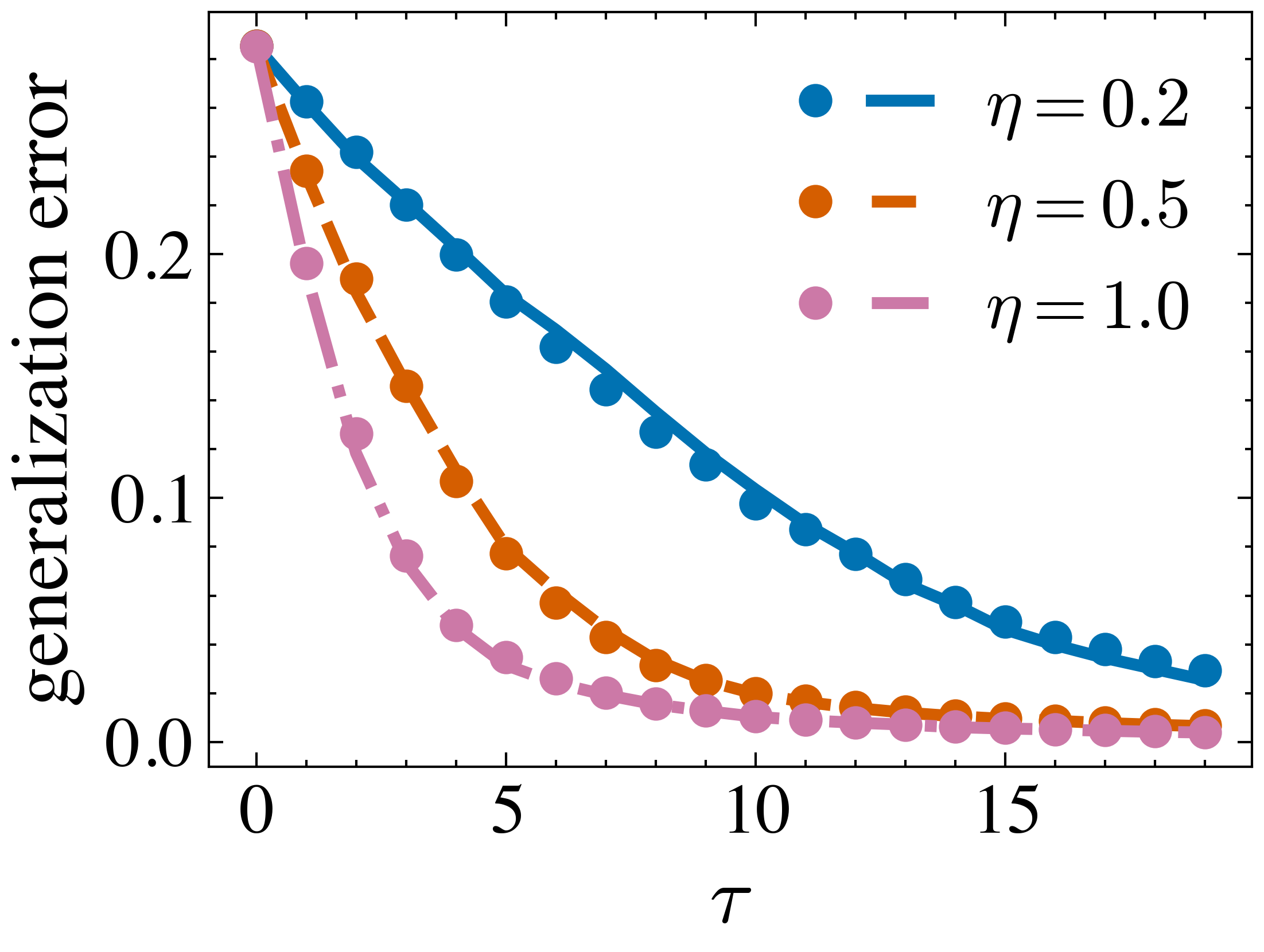}
         \caption{\centering learning rate $\eta_a = \eta_f = \eta$ ($\alpha = 1.0$ and $\sigma=\mathrm{ReLU}$)}
     \end{subfigure}
     \hfill
    \begin{subfigure}[b]{0.32\textwidth}
         \centering
         \includegraphics[width=0.99\linewidth]{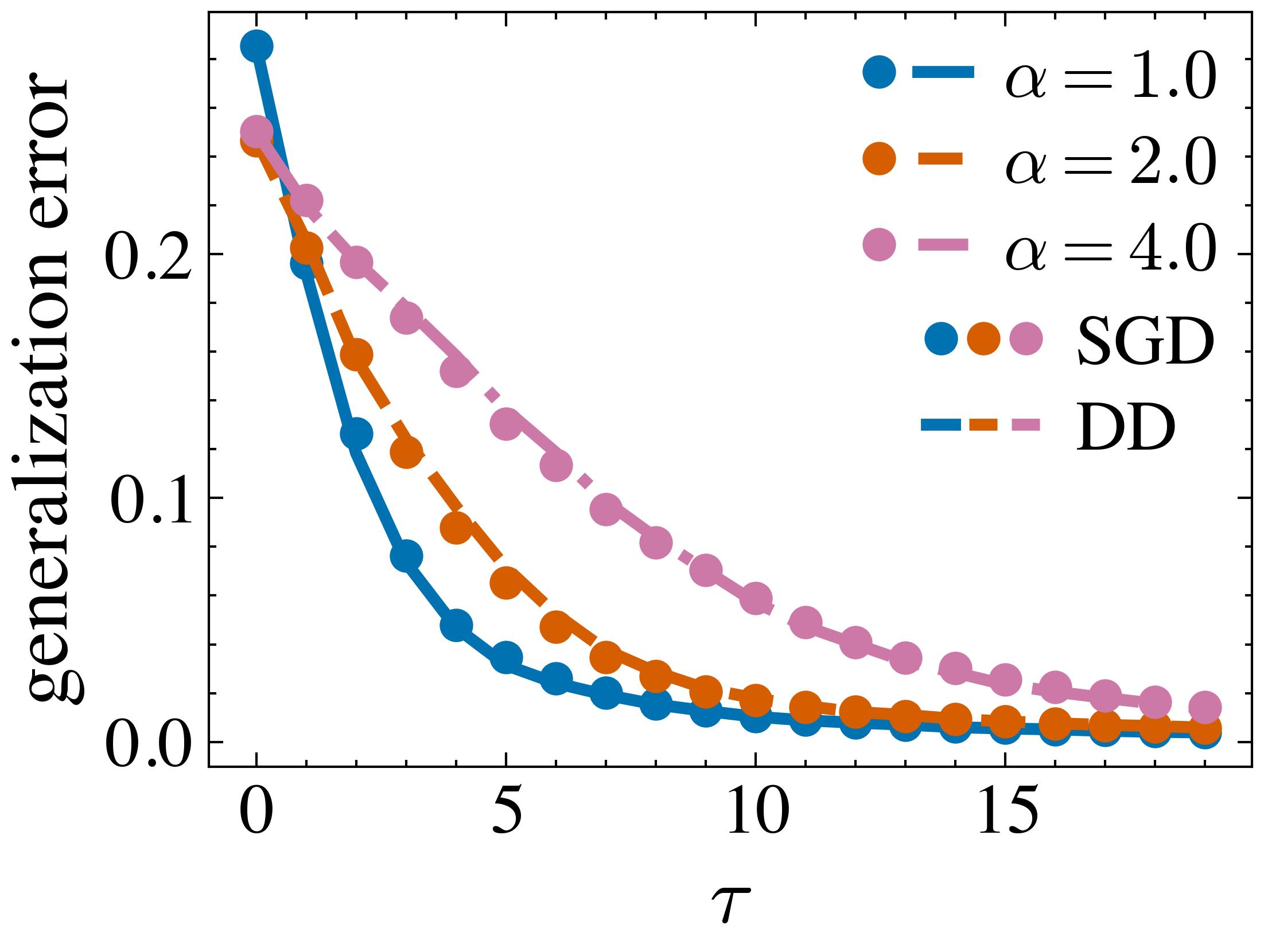}
         \caption{\centering width $\alpha = k/n$ ($\eta_a = \eta_f = 1.0$, $\sigma=\mathrm{ReLU}$)}
     \end{subfigure}
     \hfill
    \begin{subfigure}[b]{0.32\textwidth}
         \centering
         \includegraphics[width=0.99\linewidth]{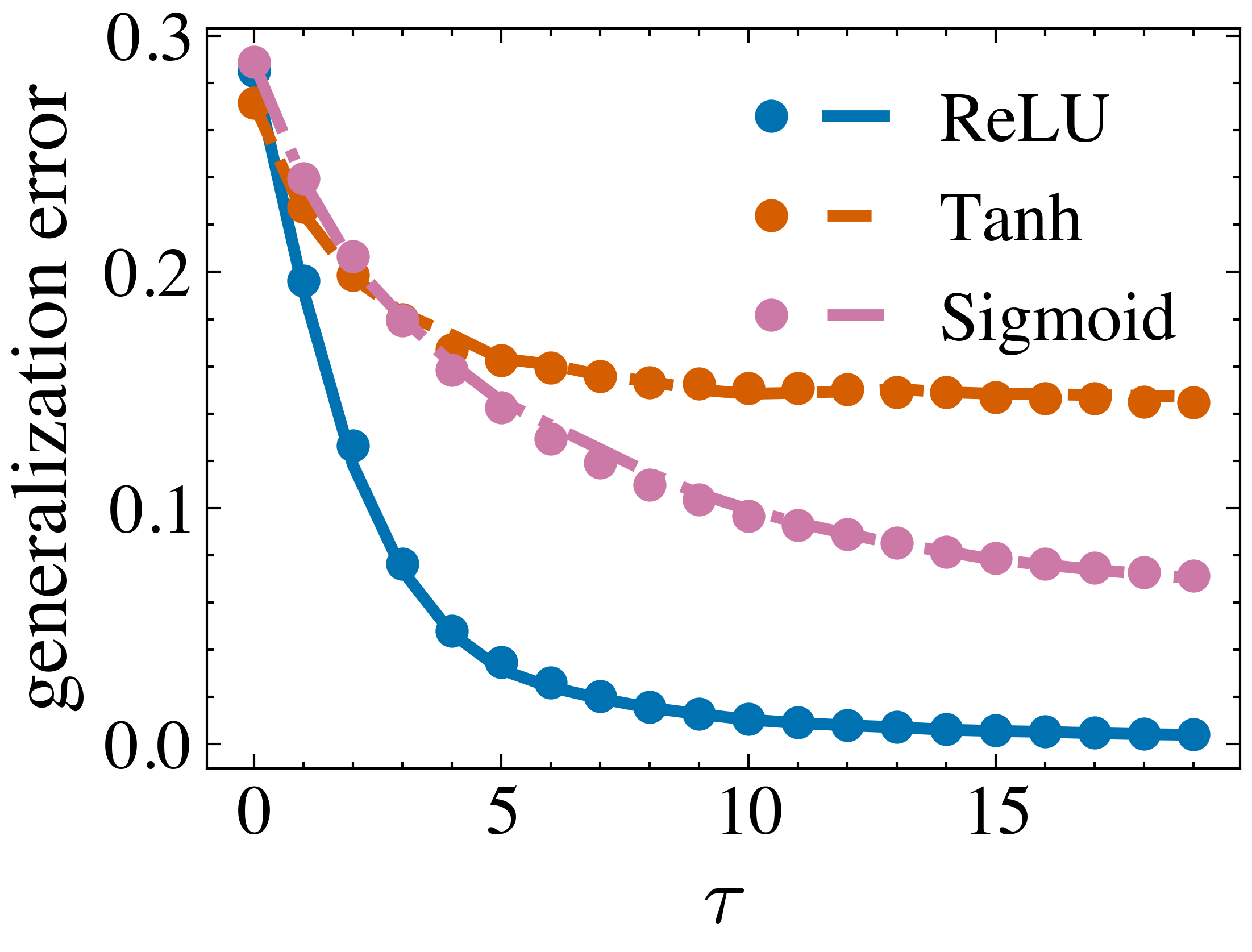}
         \caption{\centering activation function $\sigma$ ($\eta_a = \eta_f = 1.0$ and $\alpha = 1.0$)}
     \end{subfigure}

     \caption{Illustration of Theorem~\ref{thm:equivalence-spike}: asymptotic equivalence of test-error curves for the SGD and the Decomposed Dynamics under various conditions. Namely, (a) shows the effect of different learning rate $\eta_f =\eta_a = \eta$, (b) demonstrates the impact of different width $\alpha := k/n$, and finally, (c) indicates the influence of the activation function $\sigma(\cdot)$. Here, $n = 500, \tau = 20.0, k_\star = 1, g(\cdot)=\mathrm{ReLU}(\cdot)$, $\lambda=0$. Markers show SGD and lines show Decomposed Dynamics (DD).}
     \label{fig:equivalence}
\end{figure*}

Here, Proposition~\ref{prop:fast-alignment} states that the teacher-aligned part of the first layer lines up with the second layer within a vanishing fraction of training. While the proposition is proved for the oracle no-spike dynamics, Figure~\ref{fig:second_layer}b shows similarly rapid alignment under SGD.

\section{Additional simulation results}\label{sec:sims}

In this section, to illustrate the characteristics of feature learning, we compare the dynamics of four versions of two-layer neural networks: (i) the lazy training regime, where the first layer is fixed at initialization (random features \cite{RahimiRecht2007}); (ii) one-step feature learning, where the first layer is trained with one giant step of gradient descent and then frozen, as used in the literature to explain feature learning \cite{ba2022highdimensional, moniri2023theory, CuiEtAl2024OneStep, DandiEtAl2024GiantStep, demir2025asymptotic}; (iii) joint SGD training in the mean-field regime, following \eqref{eq:sgd-update-f}--\eqref{eq:sgd-update-a}; and (iv) Decomposed Dynamics (DD) with fixed $U_t$, where the complement directions stay at initialization while the signal matrix $M_t$ and the complement norms $r_t$ are the trained parameters of the first layer, updated by the components of the SGD step along $\Xi$ and along $u_{i,0}$, respectively (this variant uses the SGD step itself rather than the surrogate update \eqref{eq:dd-complement-update}; below, DD refers to this variant). While all algorithms use SGD for the second layer, the distinction lies in the training of the first layer. For random features and one-step feature learning, which serve as baselines for feature learning, the first layer is fixed while the second layer is trained. For SGD and DD, the two layers are trained simultaneously, capturing the coupled nature of feature learning.\footnote{\raggedright Code for all figures is available at \url{https://github.com/KU-MLIP/Decomposing-SGD-Dynamics-in-Neural-Networks}.}

\paragraph{Generalization dynamics ---} We start by comparing these algorithms in terms of generalization error over time in Figure~\ref{fig:comparison_of_algorithms}a. SGD ultimately outperforms one-step feature learning, which in turn improves on random features. One-step feature learning leads to a steep decrease at the beginning, while SGD enjoys a slower but steadier decrease over time. In this ReLU setting, DD with fixed complement directions improves further on SGD. Since SGD and DD grow the signal component at a similar rate (Figure~\ref{fig:comparison_of_algorithms}b), this improvement does not come from faster learning of the teacher direction. Freezing removes the teacher-induced spike channel, but it also stops the noise-driven growth of $r_t$ seen under SGD (Figure~\ref{fig:comparison_of_algorithms}c), so the comparison does not isolate the spike channel. The result is consistent with the prediction that training the complement directions can inflate variance even when the teacher-aligned component learns useful signal.

\begin{figure*}[t]
    \centering
    \begin{subfigure}[b]{0.32\textwidth}
         \centering
         \includegraphics[width=0.99\linewidth]{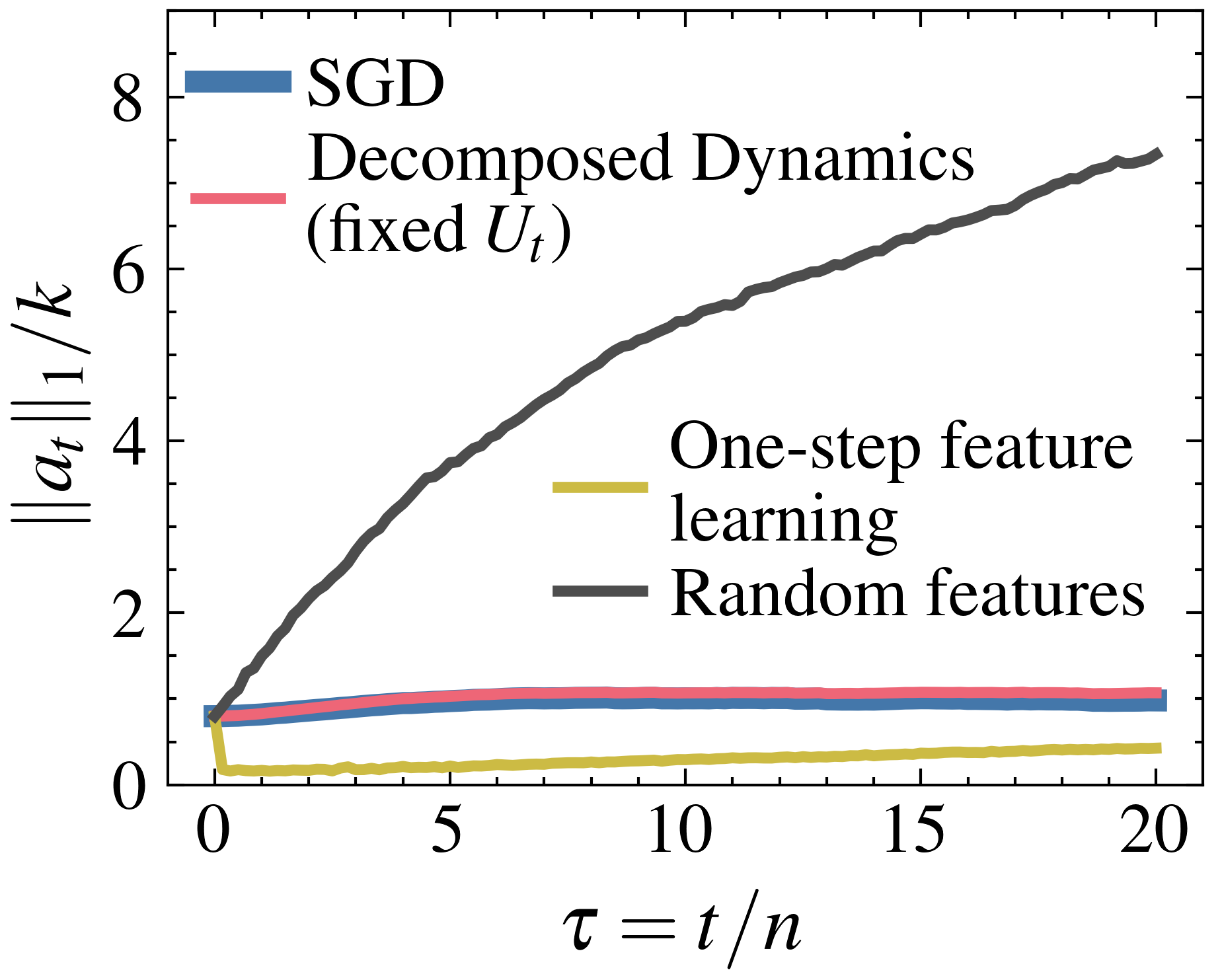}
         \caption{mean $|a|$}
     \end{subfigure}
     \hfill
    \begin{subfigure}[b]{0.32\textwidth}
         \centering
         \includegraphics[width=0.99\linewidth]{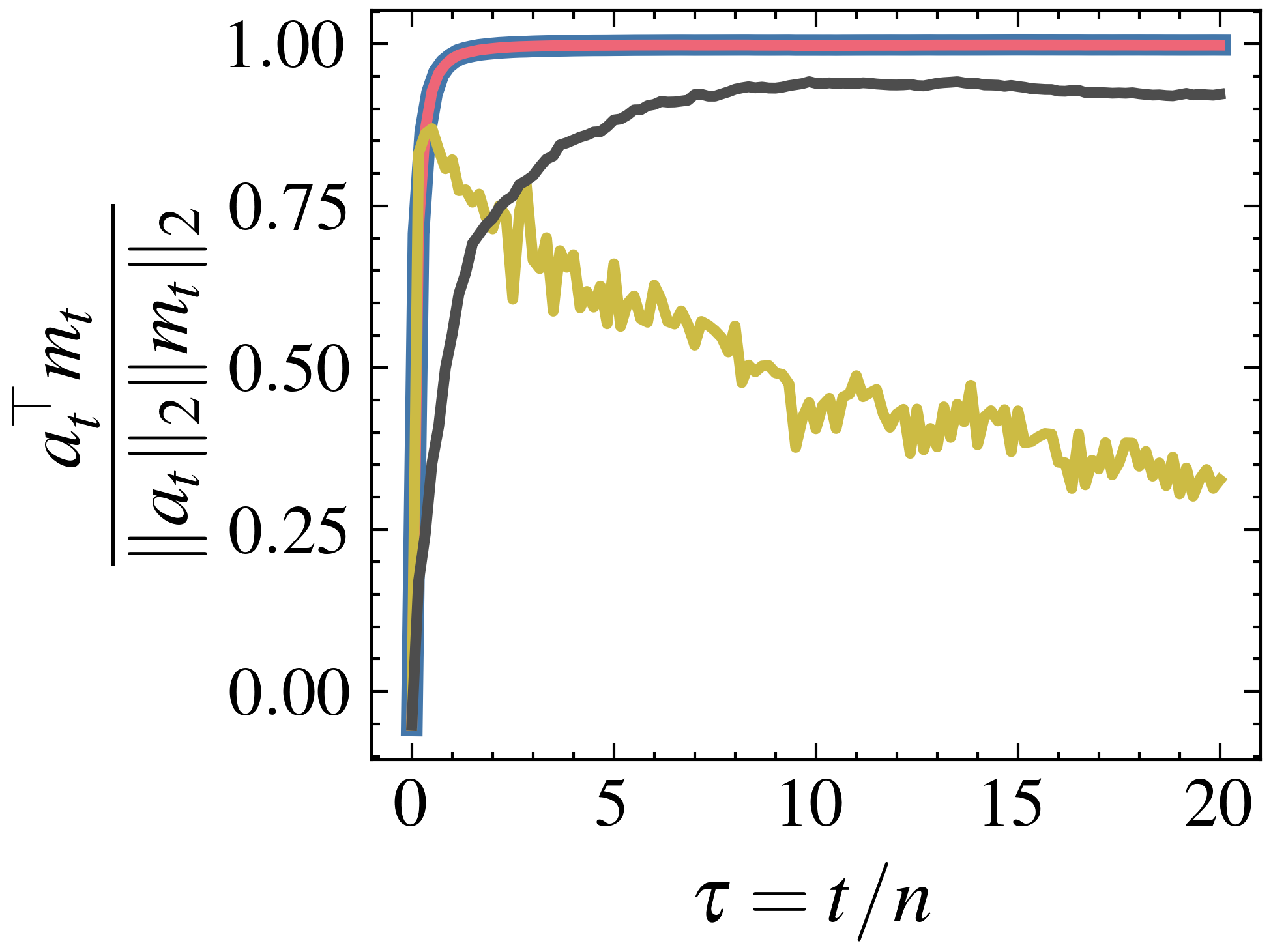}
         \caption{alignment of $a$ and $m$}
     \end{subfigure}
     \hfill
    \begin{subfigure}[b]{0.32\textwidth}
         \centering
         \includegraphics[width=0.99\linewidth]{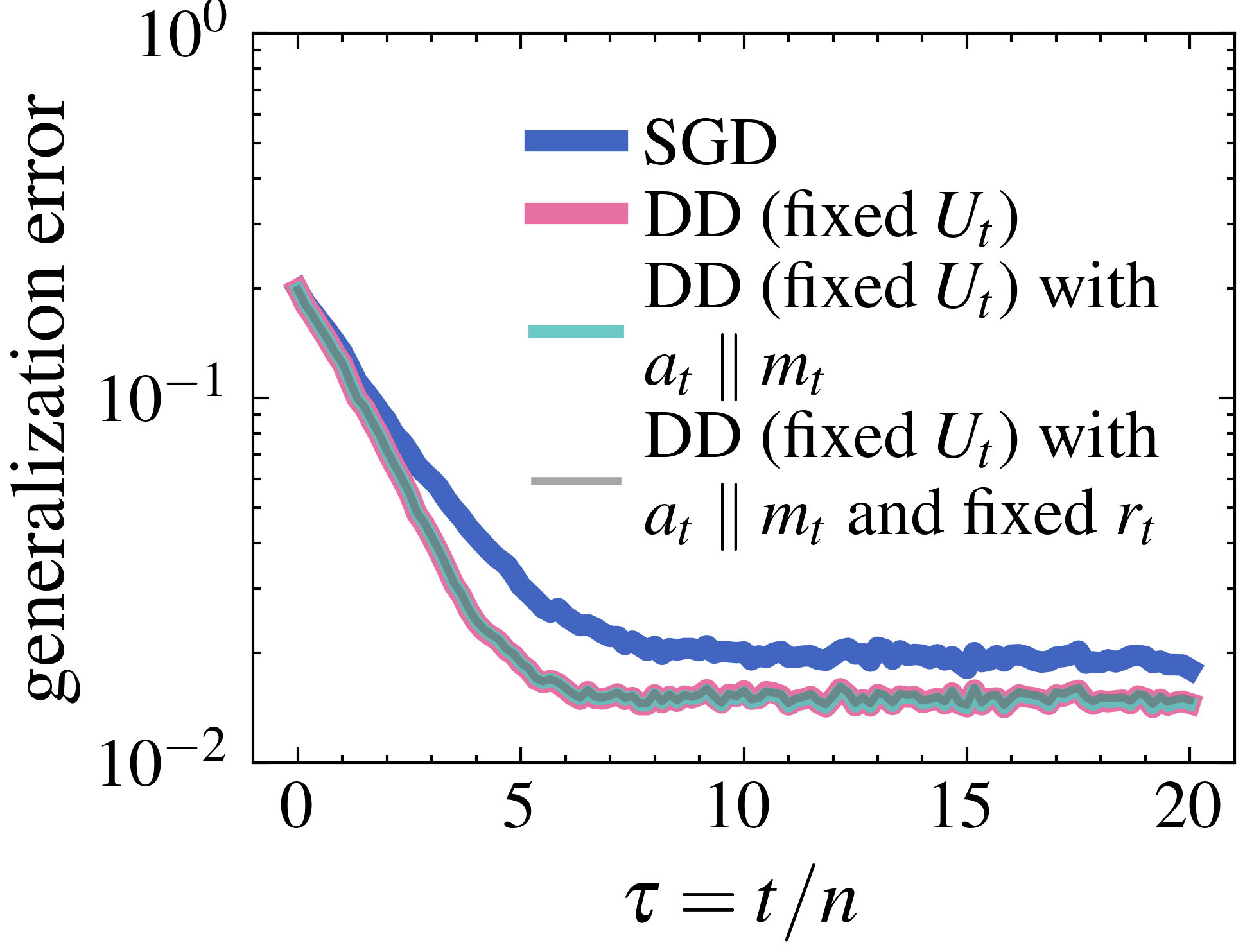}
         \caption{setting $m_t\parallel a_t$}
     \end{subfigure}
     \caption{Second-layer/signal coupling and reduced Decomposed Dynamics in the setting of Figure~\ref{fig:comparison_of_algorithms}. Panels show (a) mean second-layer magnitude $\|a_t\|_1/k$, (b) cosine alignment $a_t^\top m_t/(\|a_t\|_2\|m_t\|_2)$ for the same four algorithms, and (c) test-error curves for DD reductions that keep $U_t$ fixed and additionally impose $m_t\parallel a_t$, with and without fixed $r_t$, against SGD and DD with fixed $U_t$.}
     \label{fig:second_layer}
\end{figure*}

\paragraph{First-layer dynamics ---} Recalling $F_t=M_t\Xi^\top+\mathrm{diag}(r_t)U_t$, we track the signal matrix $M_t$ and the complement norms $r_t$ in Figures~\ref{fig:comparison_of_algorithms}b--c. Figure~\ref{fig:comparison_of_algorithms}b shows that the norm of $M_t$ rises during feature learning, indicating increasing alignment between the first-layer weights and the teacher subspace, since $M_t$ captures the aligned part of the first layer. SGD and DD show similar growth of $M_t$, whereas the norms of both $M_t$ and $r_t$ remain fixed for random features and jump sharply with the giant step of one-step feature learning. In Figure~\ref{fig:comparison_of_algorithms}c, $\|r_t\|_1/k$ grows under SGD and stays nearly constant under DD, which motivates the fixed-$r_t$ reduction tested in Figure~\ref{fig:second_layer}c and the unit-variance complement channel of Proposition~\ref{prop:fast-alignment}.

\paragraph{Second-layer dynamics ---} In Figure~\ref{fig:second_layer}a, the norm $\|a_t\|_1/k$ of the second-layer weights changes drastically for random features and one-step feature learning to compensate for the fixed first layer, whereas SGD and DD display similar, steady behavior with a slight increase over time.

\paragraph{Coupling of the layers ---} Figure~\ref{fig:second_layer}b shows the coupled dynamics of the two layers through the alignment between the signal matrix $M_t$ of the first layer and the second-layer weights $a_t$. SGD and DD have nearly identical alignment evolution over time: the alignment increases rapidly for both, consistent with Proposition~\ref{prop:fast-alignment}. One-step feature learning aligns after the large update and then gradually loses alignment, with fluctuations, while the second layer adapts to the frozen first layer. Random features also gain alignment over time, even though their first layer is fixed: the second layer upweights the fixed neurons that are already better aligned with the target subspace.

\paragraph{Unnecessary degrees of freedom ---} For DD, the near-constancy of $r_t$ and the fast alignment of $M_t$ with $a_t$ motivate us to fix $r_t$ at initialization and to set $M_t=a_t \nu_t^\top$ with a new trainable parameter $\nu_t\in\R^{k_\star}$ (updated by projecting the DD step for $M_t$ onto $a_t$). Figure~\ref{fig:second_layer}c shows that fixing $r_t$ at initialization and setting the signal matrix parallel to the second layer leaves the test-error dynamics essentially unchanged, while the number of trainable parameters drops from $k(n+1)$ to $k+k_\star$. Thus, especially when $k_\star\ll n$, many raw first-layer degrees of freedom appear unnecessary for this teacher--student task. These variants use the teacher subspace, so they are diagnostic interventions; a teacher-agnostic analogue would train a low-rank update on frozen random features.

\paragraph{Summary ---} Together, the simulations show that SGD builds spikes in the complement Gram matrix (Figure~\ref{fig:geometry_and_gram}), freezing the complement directions improves the test error in the ReLU setting (Figure~\ref{fig:comparison_of_algorithms}a), the signal component aligns quickly with the second layer (Figure~\ref{fig:second_layer}b), and a reduced model with $k+k_\star$ trainable parameters matches the frozen-direction variant (Figure~\ref{fig:second_layer}c).

\section{Conclusion}
\label{sec:conclusion}

We decomposed two-layer feature learning into a teacher-aligned signal component and a high-dimensional teacher-orthogonal complement. Even when complement updates are initially isotropic, SGD injects \emph{teacher-dependent structure} that creates \emph{spectral spikes} in the complement Gram matrix, giving a mechanism for emergent anisotropy and variance inflation. Decomposed Dynamics isolates this geometry through its complement update and matches SGD learning curves asymptotically, and the decomposition enables counterfactual experiments showing that constraining or freezing the complement can improve generalization. The same viewpoint reveals rapid cross-layer alignment between the signal component and second-layer weights, suggesting far fewer effective degrees of freedom than the raw parameterization. These results sharpen the role of nominally “irrelevant” directions and motivate interventions such as complement regularization, freezing, or low-rank training. In misspecified settings or under task shift, however, the nominal complement may carry signal, and structure learned there could help.

\begin{ack}
This work was supported in part by TÜBİTAK under the ARDEB 1001 program (Project No. 124E063). S.D. was also supported by a KUIS AI Center AI Fellowship and a TÜBİTAK BİDEB 2211 PhD Scholarship. The authors declare no competing interests.
\end{ack}

\bibliographystyle{unsrt}
\bibliography{ref}

\newpage
\appendix

\section{Related Work}
\label{app:related-work}

Our work touches several closely related lines in the theory of feature learning and high-dimensional SGD: teacher--student models, mean-field and DMFT analyses of training dynamics, Gaussian universality, spiked random matrix theory, and implicit regularization in overparameterized models. Below we situate our contributions against each, focusing on distinctions rather than a comprehensive survey.

\subsection{Teacher--Student Models and Feature Learning}

Teacher--student frameworks have played a central role in theoretical studies of representation learning, as they allow generalization and optimization dynamics to be analyzed in a controlled setting. A large body of work studies two-layer networks trained by SGD or related algorithms on targets generated by a low-dimensional teacher, with particular emphasis on how learned features align with teacher directions \cite{GoldtEtAl2020TeacherStudentSGD,DandiEtAl2024GiantStep,CuiEtAl2024OneStep}. Closely related, \cite{DamianLeeSoltanolkotabi2022} prove that, for degree-$p$ polynomial targets with a rank-$r$ expected Hessian, two-layer networks learn rank-$r$ multi-index representations with gradient descent in $O(d^2 r + d r^p)$ samples for input dimension $d$, \cite{BiettiBrunaSanfordSong2022} establish single-index recoverability with shallow networks driven by the information exponent, and \cite{MousaviHosseiniWuErdogdu2025MultiIndex} show that mean-field Langevin dynamics can learn multi-index models with complexity governed by an effective dimension. The staircase line of work \cite{AbbeBoixAdseraMisiakiewicz2022,AbbeBoixAdseraMisiakiewicz2023} characterizes which sparse and multi-index targets SGD recovers efficiently, identifying \emph{leap complexity} as the governing scale. Together, these works characterize when and how shallow networks learn teacher-relevant structure, including alignment dynamics, phase transitions, time scales, and effective dimension.

We complement these analyses by resolving SGD dynamics along \emph{both} teacher-aligned and teacher-orthogonal directions. Existing work typically concentrates on the aligned components and carries the orthogonal complement through low-dimensional state variables (fixed-width overlap matrices, norms, scalar kernels) or treats it as isotropic noise. \cite{VeigaStephanLoureiroKrzakalaZdeborova2022} extend the order-parameter ODEs of online learning to widths and learning rates that scale as powers of the input dimension; in their normalization our setting (width and learning rate both proportional to $n$) lies on the line of their phase diagram on which the SGD variance term in the update of the student overlap matrix (to leading order, the covariance injected into the complement) is of the same order as the learning term. Beyond fixed width, they use this term only through its scaling and its effect on the asymptotic risk, not through its structure across neurons. We show that the covariance SGD injects into the complement at every step contains a coherent, teacher-induced part (Lemmas~\ref{lem:complement-geometry}--\ref{lem:gram-onestep}), which produces anisotropic, spiked complement covariance (Remark~\ref{rem:bulk-plus-spikes} and Figure~\ref{fig:geometry_and_gram}), with concrete consequences for generalization (Section~\ref{sec:variance-inflation}). The fast-alignment phenomenon between $m_t$ and $a_t$ (Proposition~\ref{prop:fast-alignment}) further sharpens the timescale picture in this line.

\subsection{SGD Dynamics Beyond the NTK Regime}

A major theme in the theory of neural network training is the distinction between ``lazy'' training regimes, where features remain essentially fixed, and regimes in which features evolve substantially during optimization. The neural tangent kernel (NTK) framework formalizes the former perspective, relating training dynamics to kernel regression with random features \cite{RahimiRecht2007,JacotHG2018}. While this viewpoint has yielded powerful tools for understanding optimization and generalization, it inherently suppresses feature learning effects.

At the opposite extreme, mean-field and particle-based approaches describe feature learning at infinite width using PDE or interacting particle systems \cite{MeiMontanariNguyen2018,ChizatBach2018OT,RotskoffVandenEijnden2018,SirignanoSpiliopoulos2020}, with rigorous multilayer extensions in \cite{NguyenPham2023Mean}. \cite{WeiLeeLiuMa2019} establish a fundamental separation between mean-field (feature-learning) and NTK (lazy) regimes for two-layer networks, and \cite{Chizat2022Sparse} analyzes sparse measure-valued gradient descent in the over-parameterized limit. These analyses capture nontrivial feature evolution but often abstract away fine-grained covariance structure by averaging over particles or weights. Our equivalence proof (Theorem~\ref{thm:equivalence-spike}) rests on a stability estimate for the reduced state $(a_t,M_t,\Sigma_t)$ (Lemma~\ref{lem:S}), in the spirit of propagation of chaos, while retaining anisotropic complement covariances absent in standard mean-field descriptions.

Dynamical mean-field theory (DMFT), originating in spin-glass physics \cite{CugliandoloKurchan1994}, provides a complementary self-consistent description of SGD in high dimensions. In machine-learning settings, \cite{MignaccoKrzakalaUrbaniZdeborova2020} apply DMFT to SGD on Gaussian-mixture classification, with rigorous foundations in \cite{Gerbelot2024DMFT}; \cite{BenArousGheissariJagannath2024} establish high-dimensional limit theorems with effective dynamics and critical scalings, and \cite{SaraoMannelli2019Spiked} apply DMFT to gradient flow on a spiked matrix-tensor landscape. \cite{BordelonPehlevan2022DMFT} derive a self-consistent dynamical field theory of kernel evolution in wide networks. \cite{ArnaboldiStephanKrzakalaLoureiro2023} bridge high-dimensional and mean-field descriptions of two-layer SGD via dimensionless ODEs, showing that infinite-width dynamics remain close to a low-dimensional subspace spanned by target principal directions. Our equivalence proof for Decomposed Dynamics uses stability estimates (Lemma~\ref{lem:S}) in the spirit of \cite{Gerbelot2024DMFT,BenArousGheissariJagannath2024}, but specialized to the $(M_t,F_t^\perp)$ split of the first layer.

Our analysis sits between the lazy and mean-field regimes: features evolve nontrivially under SGD, but the decomposition $F_t = M_t\Xi^\top + F_t^\perp$ separates a low-dimensional teacher subspace from a high-dimensional complement whose structure we track. This is enough to derive an analytically tractable surrogate (Decomposed Dynamics, Theorem~\ref{thm:equivalence-spike}) whose test-error dynamics match those of SGD in the proportional limit for bounded $a_{i,0}$. The takeaway is that even when mean-field descriptions are accurate at the level of marginal distributions, higher-order structure in the teacher-orthogonal directions can shape generalization.

\subsection{Gaussian Universality and Equivalence}

A growing body of work establishes \emph{Gaussian universality} (or \emph{Gaussian equivalence}): in high-dimensional asymptotics, certain non-Gaussian feature distributions can be replaced, without changing learning curves at leading order, by Gaussian surrogates with matched first and second moments. \cite{GoldtLoureiro2022GaussEquiv} formalize this principle for shallow networks acting on data from generative models, and \cite{HuLu2023Universality} prove universality laws for random-feature models via a Lindeberg/Stein argument. \cite{MontanariSaeed2022Universality} extend universality to general empirical risk minimization in the proportional regime, and \cite{SchroderCuiDmitrievLoureiro2023} establish deterministic equivalents and error universality for deep random features.

Our results provide a complementary, training-time form of equivalence. The Decomposed Dynamics surrogate (Theorem~\ref{thm:equivalence-spike}) replaces the non-Gaussian rank-one complement increment $-(\eta_f/k)v_tg_t^\top$ by a Gaussian increment with the same conditional mean and, to leading order, the same neuron-space second moment, in line with the universality principle; the bulk-plus-spike replacement $n_\perp G_t \approx \hat c_t I_k + B_t^{\mathrm{acc}}(B_t^{\mathrm{acc}})^\top$ in the effective predictor (Eq.~\eqref{eq:yhat-eff}) further approximates the covariance of the (exactly Gaussian) complement preactivations. Whereas the works above establish equivalence at fixed (typically untrained) features, or apply it to one-pass SGD of a fixed-width student \cite{GoldtLoureiro2022GaussEquiv}, our equivalence is for the test-loss-relevant reduced state $(a_t,M_t,\Sigma_t)$ at proportional width along the SGD trajectory, which the Gaussian surrogate reproduces through its conditional mean increments.

\subsection{Random Matrix Theory and Spiked Models}

The emergence of spectral spikes in covariance or Gram matrices is a classical topic in random matrix theory, with foundational results in spiked covariance and finite-rank deformation models \cite{Johnstone2001,BaikBenArousPeche2005,BenaychGeorgesNadakuditi2009}. In these settings, low-rank structure is imposed exogenously through the data-generating process, and spectral outliers encode signal directions. Closer to neural network training, \cite{FanWang2020} establish the spectra of the Conjugate Kernel and Neural Tangent Kernel in the linear-width regime via iterated Marchenko--Pastur fixed-point equations, providing the bulk reference for training-time spectra. \cite{BaErdogduSuzukiWangWu2023} study learning in the presence of low-dimensional input structure via a spiked random matrix perspective, where spikes originate in the data covariance. Closer to our setting, one or a few large gradient steps create spikes in the first-layer weights that align with teacher-relevant (polynomial) directions and improve learning \cite{ba2022highdimensional,moniri2023theory,CuiEtAl2024OneStep,DandiEtAl2024GiantStep}.

In contrast, the spikes we identify are \emph{dynamically generated by SGD} (Remark~\ref{rem:bulk-plus-spikes}): the data distribution imposes no anisotropy in the teacher-orthogonal subspace, yet a few spectral outliers emerge from the interaction between SGD updates, the second-layer weights, and the teacher-aligned signal, above a bulk resembling Marchenko--Pastur (Figure~\ref{fig:geometry_and_gram}). This yields a training-time mechanism for spike formation, within reach of random matrix theory, that differs from classical spiked models and from one-step spikes: ours form in the teacher-orthogonal complement under online SGD and carry no target signal.

\subsection{Implicit Regularization, Variance Control, and Generalization}

A complementary line of work studies SGD through stochastic-process and diffusion approximations, relating training dynamics to stationary distributions and implicit regularization effects \cite{MandtHoffmanBlei2017,LiTaiE2017SME}. Separately, work on interpolation, ridgeless regression, and double descent highlights the central role of variance control in overparameterized models \cite{HastieEtAl2019Ridgeless,BelkinEtAl2018DoubleDescent}. Closer to the proportional regime studied here, \cite{MeiMontanari2022RF} derive precise asymptotics for the generalization error of random features and characterize the double-descent curve and the role of effective ridge regularization, while \cite{CanatarBordelonPehlevan2021} decompose generalization through the eigenstructure of the data--feature kernel, providing a spectrum-aware bias--variance picture.

Our results sharpen these perspectives for two-layer feature learning. Proposition~\ref{prop:variance-inflation} identifies the teacher-orthogonal complement as a concrete, structural source of predictor variance, arising from dynamically induced anisotropy rather than data covariance. Empirically (Figure~\ref{fig:comparison_of_algorithms}), freezing the complement directions improves generalization, exposing a much smaller effective number of degrees of freedom than the raw $k\times n$ parameterization and clarifying how such effects coexist with active feature learning in the teacher-aligned subspace.

\section{Neuron decoupling}
\label{app:decoupling}

This appendix provides a decoupling bound for the
\emph{centered} SGD drivers conditional on the teacher signal. We present the single-index case ($k_\star=1$) to
streamline notation.

\subsection{Filtration and notation}
Fix an SGD iterate $t$. Let $\mathcal F_t$ be the $\sigma$-field generated by all randomness up to time $t$
(including initialization and samples $\{x_s\}_{s<t}$), so that $(F_t,a_t)$ is $\mathcal F_t$-measurable.
At time $t$, a fresh sample $x_t\sim\mathcal N(0,I_n)$ is drawn independently of $\mathcal F_t$.

Decompose $x_t=z_{0,t}\xi+g_t$ with $z_{0,t}=\xi^\top x_t\sim\mathcal N(0,1)$ and
$g_t=P_\perp x_t\sim\mathcal N(0,P_\perp)$ independent of $z_{0,t}$.
For each neuron $i$ define the $\mathcal F_t$-measurable decomposition
\begin{equation}
    f_{i,t}=m_{i,t}\xi+r_{i,t}u_{i,t},
    \qquad m_{i,t}=\xi^\top f_{i,t},\qquad
    r_{i,t}=\|P_\perp f_{i,t}\|_2,\qquad
    u_{i,t}=\frac{P_\perp f_{i,t}}{\|P_\perp f_{i,t}\|_2}\in \xi^\perp.
    \label{eq:app-row-decomp}
\end{equation}
Then the preactivation is
\begin{equation}
    h_{i,t}=f_{i,t}^\top x_t = m_{i,t} z_{0,t}+ r_{i,t} z_{i,t},
    \qquad z_{i,t}:=u_{i,t}^\top g_t.
    \label{eq:app-preact}
\end{equation}

Define the student prediction and error
\[
    \hat y_t=\frac1k\sum_{\ell=1}^k a_{\ell,t}\sigma(h_{\ell,t}),\qquad
    y_t=g(z_{0,t}),\qquad
    e_t=\hat y_t-y_t,
\]
and the per-neuron SGD driver
\begin{equation}
    v_{i,t}:=e_t\,a_{i,t}\,\sigma'(h_{i,t}).
    \label{eq:app-v-def}
\end{equation}
Define the conditional mean and centered driver
\begin{equation}
    \mu_{i,t}(z_{0,t}) := \E[v_{i,t}\mid \mathcal F_t,z_{0,t}],
    \qquad
    \tilde v_{i,t} := v_{i,t} - \mu_{i,t}(z_{0,t}).
    \label{eq:app-centered-def}
\end{equation}

\subsection{A quantitative two-neuron decoupling bound (conditional, centered)}

We control $\Cov(\tilde v_{i,t},\tilde v_{j,t})$ via a Gaussian interpolation argument over the complement Gaussian
vector $z_t=(z_{1,t},\dots,z_{k,t})$ conditional on $(\mathcal F_t,z_{0,t})$.

\begin{lemma}[Conditional decoupling for centered drivers, weighted form]
\label{lem:app-decouple-centered}
Assume \textbf{A1}--\textbf{A5} and work on the event \textbf{A6}. Fix $\tau>0$, $t\le\tau n$ and $i\neq j$, and condition on $(\mathcal F_t,z_{0,t})$.
Let $\rho_{pq,t}:=u_{p,t}^\top u_{q,t}$, $\Delta_t:=\max_{p\neq q}|\rho_{pq,t}|\le 1$, and the empirical
coupled-second-moment $H_t:=\frac1k\sum_{p=1}^k a_{p,t}^2 r_{p,t}^2$. Define the per-pair weights
\begin{align}
    W^{(\rho)}_{ij,t} &:= |a_{i,t} a_{j,t}|\,r_{i,t}\,r_{j,t}, \nonumber\\
    W^{(\Delta)}_{ij,t} &:= |a_{i,t} a_{j,t}|\,\big(1 + r_{i,t} + r_{j,t}\big), \label{eq:app-decouple-weights}\\
    W^{(0)}_{ij,t} &:= a_{i,t}^2\,r_{i,t}^2\,|a_{j,t}| + a_{j,t}^2\,r_{j,t}^2\,|a_{i,t}| + |a_{i,t} a_{j,t}|\,H_t. \nonumber
\end{align}
Then
\begin{equation}
    \big|\Cov(\tilde v_{i,t},\tilde v_{j,t}\mid \mathcal F_t,z_{0,t})\big|
    \;\le\; C\Big[W^{(\rho)}_{ij,t}\,|\rho_{ij,t}|
                 \;+\; W^{(\Delta)}_{ij,t}\,\Delta_t
                 \;+\; \tfrac{1}{k}\,W^{(0)}_{ij,t}\Big],
    \label{eq:app-decouple-centered}
\end{equation}
where $C<\infty$ is a constant depending only on $\|\sigma\|_\infty$, $\|\sigma'\|_\infty$, $\|\sigma''\|_\infty$, $\|g\|_\infty$,
and $B_\tau$ from Lemma~\ref{lem:stability-sufficient}. The fourth-moment-type quantity $H_t$ is left explicit: it
is bounded by $B'_\tau(n)=O(\log n)$ via Lemma~\ref{lem:stability-sufficient-4th}, but this is consumed only
by downstream sums, not absorbed here.
Consequently, since the weights and $\rho_{ij,t},\Delta_t$ are $\mathcal F_t$-measurable,
\[
    \big|\E[\tilde v_{i,t}\tilde v_{j,t}\mid \mathcal F_t]\big|
    \;\le\; C\Big[W^{(\rho)}_{ij,t}\,|\rho_{ij,t}|
                 \;+\; W^{(\Delta)}_{ij,t}\,\Delta_t
                 \;+\; \tfrac{1}{k}\,W^{(0)}_{ij,t}\Big].
\]
\end{lemma}

\paragraph{Proof.}
Fix $(\mathcal F_t,z_{0,t})$. The complement projection vector $z_t=(z_{1,t},\dots,z_{k,t})$ is centered Gaussian with
covariance $\Sigma^u_t=U_tU_t^\top$, i.e., $(\Sigma^u_t)_{pq}=\rho_{pq,t}$ and $(\Sigma^u_t)_{pp}=1$.
Write $\tilde v_{i,t}=F_i(z_t)-\E[F_i(z_t)\mid\mathcal F_t,z_{0,t}]$ where
\[
F_i(z_t)= e_t(z_t)\,a_{i,t}\,\sigma'(m_{i,t}z_{0,t}+r_{i,t}z_{i,t})
\]
and $e_t(z_t)=\frac1k\sum_{\ell}a_{\ell,t}\sigma(m_{\ell,t}z_{0,t}+r_{\ell,t}z_{\ell,t})-g(z_{0,t})$.
Since subtracting constants does not affect covariance, it suffices to bound $\Cov(F_i(z_t),F_j(z_t))$.

Under \textbf{A3}, $F_i$ is Lipschitz in $z_t$ with square-integrable weak gradient.
Gaussian interpolation \cite[Lemma 1.3.1]{talagrand2010mean} for Gaussian vectors implies
\begin{equation}
\Cov(F_i(z_t),F_j(z_t))
=
\int_0^1
\E\!\left[
    \nabla F_i(Z)^\top \Sigma^u_t\, \nabla F_j(Z^{(\theta)})
    \,\middle|\,\mathcal F_t,z_{0,t}
\right] d\theta,
\label{eq:app-mehler-cov}
\end{equation}
where $(Z,Z^{(\theta)})$ are jointly Gaussian with marginal $\mathcal N(0,\Sigma^u_t)$ and cross-covariance
$\E[Z\,Z^{(\theta)\top}]=\theta\,\Sigma^u_t$.
Taking absolute values and applying Cauchy--Schwarz yields
\begin{equation}
|\Cov(F_i,F_j)|
\;\le\;
\sum_{p,q=1}^k |\rho_{pq,t}|\,
\Big(\E[(\partial_p F_i(Z))^2\mid\mathcal F_t,z_{0,t}]\Big)^{1/2}
\Big(\E[(\partial_q F_j(Z))^2\mid\mathcal F_t,z_{0,t}]\Big)^{1/2}.
\label{eq:app-cov-upper}
\end{equation}

We bound the weak derivatives using the chain rule.
For $p\neq i$,
\[
\partial_p F_i(z_t) = (\partial_p e_t(z_t))\cdot a_{i,t}\sigma'(h_{i,t})
= \frac{1}{k}\, a_{p,t}r_{p,t}\sigma'(h_{p,t})\cdot a_{i,t}\sigma'(h_{i,t}),
\]
hence by bounded $\sigma'$,
\begin{equation}
\E[(\partial_pF_i)^2\mid\mathcal F_t,z_{0,t}]
\;\le\; \frac{C}{k^2}\,a_{i,t}^2 a_{p,t}^2 r_{p,t}^2
\qquad(p\neq i).
\label{eq:app-deriv-offdiag}
\end{equation}
For $p=i$,
\[
\partial_i F_i(z_t)
= (\partial_i e_t)\,a_{i,t}\sigma'(h_{i,t})
  + e_t\cdot a_{i,t}r_{i,t}\,\sigma''(h_{i,t}),
\qquad
\partial_i e_t=\frac{1}{k}a_{i,t}r_{i,t}\sigma'(h_{i,t}),
\]
so bounded $\sigma',\sigma''$ implies
\begin{equation}
\E[(\partial_iF_i)^2\mid\mathcal F_t,z_{0,t}]
\;\le\;
C\Big(\frac{a_{i,t}^4 r_{i,t}^2}{k^2} + a_{i,t}^2 r_{i,t}^2\,\E[e_t^2\mid\mathcal F_t,z_{0,t}]\Big).
\label{eq:app-deriv-diag}
\end{equation}

On the high-probability event in \textbf{A6} we have $\frac1k\sum_\ell a_{\ell,t}^2 \le B_\tau$ and
$\frac1k\sum_\ell r_{\ell,t}^2 \le B_\tau$, and since $\sigma,g$ are bounded,
\[
    |e_t(z_t)|\;\le\; \|\sigma\|_\infty\Big(\tfrac1k\textstyle\sum_\ell a_{\ell,t}^2\Big)^{1/2} + \|g\|_\infty
    \;\le\; \|\sigma\|_\infty\sqrt{B_\tau}+\|g\|_\infty,
\]
uniformly in $z_t$. In particular $\max_\ell a_{\ell,t}^2\le \sum_\ell a_{\ell,t}^2\le kB_\tau$, so
$a_{i,t}^4/k^2\le B_\tau\,a_{i,t}^2$, and \eqref{eq:app-deriv-diag} simplifies to
\begin{equation}
    \E[(\partial_iF_i)^2\mid\mathcal F_t,z_{0,t}]\;\le\; C_1\,a_{i,t}^2\,r_{i,t}^2,
    \qquad
    \E[(\partial_pF_i)^2\mid\mathcal F_t,z_{0,t}]\;\le\;\frac{C_1}{k^2}\,a_{i,t}^2\,a_{p,t}^2\,r_{p,t}^2\quad(p\ne i),
    \label{eq:app-deriv-clean}
\end{equation}
with $C_1$ depending only on $\|\sigma\|_\infty,\|\sigma'\|_\infty,\|\sigma''\|_\infty,\|g\|_\infty,B_\tau$, and the analogous
bounds for $\partial_q F_j$. Insert \eqref{eq:app-deriv-clean} into \eqref{eq:app-cov-upper} and partition
$(p,q)\in\{1,\dots,k\}^2$ into four blocks.

\smallskip
\noindent\textit{(T1) $(p,q)\in\{i,j\}\times\{i,j\}$.}
The four cells contribute, using $|\rho_{ii}|=|\rho_{jj}|=1$ and $|\rho_{ij}|=|\rho_{ji}|$:
\begin{itemize}[leftmargin=2em,topsep=0pt,itemsep=-1ex]
    \item $(p,q)=(i,j)$: $|\rho_{ij,t}|\cdot C_1\,|a_{i,t}a_{j,t}|r_{i,t}r_{j,t}\;=\;C_1\,W^{(\rho)}_{ij,t}\,|\rho_{ij,t}|$.
    \item $(p,q)=(j,i)$: same with an extra factor $|a_{i,t}a_{j,t}|/k^2$ (both derivatives off-diagonal); subleading and absorbed into $C_1\,W^{(\rho)}_{ij,t}\,|\rho_{ij,t}|$ via $|a_{i,t}a_{j,t}|/k^2\le B_\tau/k$.
    \item $(p,q)=(i,i)$: $1\cdot\tfrac{C_1}{k}\,a_{i,t}^2 r_{i,t}^2 |a_{j,t}|$, part of $\tfrac{1}{k}\,W^{(0)}_{ij,t}$.
    \item $(p,q)=(j,j)$: symmetric, part of $\tfrac{1}{k}\,W^{(0)}_{ij,t}$.
\end{itemize}

\smallskip
\noindent\textit{(T2) $p\in\{i,j\}$, $q\notin\{i,j\}$, and (T3) symmetric.}
For $p=i$, using $|\rho_{iq,t}|\le \Delta_t$ for $q\ne i$ and Cauchy--Schwarz $\sum_q|a_{q,t}|r_{q,t}\le\sqrt{\sum a^2}\sqrt{\sum r^2}\le kB_\tau$,
\[
    \sum_{q\notin\{i,j\}}|\rho_{iq,t}|\cdot \tfrac{C_1}{k}|a_{i,t}a_{j,t}|r_{i,t}|a_{q,t}|r_{q,t}
    \;\le\; C_1 B_\tau\,\Delta_t\,|a_{i,t}a_{j,t}|r_{i,t}
    \;\le\; C_1 B_\tau\,\Delta_t\,W^{(\Delta)}_{ij,t}.
\]
The case $p=j$ produces an $r_j$-analogue carrying an additional factor $|a_{j,t}|/k\le\sqrt{B_\tau/k}$ (both derivatives off-diagonal),
absorbed into $C_1 B_\tau\,\Delta_t\,W^{(\Delta)}_{ij,t}$. (T3) is symmetric.

\smallskip
\noindent\textit{(T4) $p,q\notin\{i,j\}$.}
Split into diagonal $p=q$ and off-diagonal $p\ne q$:
\begin{itemize}[leftmargin=2em,topsep=0pt,itemsep=-1ex]
    \item Diagonal: $\sum_{p\notin\{i,j\}}1\cdot\tfrac{C_1}{k^2}|a_{i,t}a_{j,t}|a_{p,t}^2 r_{p,t}^2 \le \tfrac{C_1}{k^2}|a_{i,t}a_{j,t}|\sum_p a_{p,t}^2 r_{p,t}^2 = \tfrac{C_1}{k}\,|a_{i,t}a_{j,t}|\,H_t$,
    which is exactly the $|a_ia_j|H_t$ piece of $\tfrac{1}{k}W^{(0)}_{ij,t}$ (no fourth-moment input needed at this stage).
    \item Off-diagonal: $\sum_{p\ne q\notin\{i,j\}}|\rho_{pq,t}|\cdot\tfrac{C_1}{k^2}|a_{i,t}a_{j,t}||a_{p,t}a_{q,t}|r_{p,t}r_{q,t}
    \le \Delta_t\,\tfrac{C_1}{k^2}|a_{i,t}a_{j,t}|\big(\sum_p|a_{p,t}|r_{p,t}\big)^2 \le C_1 B_\tau^2\,\Delta_t\,|a_{i,t}a_{j,t}|$,
    which is $\le C_1 B_\tau^2\,\Delta_t\,W^{(\Delta)}_{ij,t}$.
\end{itemize}

Summing (T1)--(T4) and absorbing $B_\tau,B_\tau^2$ into a single constant $C$ yields
\eqref{eq:app-decouple-centered}.\qed

\section{Proof of Lemma~\ref{lem:complement-geometry} (complement geometry)}
\label{app:complement-geometry-proof}

\subsection{Decomposition of the driver covariance accumulator}

Expanding $v_s v_s^\top$ gives
\[
v_s v_s^\top = \tilde v_s\tilde v_s^\top + \mu_s\mu_s^\top + \mu_s\tilde v_s^\top + \tilde v_s\mu_s^\top,
\]
where $\mu_s=\mu_s(z_{0,s})$ is the conditional mean vector and $\tilde v_s=v_s-\mu_s$.
Summing and inserting into \eqref{eq:C-accum} yields \eqref{eq:C-decomp} with
\[
\widetilde C_t=\frac1n I_k+\Big(\frac{\eta_f}{k}\Big)^2\sum_{s=0}^{t-1}\tilde v_s\tilde v_s^\top,\qquad
S_s=\E[\mu_s\mu_s^\top\mid\mathcal F_s],
\]
and
\begin{align*}
R_t
&=
\Big(\frac{\eta_f}{k}\Big)^2\sum_{s=0}^{t-1}\Big(\mu_s\tilde v_s^\top+\tilde v_s\mu_s^\top\Big)
+\Big(\frac{\eta_f}{k}\Big)^2\sum_{s=0}^{t-1}\Big(\mu_s\mu_s^\top-\E[\mu_s\mu_s^\top\mid\mathcal F_s]\Big).
\end{align*}

\paragraph{Frobenius-level martingale control for $R_t$.}
Let $\mathcal G_s := \sigma(\mathcal F_s, z_{0,s})$ denote the $\sigma$-field obtained by enlarging
$\mathcal F_s$ with $z_{0,s}$. By definition, $\mu_s$ is $\mathcal G_s$-measurable, and
$\E[\tilde v_s\mid\mathcal G_s]=0$, hence $\mu_s\tilde v_s^\top + \tilde v_s\mu_s^\top$ is a martingale difference
for the filtration $(\mathcal G_{s+1})_{s\ge 0}$. Separately,
$\mu_s\mu_s^\top - \E[\mu_s\mu_s^\top\mid\mathcal F_s]$ is centered under the original $\mathcal F_s$ (with
$S_s := \E[\mu_s\mu_s^\top\mid\mathcal F_s]$), and is a martingale difference for
$(\mathcal F_{s+1})_{s\ge 0}$ since $\mu_s\mu_s^\top \in \mathcal F_{s+1}$.

The martingale bounds in this appendix are applied to the increments stopped at the $(\mathcal F_s)$-stopping time $T:=\inf\{s:\|a_s\|_2^2>kB_\tau\}\wedge\lfloor\tau n\rfloor$, for which they hold whenever $s<T$. The stopped processes $R_{t\wedge T}$ and $N_{t\wedge T}$ (below) are martingales with the stated second moments, and $T=\lfloor\tau n\rfloor$ on \textbf{A6}, so they coincide with $R_t$ and $N_t$ there.
For $s<T$ (so $\|a_s\|_2^2\le kB_\tau$) and bounded $\sigma,\sigma'$, we have
\[
\|\mu_s\|_2^2 \le \E[\|v_s\|_2^2\mid\mathcal G_s]\lesssim \|a_s\|_2^2 = O(k),
\qquad
\E[\|\tilde v_s\|_2^2\mid\mathcal G_s]\lesssim \|a_s\|_2^2 = O(k),
\]
and therefore
\[
\E\big[\|\mu_s\tilde v_s^\top+\tilde v_s\mu_s^\top\|_F^2\mid\mathcal G_s\big]
\;\le\; C\,\|\mu_s\|_2^2\,\E[\|\tilde v_s\|_2^2\mid\mathcal G_s]
\;=\;O(k^2),
\]
and similarly $\E[\|\mu_s\mu_s^\top-\E[\mu_s\mu_s^\top\mid\mathcal F_s]\|_F^2\mid\mathcal F_s]=O(k^2)$.
Hence, by orthogonality of martingale differences in $L^2(\|\cdot\|_F)$,
\[
\E\big[\|R_t\|_F^2\big]
\;\lesssim\;
\Big(\frac{\eta_f}{k}\Big)^4\sum_{s=0}^{t-1} O(k^2)
\;=\; O\Big(\frac{t}{k^2}\Big)
\;=\; O\Big(\frac{1}{n}\Big),
\]
since $t\le \tau n$ and $k\asymp n$.
By Doob's maximal inequality,
$\E[\sup_{t\le \tau n}\|R_t\|_F^2]\le 4\,\E[\|R_{\lfloor\tau n\rfloor}\|_F^2]=O(1/n)$,
so $\sup_{t\le\tau n}\|R_t\|_F \le \varepsilon_n$ with high probability for some deterministic $\varepsilon_n\to 0$.
This gives the remainder bound in \eqref{eq:C-decomp}.

\subsection{Residual isotropy}

By Lemma~\ref{lem:app-decouple-centered}, on the event \textbf{A6}, and using the trivial bounds
$|\rho_{ij,t}|,\Delta_t\le 1$,
\begin{equation}
    \big|\E[\tilde v_{i,t}\tilde v_{j,t}\mid\mathcal F_t]\big|
    \;\le\; C\,\Omega_{ij,t},
    \qquad
    \Omega_{ij,t}\;:=\; W^{(\rho)}_{ij,t} + W^{(\Delta)}_{ij,t} + \tfrac{1}{k}\,W^{(0)}_{ij,t},
    \qquad i\ne j,
    \label{eq:app-decouple-envelope}
\end{equation}
with the weights from \eqref{eq:app-decouple-weights} and $C$ the constant of Lemma~\ref{lem:app-decouple-centered}. By the explicit form of the
weights, A6, and Lemma~\ref{lem:stability-sufficient-4th} (which gives $H_t\le B'_\tau(n)=O(\log n)$ and
$\sum_p a_p^2 r_p^2 \le k B'_\tau(n)$), the per-pair envelope satisfies
\begin{equation}
    \sum_{i\ne j}\Omega_{ij,t}^2 \;\le\; k^2\,\mathrm{polylog}(n),
    \label{eq:app-Omega-sum}
\end{equation}
the dominant contribution being $\sum_{i,j}(W^{(\rho)}_{ij,t})^2 = \big(\sum_p a_{p,t}^2 r_{p,t}^2\big)^2 \le (kB'_\tau(n))^2$;
the $W^{(\Delta)}$ and $\tfrac{1}{k}W^{(0)}$ contributions are bounded by $O(k^2\log n)$ and $O(k\log^2 n)$, respectively.

Write
\[
\widetilde C_t = \frac1n I_k + \Big(\frac{\eta_f}{k}\Big)^2\sum_{s<t}\tilde v_s\tilde v_s^\top,
\qquad
c_t=\frac1k\Tr(\widetilde C_t),
\]
and decompose into drift plus martingale, $\widetilde C_t = D_t + N_t$, with
$D_t := \frac1n I_k + \Big(\frac{\eta_f}{k}\Big)^2\sum_{s<t}\E[\tilde v_s\tilde v_s^\top\mid\mathcal F_s]$
and $N_t$ the centered remainder. Then
$\|\widetilde C_t-c_tI_k\|_F \le \|D_t-\tfrac1k\Tr(D_t)I_k\|_F + 2\|N_t\|_F$.

\emph{Drift, off-diagonal control.}
For $i\ne j$, by \eqref{eq:app-decouple-envelope} and Cauchy--Schwarz on the time sum,
\[
    |(D_t)_{ij}|^2 \;\le\; \frac{C^2\eta_f^4\,t}{k^4}\sum_{s<t}\Omega_{ij,s}^2.
\]
Summing over $i\ne j$, applying \eqref{eq:app-Omega-sum} at each $s<t$, and using $t\le\tau n\lesssim k$,
\[
    \sum_{i\ne j}|(D_t)_{ij}|^2
    \;\le\; \frac{C^2\eta_f^4\,t^2}{k^4}\cdot k^2\,\mathrm{polylog}(n)
    \;=\; O(\mathrm{polylog}(n)),
    \qquad\text{hence}\qquad
    \frac{1}{k^2}\sum_{i\ne j}(D_t)_{ij}^2 \;=\; o(1).
\]
Combined with the diagonal term,
\[
\frac{1}{k^2}\Big\|D_t-\tfrac1k\Tr(D_t)I_k\Big\|_F^2
\le
\frac{1}{k^2}\sum_{i\ne j}(D_t)_{ij}^2
+\frac{1}{k^2}\sum_i\Big((D_t)_{ii}-\tfrac1k\Tr(D_t)\Big)^2
=o(1),
\]
where the diagonal term is $o(1)$ as follows. Since $D_t = \frac1n I_k + (\eta_f/k)^2\sum_{s<t}\E[\tilde v_s\tilde v_s^\top\mid\mathcal F_s]$
is a sum of PSD matrices, $D_t\succeq 0$, and taking trace gives
\[
    \Tr(D_t) \;=\; \frac{k}{n} + \Big(\frac{\eta_f}{k}\Big)^2\sum_{s<t}\E\big[\|\tilde v_s\|_2^2\,\big|\,\mathcal F_s\big].
\]
On \textbf{A6}, $\E[\|\tilde v_s\|_2^2\mid\mathcal F_s]\le C\sum_p a_{p,s}^2 = O(k)$ uniformly, so
$\Tr(D_t)\le k/n + (\eta_f/k)^2\cdot t\cdot O(k) = O(1)$ since $t\le\tau n\asymp k$.
Then $\sum_i((D_t)_{ii}-\tfrac1k\Tr(D_t))^2 \le \sum_i(D_t)_{ii}^2 \le \|D_t\|_F^2 \le \Tr(D_t)^2 = O(1)$ by PSD-ness,
giving $\frac{1}{k^2}\sum_i((D_t)_{ii}-\tfrac1k\Tr(D_t))^2 = o(1)$.

\emph{Martingale.}
Since $\|\tilde v_s\tilde v_s^\top\|_F=\|\tilde v_s\|_2^2$ and $|\Tr(N_t)|\le \sqrt{k}\|N_t\|_F$,
it suffices to control $\|N_t\|_F$.
Using orthogonality of martingale differences in $L^2(\|\cdot\|_F)$,
\[
\E[\|N_t\|_F^2]
\;\le\;
\Big(\frac{\eta_f}{k}\Big)^4\sum_{s=0}^{t-1} \E\big[\|\tilde v_s\tilde v_s^\top\|_F^2\big]
=
\Big(\frac{\eta_f}{k}\Big)^4\sum_{s=0}^{t-1} \E[\|\tilde v_s\|_2^4].
\]
For $s<T$ and bounded $\sigma,\sigma'$, we have $\|\tilde v_s\|_2\le \|v_s\|_2+\|\mu_s\|_2\lesssim \|a_s\|_2$ and
$\|a_s\|_2^2\le kB_\tau$, hence $\|\tilde v_s\|_2^4=O(k^2)$ and
\[
\E[\|N_t\|_F^2] \;=\; O\Big(\frac{t}{k^2}\Big)=O\Big(\frac{1}{n}\Big).
\]
By Doob's maximal inequality, $\E[\sup_{t\le \tau n}\|N_t\|_F^2]=O(1/n)$, hence
$\sup_{t\le\tau n}\|N_t\|_F = o(1)$ with high probability, and therefore
\[
\sup_{t\le\tau n}\frac{1}{k^2}\|N_t-\tfrac1k\Tr(N_t)I_k\|_F^2
\;\le\;
\sup_{t\le\tau n}\frac{4}{k^2}\|N_t\|_F^2
\;=\;o(1).
\]

Combining the drift and martingale bounds yields \eqref{eq:residual-isotropy} on the high-probability events
specified by Lemmas~\ref{lem:stability-sufficient}--\ref{lem:stability-sufficient-4th}, which jointly hold with
probability $1-o(1)$. The drift contribution is $o(1)$ in normalized Frobenius norm via the weighted form of
Lemma~\ref{lem:app-decouple-centered} together with the bound $H_t\le B'_\tau(n)$ from
Lemma~\ref{lem:stability-sufficient-4th}, and the martingale contribution is $o(1)$ by Doob's inequality
applied to the second-moment bound on $\|\tilde v_s\|_2^2$.\qed

\section{Proof of Lemma~\ref{lem:gram-onestep} (normalized complement Gram recursion)}
\label{app:gram-onestep-proof}

We prove Lemma~\ref{lem:gram-onestep}. For clarity we write the argument in the single-index case ($k_\star=1$);
the multi-index case follows by conditioning on $z_{\star,t}=\Xi^\top x_t$ throughout.

\subsection{Exact one-step identity}
Recall $F^{\perp}_t:=F_tP_\perp\in\R^{k\times n}$ and the complement update
\begin{equation}
F^{\perp}_{t+1}=F^{\perp}_t-\frac{\eta_f}{k}v_t g_t^\top,
\qquad g_t:=P_\perp x_t\sim\mathcal N(0,P_\perp),
\label{eq:app-complement-update}
\end{equation}
where $x_t$ is fresh and independent of $\mathcal F_t$, hence $g_t$ is independent of $\mathcal F_t$.
Let $n_\perp:=\Tr(P_\perp)$ and define
\[
G_t:=\frac{1}{n_\perp}F^{\perp}_t(F^{\perp}_t)^\top\in\R^{k\times k},
\qquad
h_t^\perp:=F^{\perp}_t g_t\in\R^k.
\]
Expanding gives the exact identity
\begin{equation}
G_{t+1}
=
G_t
-\frac{\eta_f}{k\,n_\perp}\Big(v_t h_t^{\perp\top}+h_t^\perp v_t^\top\Big)
+\Big(\frac{\eta_f}{k}\Big)^2\frac{\|g_t\|_2^2}{n_\perp}\,v_tv_t^\top.
\label{eq:app-gram-exact}
\end{equation}
Taking $\E_t[\cdot]$ yields
\begin{align}
\E_t[G_{t+1}]
&=
G_t
-\frac{\eta_f}{k\,n_\perp}\Big(\E_t[v_t h_t^{\perp\top}]+\E_t[h_t^\perp v_t^\top]\Big)
+\Big(\frac{\eta_f}{k}\Big)^2\E_t\!\left[\frac{\|g_t\|_2^2}{n_\perp}\,v_tv_t^\top\right].
\label{eq:app-gram-start}
\end{align}

\subsection{Quadratic term: remove $\|g_t\|_2^2/n_\perp$}
Let $\Delta_t:=\|g_t\|_2^2/n_\perp-1$. Since $g_t\sim\mathcal N(0,P_\perp)$,
\[
\E_t[\Delta_t]=0,
\qquad
\E_t[\Delta_t^2]=\Var(\|g_t\|_2^2/n_\perp)=\frac{2}{n_\perp}.
\]
Define
\[
E_t^{(g)}
:=
\Big(\frac{\eta_f}{k}\Big)^2
\E_t\!\big[\Delta_t\,v_tv_t^\top\big].
\]
By conditional Cauchy--Schwarz,
\[
\| \E_t[\Delta_t v_tv_t^\top]\|_F
\le
\sqrt{\E_t[\Delta_t^2]}\cdot \sqrt{\E_t[\|v_tv_t^\top\|_F^2]}
=
\sqrt{\frac{2}{n_\perp}}\cdot \sqrt{\E_t[\|v_t\|_2^4]}.
\]
Under \textbf{A2--A3}, $|\sigma|$ and $|\sigma'|$ are bounded and $g$ is bounded, hence
$|e_t|\lesssim 1+\|a_t\|_2/\sqrt{k}$.
On the \textbf{A6} event, $\sum_i a_{i,t}^2=O(k)$ uniformly over $t\le\tau n$, so $|e_t|=O(1)$, $\|v_t\|_2^2\lesssim \sum_{i=1}^k a_{i,t}^2=O(k)$ and $\E_t[\|v_t\|_2^4]=O(k^2)$.
Thus
\[
\| \E_t[\Delta_t v_tv_t^\top]\|_F
=
O\!\Big(\frac{k}{\sqrt{n_\perp}}\Big),
\qquad\Rightarrow\qquad
\|E_t^{(g)}\|_F
=
O\!\Big(\frac{1}{k\sqrt{n_\perp}}\Big)
=
O\!\Big(\frac{1}{n^{3/2}}\Big),
\]
since $k\asymp n_\perp\asymp n$.
Therefore,
\begin{equation}
\Big(\frac{\eta_f}{k}\Big)^2\E_t\!\left[\frac{\|g_t\|_2^2}{n_\perp}\,v_tv_t^\top\right]
=
\Big(\frac{\eta_f}{k}\Big)^2 K_t
+E_t^{(g)},
\qquad
K_t:=\E_t[v_tv_t^\top]=S_t+\E_t[\tilde v_t\tilde v_t^\top].
\label{eq:app-quad-Kt}
\end{equation}
The cross terms in $K_t$ vanish because $\mu_t$ is a function of $z_{0,t}$ and $\E_t[\tilde v_t\mid z_{0,t}]=0$.

\subsection{Linear term: exact Stein identity}
Condition on $(\mathcal F_t,z_{0,t})$. Since the rows of $F_t^\perp$ lie in $\mathrm{range}(P_\perp)$, $h_t^\perp=F_t^\perp g_t\sim\mathcal N(0,\Sigma_t)$ with $\Sigma_t:=F_t^\perp(F_t^\perp)^\top=n_\perp G_t$, and Gaussian integration by parts in $g_t\sim\mathcal N(0,P_\perp)$ \cite{Nourdin2012malliavin}, using $\nabla_{g}v_{i,t}=F_t^{\perp\top}\partial v_{i,t}/\partial h_t^\perp$ (a.e., by \textbf{A3}; valid also when $\Sigma_t$ is singular, e.g.,\ $k>n_\perp$), gives, for all $i,j$,
\[
\E\big[v_{i,t}h^\perp_{j,t}\,\big|\,\mathcal F_t,z_{0,t}\big]=\sum_{\ell=1}^k(\Sigma_t)_{j\ell}\,\E\Big[\frac{\partial v_{i,t}}{\partial h^\perp_{\ell,t}}\,\Big|\,\mathcal F_t,z_{0,t}\Big],
\]
\[
\frac{\partial v_{i,t}}{\partial h^\perp_{\ell,t}}=\frac1k\,a_{i,t}a_{\ell,t}\sigma'(h_{i,t})\sigma'(h_{\ell,t})+\delta_{i\ell}\,e_ta_{i,t}\sigma''(h_{i,t}).
\]
Since $\Sigma_t$ is $\mathcal F_t$-measurable, averaging over $z_{0,t}$ gives $\E_t[v_th_t^{\perp\top}]=J_t\Sigma_t=n_\perp J_tG_t$ exactly, and, by symmetry of $J_t$ and $\Sigma_t$, $\E_t[h_t^\perp v_t^\top]=n_\perp G_tJ_t$. Hence
\begin{equation}
\frac{\eta_f}{k\,n_\perp}\Big(\E_t[v_t h_t^{\perp\top}]+\E_t[h_t^\perp v_t^\top]\Big)=\frac{\eta_f}{k}\big(J_tG_t+G_tJ_t\big).
\label{eq:app-lin-stein}
\end{equation}

\subsection{Assemble terms and conclude}
Combining \eqref{eq:app-gram-start}, \eqref{eq:app-quad-Kt} and \eqref{eq:app-lin-stein} gives \eqref{eq:gram-onestep} with $E_t:=E_t^{(g)}$, so that $\|E_t\|_F=O(n^{-3/2})$ uniformly over $t\le\tau n$ (with probability $1-o(1)$), which is \eqref{eq:gram-onestep-rem}.\qed

\section{Teacher-induced covariance is effectively finite rank}
\label{app:teacher-finite-rank}

This appendix proves that the \emph{teacher-induced} neuron-space covariance
\[
    S_t \;=\; \E_{z_\star}\big[\mu_t(z_\star)\mu_t(z_\star)^\top \,\big|\,\mathcal F_t\big],
    \qquad
    \mu_{i,t}(z_\star):=\E[v_{i,t}\mid \mathcal F_t,z_\star],
\]
is well-approximated (in normalized trace norm, hence in normalized Frobenius norm) by a matrix of rank at most $N_D = \binom{k_\star+D}{D}$,
which grows only polylogarithmically in $n$ for a truncation degree such as $D=\lceil(\log n)^2\rceil$, for which $\log n/(D+1)\to0$,
and that the \emph{accumulated} structure is controlled uniformly over time.

Throughout, $z_\star\sim\mathcal N(0,I_{k_\star})$ with $k_\star=O(1)$ fixed.

\subsection{Hermite basis and notation}

Let $\gamma_{k_\star}$ denote the standard Gaussian measure on $\R^{k_\star}$.
Let $\{H_\alpha\}_{\alpha\in\mathbb N^{k_\star}}$ be the multivariate orthonormal Hermite basis
for $L^2(\gamma_{k_\star})$ \cite{Nourdin2012malliavin}, indexed by multi-indices $\alpha=(\alpha_1,\dots,\alpha_{k_\star})$ with
total degree $|\alpha|=\sum_j \alpha_j$, and satisfying
$\E[H_\alpha(z_\star)H_\beta(z_\star)]=\mathbf 1\{\alpha=\beta\}$.

For each $i$, expand (in $L^2(\gamma_{k_\star})$)
\begin{equation}
    \mu_{i,t}(z_\star)=\sum_{\alpha\in\mathbb N^{k_\star}} \beta_{i,t}^{(\alpha)}\,H_\alpha(z_\star),
    \qquad
    \beta_{i,t}^{(\alpha)}:=\E_{z_\star}[\mu_{i,t}(z_\star)H_\alpha(z_\star)\mid \mathcal F_t].
    \label{eq:app-hermite-expansion}
\end{equation}
Define coefficient vectors $b_t^{(\alpha)}\in\R^k$ by stacking coefficients across neurons:
\begin{equation}
    b_t^{(\alpha)} := \big(\beta_{1,t}^{(\alpha)},\dots,\beta_{k,t}^{(\alpha)}\big)^\top,
    \qquad \alpha\in\mathbb N^{k_\star}.
    \label{eq:app-balpha-def}
\end{equation}

\subsection{Exact PSD decomposition of $S_t$}

\begin{lemma}[Hermite coefficient decomposition]
\label{lem:app-St-decomp}
Conditional on $\mathcal F_t$,
\begin{equation}
    S_t \;=\;\sum_{\alpha\in\mathbb N^{k_\star}} b_t^{(\alpha)}\big(b_t^{(\alpha)}\big)^\top,
    \label{eq:app-St-rank1-sum}
\end{equation}
where the series converges in Frobenius norm.
\end{lemma}

\paragraph{Proof.}
Insert \eqref{eq:app-hermite-expansion} into $\mu_t(z_\star)\mu_t(z_\star)^\top$ and take $\E_{z_\star}[\cdot\mid\mathcal F_t]$.
Orthonormality removes cross-terms, yielding \eqref{eq:app-St-rank1-sum}.
Convergence follows from Parseval:
$\E[\mu_{i,t}(z_\star)^2\mid\mathcal F_t]=\sum_\alpha (\beta_{i,t}^{(\alpha)})^2<\infty$ under bounded moments. \qed

\subsection{Spectral tail control from Lipschitz regularity}

\begin{lemma}[Hermite tail bound via Gaussian Sobolev energy]
\label{lem:app-hermite-tail}
Let $f\in W^{1,2}(\gamma_{k_\star})$ with Hermite coefficients $\{\hat f(\alpha)\}$ in the basis $\{H_\alpha\}$.
Then for any integer $D\ge 0$,
\begin{equation}
    \sum_{|\alpha|>D} \hat f(\alpha)^2
    \;\le\; \frac{1}{D+1}\,\E\big[\|\nabla f(z_\star)\|_2^2\big],
    \qquad z_\star\sim\mathcal N(0,I_{k_\star}),
    \label{eq:app-tail-bound}
\end{equation}
where $\nabla f$ is the weak gradient (defined a.e.).
\end{lemma}

\paragraph{Proof.}
Let $f=\sum_{m\ge 0} f^{(m)}$ be the orthogonal decomposition into homogeneous Wiener chaoses
(total degree $m$). Then $\|f^{(m)}\|_2^2=\sum_{|\alpha|=m}\hat f(\alpha)^2$.
A classical identity for the OU generator \cite{Nourdin2012malliavin} gives
\[
    \E[\|\nabla f\|_2^2] \;=\; \sum_{m\ge 1} m\,\|f^{(m)}\|_2^2.
\]
Therefore,
\[
    \sum_{m>D}\|f^{(m)}\|_2^2
    \;\le\; \frac{1}{D+1}\sum_{m>D} m\,\|f^{(m)}\|_2^2
    \;\le\; \frac{1}{D+1}\,\E[\|\nabla f\|_2^2],
\]
which is \eqref{eq:app-tail-bound} after re-expanding by multi-indices. \qed

\subsection{Uniform Sobolev bound for $\mu_{i,t}(z_\star)$}

\begin{lemma}[Averaged Gaussian Sobolev bound for conditional mean drivers]
\label{lem:app-mu-sobolev}
Assume \textbf{A1}--\textbf{A5}. Fix $\tau>0$ and $t\le \tau n$.
There exists a sequence $G(n)$, with $G^2(n)=O(\log n)$, depending only on $L_g$, $\|g\|_\infty$, the bounds
$\|\sigma\|_\infty,\|\sigma'\|_\infty,\|\sigma''\|_\infty$, the constant $B_\tau$ from Lemma~\ref{lem:stability-sufficient}, and
the polylog rate $B'_\tau(n)=O(\log n)$ from Lemma~\ref{lem:stability-sufficient-4th}, such that with
probability $1-o(1)$,
\begin{equation}
    \frac{1}{k}\sum_{i=1}^k\,\E_{z_\star}\!\left[\|\nabla_{z_\star}\mu_{i,t}(z_\star)\|_2^2 \,\middle|\, \mathcal F_t\right]
    \;\le\; G^2(n),
    \quad\text{i.e.,}\quad
    \sum_{i=1}^k\,\E_{z_\star}\!\left[\|\nabla_{z_\star}\mu_{i,t}\|_2^2 \,\middle|\, \mathcal F_t\right] \le k\,G^2(n).
    \label{eq:app-mu-grad-bound}
\end{equation}
\end{lemma}

\paragraph{Proof.}
Fix $t$ and condition on $\mathcal F_t$. Recall
\[
    v_{i,t}=e_t\,a_{i,t}\,\sigma'(h_{i,t}),
    \qquad
    e_t=\frac1k\sum_{\ell=1}^k a_{\ell,t}\sigma(h_{\ell,t})-g(z_\star),
    \qquad
    h_{\ell,t}=m_{\ell,t}^\top z_\star + r_{\ell,t}z_{\ell,t},
\]
with $z_{\ell,t}$ complement Gaussians independent of $z_\star$ given $\mathcal F_t$.
By \textbf{A2} and Rademacher's theorem, $g$ has a weak gradient with $\|\nabla g\|_2\le L_g$ a.e. Since
$\sigma\in W^{2,\infty}$, the map $z_\star\mapsto e_t$ is a.e.\ differentiable with
\[
    \nabla_{z_\star} e_t
    \;=\;\frac1k\sum_{\ell=1}^k a_{\ell,t}\sigma'(h_{\ell,t})\,m_{\ell,t} \;-\;\nabla g(z_\star),
\]
and $\nabla_{z_\star}\sigma'(h_{i,t}) = \sigma''(h_{i,t})\,m_{i,t}$ a.e.

Commuting derivative and conditional expectation (justified by dominated convergence using bounded
$\sigma, \sigma', \sigma''$),
\[
    \nabla_{z_\star}\mu_{i,t}(z_\star)
    \;=\;\E\!\left[\,a_{i,t}\,\sigma'(h_{i,t})\,\nabla_{z_\star}e_t
              + e_t\,a_{i,t}\,\sigma''(h_{i,t})\,m_{i,t}
        \,\Big|\,\mathcal F_t,z_\star\right].
\]
Applying Jensen and Cauchy--Schwarz to bound $\|\nabla \mu_{i,t}\|_2^2$,
\begin{equation}
    \|\nabla_{z_\star}\mu_{i,t}(z_\star)\|_2^2
    \;\le\;
    2 \|\sigma'\|_\infty^2\,a_{i,t}^2\,\E[\|\nabla e_t\|_2^2 \mid \mathcal F_t, z_\star]
    + 2 \|\sigma''\|_\infty^2\,a_{i,t}^2\,\|m_{i,t}\|_2^2\,\E[e_t^2 \mid \mathcal F_t, z_\star].
    \label{eq:app-mu-grad-pointwise}
\end{equation}
On the high-probability event of Lemma~\ref{lem:stability-sufficient}, $|e_t|\le C$ uniformly, and by
Cauchy--Schwarz on the formula for $\nabla e_t$,
$\|\nabla e_t\|_2^2 \le 2\|\sigma'\|_\infty^2 \frac{1}{k^2}(\sum_\ell a_\ell^2)(\sum_\ell\|m_\ell\|^2) + 2 L_g^2
\le 2\|\sigma'\|_\infty^2 B_\tau^2 + 2 L_g^2$.

Averaging \eqref{eq:app-mu-grad-pointwise} over $z_\star$ and summing over $i$,
\[
    \sum_{i=1}^k \E_{z_\star}\!\big[\|\nabla \mu_{i,t}\|_2^2 \mid \mathcal F_t\big]
    \;\le\;
    C\sum_{i=1}^k a_{i,t}^2 + C\sum_{i=1}^k a_{i,t}^2\|m_{i,t}\|_2^2
    \;\le\;
    C k B_\tau + C k B'_\tau(n),
\]
where the first bound uses Lemma~\ref{lem:stability-sufficient} and the second uses
\eqref{eq:app-stab-4th-coupled} from Lemma~\ref{lem:stability-sufficient-4th}. Setting
$G^2(n) := C(B_\tau + B'_\tau(n)) = O(\log n)$ yields \eqref{eq:app-mu-grad-bound}.\qed

\subsection{Finite-rank approximation of $S_t$}

\begin{theorem}[Teacher-induced covariance is approximately finite-rank (normalized trace)]
\label{thm:app-finite-rank}
Assume \textbf{A1}--\textbf{A5} and work on the event \textbf{A6}; fix $\tau>0$, $t\le \tau n$.
For any integer $D\ge 0$, define the truncated PSD matrix
\begin{equation}
    S_t^{(\le D)} \;:=\; \sum_{|\alpha|\le D} b_t^{(\alpha)}\big(b_t^{(\alpha)}\big)^\top,
    \label{eq:app-St-trunc}
\end{equation}
where $b_t^{(\alpha)}$ are from \eqref{eq:app-balpha-def}.
Then $S_t^{(\le D)}$ has rank at most
\begin{equation}
    N_D \;:=\; \#\{\alpha\in\mathbb N^{k_\star}:|\alpha|\le D\} \;=\; \binom{k_\star+D}{D},
    \label{eq:app-ND}
\end{equation}
and with probability $1-o(1)$,
\begin{equation}
    \frac{1}{k}\,\mathrm{tr}\big(S_t - S_t^{(\le D)}\big)
    \;\le\; \frac{G^2(n)}{D+1},
    \qquad
    G^2(n) = O(\log n),
    \label{eq:app-finite-rank-bound}
\end{equation}
and hence the same bound for $\frac1k\|S_t-S_t^{(\le D)}\|_F$, where $G^2(n)$ is the Sobolev constant from Lemma~\ref{lem:app-mu-sobolev} (which depends on $B_\tau$ and on
the coupled bound $B'_\tau(n) = O(\log n)$ from Lemma~\ref{lem:stability-sufficient-4th}).
\end{theorem}

\paragraph{Proof.}
By Lemma~\ref{lem:app-St-decomp}, $S_t-S_t^{(\le D)}=\sum_{|\alpha|>D} b^{(\alpha)}b^{(\alpha)\top}$ is PSD, hence
$\|S_t-S_t^{(\le D)}\|_F \le \mathrm{tr}(S_t-S_t^{(\le D)})$ (by PSD-ness of $S_t - S_t^{(\le D)}$, which is a sum of rank-1 PSD terms).
Moreover,
\[
\mathrm{tr}(S_t-S_t^{(\le D)})
=\sum_{i=1}^k \sum_{|\alpha|>D} (\beta_{i,t}^{(\alpha)})^2.
\]
Apply Lemma~\ref{lem:app-hermite-tail} to each $\mu_{i,t}$ conditional on $\mathcal F_t$ and sum over $i$:
\[
\mathrm{tr}(S_t-S_t^{(\le D)})
\le \frac{1}{D+1}\sum_{i=1}^k \E_{z_\star}\!\left[\|\nabla\mu_{i,t}(z_\star)\|_2^2 \,\middle|\, \mathcal F_t\right]
\le \frac{kG^2(n)}{D+1}
\]
with high probability by Lemma~\ref{lem:app-mu-sobolev}. Divide by $k$ to obtain \eqref{eq:app-finite-rank-bound}.
\qed

\subsection{A time-uniform truncation and control of the accumulated covariance}

Fix a macroscopic horizon $T_{\max}=\lfloor \tau n\rfloor$.
For each time $s$, recall $S_s=\E_{z_\star}[\mu_s(z_\star)\mu_s(z_\star)^\top\mid\mathcal F_s]$ and let
$S_s^{(\le D)}$ be \eqref{eq:app-St-trunc}.

\begin{lemma}[Uniform-in-time Hermite tail bound]
\label{lem:app-uniform-tail}
Assume \textbf{A1}--\textbf{A5} and work on the event \textbf{A6}; fix $\tau>0$. With probability $1-o(1)$, for all $s\le T_{\max}$ and all integers $D\ge 0$,
\begin{equation}
    \frac{1}{k}\,\mathrm{tr}\!\big(S_s - S_s^{(\le D)}\big)
    \;\le\; \frac{G^2(n)}{D+1}, \qquad G^2(n)=O(\log n),
    \label{eq:app-uniform-trace-tail}
\end{equation}
where $G^2(n)$ is the (slowly-growing) Sobolev constant from Lemma~\ref{lem:app-mu-sobolev}.
Consequently,
\begin{equation}
    \frac{1}{k}\,\big\|S_s - S_s^{(\le D)}\big\|_F
    \;\le\; \frac{G^2(n)}{D+1}.
    \label{eq:app-uniform-frob-tail}
\end{equation}
\end{lemma}

\paragraph{Proof.}
Apply Lemma~\ref{lem:app-hermite-tail} to each $\mu_{i,s}$ conditional on $\mathcal F_s$ and sum over $i$.
Although Lemma~\ref{lem:app-mu-sobolev} is stated at a fixed time, its proof's bound on
$\sum_i\E_{z_\star}[\|\nabla\mu_{i,s}\|_2^2\mid\mathcal F_s]$ depends only on the quantities bounded in \textbf{A6} and on
$\sum_i a_{i,s}^2\|m_{i,s}\|_2^2$, all of which are controlled uniformly in $s\le T_{\max}$ on a single
$1-o(1)$ event by Lemmas~\ref{lem:stability-sufficient}--\ref{lem:stability-sufficient-4th} (which themselves
give $\sup_{s\le\tau n}$ statements). Hence the same Sobolev constant $G^2(n)$ applies to every $s\le T_{\max}$
on this joint event. Finally, use PSD-ness to pass from trace to Frobenius norm. \qed

\begin{theorem}[Accumulated teacher-induced covariance is uniformly approximately finite-mode]
\label{thm:app-accumulated-finite}
Assume \textbf{A1}--\textbf{A5} and work on the event \textbf{A6}; fix $\tau>0$.
Define the accumulated teacher-induced covariance
\[
    \Sigma_t^{\mathrm{teach}} \;:=\; \sum_{s=0}^{t-1} S_s,
    \qquad
    \Sigma_t^{\mathrm{teach},(\le D)} \;:=\; \sum_{s=0}^{t-1} S_s^{(\le D)}.
\]
(We use the superscript $\mathrm{teach}$ to disambiguate from the complement covariance
$\Sigma_t = n_\perp G_t$ used in Appendix~\ref{app:equivalence-proof}.)
Then with probability $1-o(1)$, uniformly for all $1\le t\le T_{\max}$ and all $D\ge 0$,
\begin{equation}
    \frac{1}{kt}\,\mathrm{tr}\big(\Sigma_t^{\mathrm{teach}}-\Sigma_t^{\mathrm{teach},(\le D)}\big)
    \;\le\; \frac{G^2(n)}{D+1}, \qquad G^2(n)=O(\log n),
    \label{eq:app-accumulated-frob}
\end{equation}
and hence the same bound for $\frac{1}{kt}\|\Sigma_t^{\mathrm{teach}}-\Sigma_t^{\mathrm{teach},(\le D)}\|_F$.
Moreover, the truncation uses a time-invariant teacher feature map of dimension
$R=N_D=\binom{k_\star+D}{D}$.
\end{theorem}

\paragraph{Proof.}
By triangle inequality and Lemma~\ref{lem:app-uniform-tail},
\[
    \big\|\Sigma_t^{\mathrm{teach}}-\Sigma_t^{\mathrm{teach},(\le D)}\big\|_F
    \;\le\;\sum_{s=0}^{t-1}\big\|S_s-S_s^{(\le D)}\big\|_F
    \;\le\;\sum_{s=0}^{t-1}\mathrm{tr}(S_s-S_s^{(\le D)})
    \;\le\; t\,k\,\frac{G^2(n)}{D+1},
\]
and since the second sum equals $\mathrm{tr}(\Sigma_t^{\mathrm{teach}}-\Sigma_t^{\mathrm{teach},(\le D)})$, division by $kt$ yields \eqref{eq:app-accumulated-frob}. The “time-invariant feature map” claim follows from the
fact that the truncation uses the fixed subspace $\mathcal V_D=\mathrm{span}\{H_\alpha:|\alpha|\le D\}$, which depends
only on the (static) teacher variable distribution $z_\star\sim\mathcal N(0,I_{k_\star})$. \qed

\subsection{Time-invariant structure source}

Define the fixed feature map $\varphi_D:\R^{k_\star}\to\R^{R}$ by listing all Hermite polynomials up to degree $D$:
\[
    \varphi_D(z_\star) \;:=\; \big(H_\alpha(z_\star)\big)_{|\alpha|\le D},
    \qquad R=N_D.
\]
For each time $s$, define the loading matrix $B_s\in\R^{k\times R}$ whose columns are the coefficient vectors
$b_s^{(\alpha)}$ for $|\alpha|\le D$:
\begin{equation}
    \mu_s^{(\le D)}(z_\star)
    \;=\;\sum_{|\alpha|\le D} b_s^{(\alpha)}\,H_\alpha(z_\star)
    \;=\; B_s\,\varphi_D(z_\star),
    \label{eq:app-fixed-source-factorization}
\end{equation}
so that
\begin{equation}
    S_s^{(\le D)}
    \;=\;\E_{z_\star}\!\big[\mu_s^{(\le D)}(z_\star)\mu_s^{(\le D)}(z_\star)^\top\mid\mathcal F_s\big]
    \;=\; B_s B_s^\top,
    \label{eq:app-StD-BBT}
\end{equation}
because $\E[\varphi_D(z_\star)\varphi_D(z_\star)^\top]=I_R$ by orthonormality.

\paragraph{Accumulated loading matrix.}
Iterating the mean Gram recursion of Lemma~\ref{lem:gram-onestep}, treating $J_q$ and $K_s$ as deterministic, gives, up to the propagated initial Gram and the second-order terms $(\eta_f/k)^2J_qG_qJ_q$,
$G_t \approx (\eta_f/k)^2 \sum_{s<t}\Pi_{s,t}\,K_s\,\Pi_{s,t}^\top$ with damping product
$\Pi_{s,t}:=(I-\tfrac{\eta_f}{k}J_{t-1})\cdots(I-\tfrac{\eta_f}{k}J_{s+1})$ (empty product $=I$; the contribution injected at step $s$ is
propagated through the damping at steps $s+1,\dots,t-1$). Splitting $K_s=S_s+\E_s[\tilde v_s\tilde v_s^\top]$ and
truncating $S_s\approx S_s^{(\le D)} = B_s B_s^\top$, we define the \emph{propagated accumulated loading matrix}
at the covariance scale $\Sigma_t = n_\perp G_t$:
\begin{equation}
    B_t^{\mathrm{acc}}
    \;:=\;
    \frac{\eta_f\sqrt{n_\perp}}{k}\big[\Pi_{0,t}B_0\,\big|\,\Pi_{1,t}B_1\,\big|\,\cdots\,\big|\,\Pi_{t-1,t}B_{t-1}\big]
    \;\in\;\R^{k\times tR},
    \label{eq:app-Bacc-def}
\end{equation}
so that
$B_t^{\mathrm{acc}}(B_t^{\mathrm{acc}})^\top = n_\perp(\eta_f/k)^2 \sum_{s<t}\Pi_{s,t}\,S_s^{(\le D)}\,\Pi_{s,t}^\top$
captures the propagated teacher-induced contribution to $\Sigma_t$ (at the same scale used in the bulk-plus-spike
approximation of the main text and the variance-inflation channel \eqref{eq:yhat-eff}). The column count $tR$ may be
of order $n$, so $B_t^{\mathrm{acc}}$ is generally not low-rank in neuron space, but its columns are indexed by
the fixed Hermite feature modes $\{H_\alpha:|\alpha|\le D\}$. The construction also assigns the propagated initial Gram and the propagated centered part to the bulk $\hat c_tI_k$, so it gives a qualitative picture of $\Sigma_t$.

\begin{remark}[What ``accumulated finite mode'' means here]
Theorem~\ref{thm:app-accumulated-finite} shows that, uniformly over macroscopic training time,
the accumulated teacher-induced covariance per unit time is well-approximated (in normalized trace norm)
by the sum of the same fixed-mode truncations:
\[
    \frac{1}{t}\sum_{s=0}^{t-1} S_s \approx \frac{1}{t}\sum_{s=0}^{t-1} B_s B_s^\top,
    \qquad \text{with a fixed structure source } \varphi_D(z_\star)\in\R^R.
\]
The structure source $\varphi_D$ is determined only by the static teacher variable $z_\star$ and does not rotate with time.
However, while each per-step factor $S_s^{(\le D)}=B_s B_s^\top$ has rank $\le R$, the accumulated covariance
$\sum_{s<t}B_s B_s^\top$ in $\R^{k\times k}$ may have rank up to $\min(tR, k)$, since the neuron-side column
spaces of $B_s$ vary with $s$. The ``finite-mode'' guarantee is therefore on the teacher side ($\varphi_D\in\R^R$
fixed), not on the neuron-side rank of the accumulated covariance. The accumulated loading matrix $B_t^{\mathrm{acc}}$
defined in \eqref{eq:app-Bacc-def} is used in the variance-inflation analysis of
Proposition~\ref{prop:variance-inflation} (Appendix~\ref{app:variance-inflation}).
\end{remark}

\section{Proof of Theorem~\ref{thm:equivalence-spike} (asymptotic equivalence)}
\label{app:equivalence-proof}

The proof has four steps:
\begin{enumerate}[topsep=0pt,itemsep=-1ex,partopsep=1ex,parsep=1ex]
    \item Identify the test loss as a functional of the state $(a_t, M_t, \Sigma_t)$ with
    $\Sigma_t = n_\perp G_t = F_t^\perp(F_t^\perp)^\top$ (Sec.~\ref{app:eq-step1}).
    \item Match drift and diffusion of the parameters $(a_t, M_t)$ between SGD and DD
    (Sec.~\ref{app:eq-step2}).
    \item Compute the conditional mean of the complement increment by Stein's identity, show that the
    directional decoherence $\rho^{\mathrm{dec}}$ is produced by $\zeta_t$, and derive the spike covariance $\Gamma_t$ (Sec.~\ref{app:eq-step3}).
    \item Prove stability estimates for the reduced state (Lemma~\ref{lem:S}) and transport the matching of conditional mean increments to the test loss by a
    discrete Gr\"onwall argument on the reduced state (Sec.~\ref{app:eq-step4}).
\end{enumerate}

\subsection{Step 1: Test loss as a functional of the covariance state}
\label{app:eq-step1}

The test loss $\mathcal L_t = \frac12\E_{x,y}[(\hat y(x; F_t, a_t)-y)^2]$ depends on $F_t$ only through the
joint distribution of the preactivations $h_{i,t} = m_{i,t}^\top z_\star + h_{i,t}^\perp$. Since
$z_\star = \Xi^\top x \sim\mathcal N(0,I_{k_\star})$ and $h_t^\perp \mid \mathcal F_t \sim \mathcal N(0, \Sigma_t)$
\emph{exactly}, with covariance state
\begin{equation}
    \Sigma_t \;:=\; n_\perp G_t \;=\; F_t^\perp(F_t^\perp)^\top \;\in\;\R^{k\times k},
    \label{eq:cov-state}
\end{equation}
the test loss is a deterministic functional
\begin{equation}
    \mathcal L_t \;=\; \Phi(a_t, M_t, \Sigma_t)
    \;:=\;
    \tfrac12\,\E_{z_\star, h^\perp}\!\Big[\Big(\tfrac1k\textstyle\sum_{i=1}^k a_{i,t}\,\sigma(m_{i,t}^\top z_\star + h_i^\perp) - g(z_\star)\Big)^2\Big],
    \label{eq:test-loss-functional}
\end{equation}
where $h^\perp \sim \mathcal N(0, \Sigma_t)$ is jointly Gaussian. Step~4 uses that $\Phi$ is Lipschitz in the norm $\|\cdot\|_\star$ on the region $\mathcal R_n$ defined there. Moreover, given $\mathcal F_t$ the SGD step depends on the fresh sample only through $(z_{\star,t},h_t^\perp,\|g_t\|_2^2)$, and $\|g_t\|_2^2=h_t^{\perp\top}\Sigma_t^{+}h_t^\perp+\chi_t$ with $\chi_t\sim\chi^2_{n_\perp-\mathrm{rank}(\Sigma_t)}$ independent of $(z_{\star,t},h_t^\perp)$. Hence the conditional law of $X_{t+1}:=(a_{t+1},M_{t+1},\Sigma_{t+1})$ depends on $F_t^\perp$ only through $\Sigma_t$, and $(X_t)$ is a Markov chain; the same holds for Decomposed Dynamics, whose complement rows stay in $\mathrm{range}(P_\perp)$ by \eqref{eq:dd-complement-update} and whose noise $\zeta_t$ is rotation invariant within $\mathrm{range}(P_\perp)$. (The structural decomposition $\Sigma_t\approx \hat c_t I + B_t^{\mathrm{acc}}(B_t^{\mathrm{acc}})^\top$
of Appendix~\ref{app:teacher-finite-rank} is not used in the equivalence proof; it serves as the interpretive
framework in the variance-inflation analysis of Proposition~\ref{prop:variance-inflation}.)

\subsection{Step 2: drift+diffusion of $(a_t, M_t)$ match between SGD and DD}
\label{app:eq-step2}

By construction (Theorem~\ref{thm:equivalence-spike} item 2(i)), Decomposed Dynamics applies the SGD updates
\eqref{eq:sgd-update-f}--\eqref{eq:sgd-update-a} verbatim to $(a_t^{\mathrm{DD}}, M_t^{\mathrm{DD}})$:
\begin{align*}
    a_{i,t+1}^{\mathrm{DD}} &= a_{i,t}^{\mathrm{DD}} - \tfrac{\eta_a}{k}\big(e_t^{\mathrm{DD}}\,\sigma(h_{i,t}^{\mathrm{DD}}) + \lambda a_{i,t}^{\mathrm{DD}}\big),\\
    m_{i,t+1}^{\mathrm{DD}} &= m_{i,t}^{\mathrm{DD}} - \tfrac{\eta_f}{k}\,e_t^{\mathrm{DD}}\,a_{i,t}^{\mathrm{DD}}\,\sigma'(h_{i,t}^{\mathrm{DD}})\,z_{\star,t},
\end{align*}
with $h_{i,t}^{\mathrm{DD}} = (m_{i,t}^{\mathrm{DD}})^\top z_\star + h_{i,t}^{\perp,\mathrm{DD}}$ and
$h_t^{\perp,\mathrm{DD}} \mid \mathcal F_t \sim \mathcal N(0, \Sigma_t^{\mathrm{DD}})$ where
$\Sigma_t^{\mathrm{DD}} := F_t^{\perp,\mathrm{DD}}(F_t^{\perp,\mathrm{DD}})^\top$. Both channels are exact
Gaussians; equivalence reduces to comparing the evolutions of $\Sigma_t$ and $\Sigma_t^{\mathrm{DD}}$ (Steps~3--4).

The conditional laws of the $(a_{i,t},m_{i,t})$ increments under SGD and under DD agree whenever the complete
reduced states $(a_t,M_t,\Sigma_t)$ and $(a_t^{\mathrm{DD}},M_t^{\mathrm{DD}},\Sigma_t^{\mathrm{DD}})$ agree (Step~1); by Lemma~\ref{lem:S}~(S1), on the region $\mathcal R_n$ of Step~4 the
mismatch of the conditional means of these increments is controlled by the $\|\cdot\|_\star$-distance of the states. Step~3 below shows that DD's
complement update has the same conditional mean increment as SGD's; together with the matching here, this is the input to Step~4. 

\subsection{Step 3: complement drift and spike covariance}
\label{app:eq-step3}

Write the SGD complement matrix as $F^{\perp}_t=F_tP_\perp$ and recall the exact rank-one recursion
\[
\Delta F^{\perp}_t := F^{\perp}_{t+1}-F^{\perp}_t \;=\; -\frac{\eta_f}{k}\, v_t g_t^\top,
\qquad g_t=P_\perp x_t\sim\mathcal N(0,P_\perp),
\]
where $v_{i,t}=e_t a_{i,t}\sigma'(h_{i,t})$ and $h_{i,t}=m_{i,t}^\top z_{\star,t} + (f^\perp_{i,t})^\top g_t$.

\paragraph{(i) Conditional mean: the friction term and the full drift.}
Fix a neuron $i$ and condition on $\mathcal F_t$. Let $f^\perp_{i,t}\in\R^n$ be the $i$-th row of $F^{\perp}_{t}$ and $r_{i,t}^2=\|f^\perp_{i,t}\|_2^2$.
Define the complement preactivation (scalar)
\[
h^{\perp}_{i,t} := (f^\perp_{i,t})^\top g_t.
\]
By Gaussian regression in $\mathrm{range}(P_\perp)$, we may decompose
\begin{equation}
g_t \;=\; \frac{h^{\perp}_{i,t}}{r_{i,t}^2}\,f^\perp_{i,t} \;+\; g^{(i)}_{\perp,t},
\qquad
g^{(i)}_{\perp,t}\ \perp\ f^\perp_{i,t},
\qquad
\E\!\left[g^{(i)}_{\perp,t}g^{(i)\top}_{\perp,t}\,\middle|\,\mathcal F_t\right]
=
P_\perp-\frac{f^\perp_{i,t}(f^\perp_{i,t})^\top}{r_{i,t}^2}.
\label{eq:regression-decomp}
\end{equation}
Substituting into the row update $f^\perp_{i,t+1}=f^\perp_{i,t}-(\eta_f/k)v_{i,t}g_t$ gives the exact identity
\begin{equation}
f^\perp_{i,t+1}
=
\Big(1-\frac{\eta_f}{k}\frac{v_{i,t}h^{\perp}_{i,t}}{r_{i,t}^2}\Big)\,f^\perp_{i,t}
-\frac{\eta_f}{k}\,v_{i,t}\,g^{(i)}_{\perp,t}.
\label{eq:row-update-decomp}
\end{equation}
The second term $-(\eta_f/k) v_{i,t}\,g^{(i)}_{\perp,t}$ is orthogonal in $\mathrm{range}(P_\perp)$ to
$f^\perp_{i,t}$ \emph{by construction} (since $g^{(i)}_{\perp,t}\perp f^\perp_{i,t}$ from the regression
decomposition \eqref{eq:regression-decomp}). Its conditional mean
$\E[v_{i,t}\,g^{(i)}_{\perp,t} \mid \mathcal F_t]$ is generally nonzero — by Stein's identity it equals
$\Sigma'_t\,\E[\nabla_{g^{(i)}_\perp} v_{i,t} \mid \mathcal F_t]$ where
$\Sigma'_t = P_\perp - f^\perp_{i,t}(f^\perp_{i,t})^\top/r_{i,t}^2$ — but it lives in
$\mathrm{range}(\Sigma'_t)$, the orthogonal complement of $f^\perp_{i,t}$ within $\mathrm{range}(P_\perp)$. The
projection onto $f^\perp_{i,t}$ is therefore zero, but the orthogonal mean is first order in $\eta_f/k$ and
coherent across rows; it is part of the full drift computed below.

Taking $\E[\cdot\mid\mathcal F_t]$ in \eqref{eq:row-update-decomp} and projecting onto the direction
$f^\perp_{i,t}/r_{i,t}$ yields the scalar friction identity
\[
\E[(f^\perp_{i,t+1})^\top u_{i,t}\mid \mathcal F_t]
=
\big(1-\rho^{\mathrm{fric}}_{i,t}\big)\,r_{i,t},
\qquad
\rho^{\mathrm{fric}}_{i,t}
=
\frac{\eta_f}{k}\frac{\E[v_{i,t}h^{\perp}_{i,t}\mid \mathcal F_t]}{r_{i,t}^2},
\]
which is the projection on $f^\perp_{i,t}$ of the full drift computed next.

\smallskip\emph{Full conditional mean.}
Summing the $f^\perp_{i,t}$-aligned and orthogonal parts, Gaussian integration by parts over $g_t\sim\mathcal N(0,P_\perp)$ given $(\mathcal F_t,z_{\star,t})$ gives $\E[v_{i,t}g_t\mid\mathcal F_t,z_{\star,t}]=\sum_{\ell}\E[\partial v_{i,t}/\partial h^\perp_{\ell,t}\mid\mathcal F_t,z_{\star,t}]\,f^\perp_{\ell,t}$, hence, after averaging over $z_{\star,t}$,
\[
\E_t[\Delta F_t^\perp]=-\frac{\eta_f}{k}J_tF_t^\perp=-\widetilde\Lambda_tF_t^\perp,
\]
with $J_t$ of Lemma~\ref{lem:gram-onestep}. Its projection on $f^\perp_{i,t}$ is the friction term above; its off-diagonal entries couple the rows. Since $\E_t[\zeta_t]=0$, \eqref{eq:dd-complement-update} has exactly this conditional mean increment.

\paragraph{(ii) Directional decoherence.}
The orthogonal perturbation in \eqref{eq:row-update-decomp} rotates the direction of $f^\perp_{i,t}$ within
$\mathrm{range}(P_\perp)$. We compute the rate of this decoherence and show that Decomposed Dynamics
reproduces it through $\zeta_t$.
Let $u_{i,t}=f^\perp_{i,t}/r_{i,t}$ and define the orthogonal increment
\[
\delta^{\perp}_{i,t}:=-\frac{\eta_f}{k}\,v_{i,t}\,g^{(i)}_{\perp,t},
\qquad u_{i,t}^\top \delta^{\perp}_{i,t}=0.
\]
A second-order Taylor expansion of the normalization map $x\mapsto x/\|x\|_2$ yields, for neurons with $r_{i,t}$ bounded below and $|a_{i,t}|\le A_n$ (event \eqref{eq:app-stab-Esup-a}),
\begin{equation}
\E\!\left[u_{i,t+1}^\top u_{i,t}\,\middle|\,\mathcal F_t\right]
=
1-\frac{1}{2r_{i,t}^2}\,\E\!\left[\|\delta^{\perp}_{i,t}\|_2^2\,\middle|\,\mathcal F_t\right]
+o\!\Big(\frac{1}{n}\Big).
\label{eq:angle-decay-taylor}
\end{equation}
We use $\E\big[\|g^{(i)}_{\perp,t}\|_2^2 \mid\mathcal F_t\big] = \mathrm{Tr}(P_\perp - f_i^\perp(f_i^\perp)^\top/r_i^2)
= n_\perp - 1$ and write $\E_t[v_{i,t}^2 \|g^{(i)}_\perp\|^2]
= \E_t[v_{i,t}^2]\,(n_\perp-1) + \mathrm{Cov}_t(v_{i,t}^2, \|g^{(i)}_\perp\|^2)$. The covariance is bounded by
Cauchy--Schwarz: $|\mathrm{Cov}_t(v_i^2, \|g^{(i)}_\perp\|^2)| \le \sqrt{\mathrm{Var}_t(v_i^2)}\sqrt{\mathrm{Var}_t(\|g^{(i)}_\perp\|^2)}$,
with $\mathrm{Var}_t(\|g^{(i)}_\perp\|^2) = 2(n_\perp-1)$ (chi-square) and
$\mathrm{Var}_t(v_i^2) \le \E_t[v_i^4] \le C a_{i,t}^4$ pointwise on the high-probability event of
Lemma~\ref{lem:stability-sufficient} (since $|v_i|\le C|a_i|$ from $|e_t|,\|\sigma'\|_\infty\le C$ on
the same event). Hence
\[
\E\!\left[\|\delta^{\perp}_{i,t}\|_2^2\,\middle|\,\mathcal F_t\right]
\;=\;
\Big(\frac{\eta_f}{k}\Big)^2 \Big[\,\E[v_{i,t}^2\mid\mathcal F_t]\,(n_\perp-1) + O\big(a_{i,t}^2\sqrt{n_\perp}\big)\,\Big],
\]
where, for such neurons, the $O$-term contributes $O(n^{-3/2}\log n)=o(1/n)$ to \eqref{eq:angle-decay-taylor}.
Inserting into \eqref{eq:angle-decay-taylor} and using $n_\perp-1=n_\perp+o(n)$ yields the leading-order decay
\[
\E[u_{i,t+1}^\top u_{i,t}\mid\mathcal F_t]
=
1-\rho^{\mathrm{dec}}_{i,t}+o\!\Big(\frac{1}{n}\Big),
\qquad
\rho^{\mathrm{dec}}_{i,t}
=
\frac{n_\perp}{2}\Big(\frac{\eta_f}{k}\Big)^2\frac{\E[v_{i,t}^2\mid\mathcal F_t]}{r_{i,t}^2}.
\]
For Decomposed Dynamics, row $i$ of $\zeta_t$ has $\E_t\|\zeta_{i,t}\|_2^2=n_\perp(\eta_f/k)^2\E_t[v_{i,t}^2]$, and the drift $-\widetilde\Lambda_tF_t^\perp$ changes $u_{i,t}^\top u_{i,t+1}$ only at order $o(1/n)$, so \eqref{eq:dd-complement-update} gives the same $\E_t[u_{i,t+1}^\top u_{i,t}]=1-\rho^{\mathrm{dec}}_{i,t}+o(1/n)$.

\paragraph{(iii) Spike covariance.}
The SGD complement increment satisfies
\[
\Delta F^{\perp}_t\Delta (F^{\perp}_t)^\top
=
\Big(\frac{\eta_f}{k}\Big)^2 v_t v_t^\top\,\|g_t\|_2^2.
\]
Writing $\|g_t\|_2^2 = n_\perp + (\|g_t\|_2^2 - n_\perp)$ with the second term having
$\E_t[\,\cdot\,] = 0$ and $\mathrm{Var}_t(\,\cdot\,) = 2 n_\perp$ (chi-square), Cauchy--Schwarz on the
covariance gives, entry-wise,
\[
\E_t\!\big[(v_i v_j)\,\|g_t\|_2^2\big]
= n_\perp\,\E_t[v_i v_j] + O\!\big(\sqrt{n_\perp}\cdot\sqrt{\E_t[v_i^2 v_j^2]}\big)
= n_\perp\,(K_t)_{ij} + O\!\big(|a_i a_j|\sqrt{n_\perp}\big),
\]
hence in Frobenius norm,
\[
\E[\Delta F^{\perp}_t\Delta (F^{\perp}_t)^\top\mid\mathcal F_t]
=
n_\perp\,\Gamma_t+E_t^{\mathrm{cov}},\qquad
\|E_t^{\mathrm{cov}}\|_F\le C(\eta_f/k)^2\sqrt{n_\perp}\,\|a_t\|_2^2=O(n^{-1/2}),
\]
with $E_t^{\mathrm{cov}}=n_\perp E_t$ for $E_t$ of Lemma~\ref{lem:gram-onestep}, and $\Gamma_t = (\eta_f/k)^2 K_t$ as defined in \eqref{eq:gamma-lambdatotal-def}. In the surrogate,
$\mathrm{vec}(\zeta_t^\top)\sim\mathcal N(0,\Gamma_t\otimes P_\perp)$ ensures $\E[\zeta_t\zeta_t^\top\mid\mathcal F_t]
=\mathrm{Tr}(P_\perp)\Gamma_t = n_\perp\,\Gamma_t$, matching the leading-order covariance of the SGD increment.

\paragraph{(iv) Summary of Step 3.}
The update \eqref{eq:dd-complement-update} has SGD's conditional mean increment $-\widetilde\Lambda_tF_t^\perp$ (i) and its leading neuron-space second moment $n_\perp\Gamma_t$ (iii); the directional decoherence (ii) is produced by $\zeta_t$. Higher moments are not matched: SGD's linear Gram increment $-\frac{\eta_f}{k}(v_th_t^{\perp\top}+h_t^\perp v_t^\top)$ correlates $v_t$ with the same $h_t^\perp$, whereas, given $\mathcal F_t$, $\zeta_t$ is a centered Gaussian independent of the fresh sample, so the conditional variances of projections of $\Delta\Sigma_t$ differ.

\subsection{Step 4: Gr\"onwall argument for the test loss}
\label{app:eq-step4}

For either procedure write $X_t=(a_t,M_t,\Sigma_t)$, a Markov chain with $\mathcal L_t=\Phi(X_t)$ (Step~1). For $x=(a,M,\Sigma)$ let
$\|x\|_\star:=k^{-1/2}\big(\|a\|_2+\|M\|_F+\|\mathrm{diag}(\Sigma)\|_2\big)+k^{-1}\|\Sigma\|_2$, where $\|\mathrm{diag}(\Sigma)\|_2:=(\sum_i\Sigma_{ii}^2)^{1/2}$, let
$\mathcal R_n:=\{x:\Sigma\succeq0,\ \max_i|a_i|\le\bar A_\tau,\ \max_i(\|m_i\|_2^2+\Sigma_{ii})\le\bar Q_\tau\}$ with $\bar A_\tau:=e^{2C_a\tau/\alpha}(A+1)$ for $C_a:=\eta_a\|\sigma\|_\infty\max(\|\sigma\|_\infty,\|g\|_\infty)$ and $A$ the bound on $|a_{i,0}|$ in Theorem~\ref{thm:equivalence-spike} (for $n$ large, $\bar A_\tau$ bounds $\max_i|a_{i,t}|$; proof of Lemma~\ref{lem:S}) and $\bar Q_\tau$ from Lemma~\ref{lem:S}, and let $b(x)$ be the conditional mean increment of Decomposed Dynamics at state $x$.

\begin{lemma}[Stability of the reduced dynamics]
\label{lem:S}
Assume \textbf{A1}--\textbf{A5} and $|a_{i,0}|\le A$ almost surely, and let Decomposed Dynamics update $(a_t,M_t)$ verbatim by the SGD rule, as in Step~2. There are $\bar Q_\tau,L,L_\Phi<\infty$, independent of $n$, such that, for all $n$ large enough:
\textbf{(S1)} $\|b(x)-b(y)\|_\star\le\frac{L}{n}\|x-y\|_\star$ and $|\Phi(x)-\Phi(y)|\le L_\Phi\|x-y\|_\star$ for all $x,y\in\mathcal R_n$;
\textbf{(S2)} with probability $1-o(1)$, $X_t^{\mathrm{SGD}}\in\mathcal R_n$ and $X_t^{\mathrm{DD}}\in\mathcal R_n$ for all $t\le\tau n$;
\textbf{(S3)} for each procedure, the martingale part $\mathcal M_t:=\sum_{s<t}\big(X_{s+1}-X_s-\E[X_{s+1}-X_s\mid X_s]\big)$ satisfies $\sup_{t\le\tau n}\|\mathcal M_t\|_\star=O_{\P}(n^{-1/2})$.
\end{lemma}

\begin{proof}
Write $s_j:=\|\sigma^{(j)}\|_\infty$, finite for $j\le3$ by \textbf{A3}; below, $C$ denotes constants that depend only on $\tau$, $\alpha$, $A$, $\eta_a$, $\eta_f$, $\lambda$, $k_\star$, $\|g\|_\infty$ and $s_0,\dots,s_3$, and $n$ is large enough that $n/k\le2/\alpha$ and $\eta_a\lambda\le k$.

\emph{Drift map.} Fix $x\in\mathcal R_n$, let $z\sim\mathcal N(0,I_{k_\star})$ and $h^\perp\sim\mathcal N(0,\Sigma)$ be independent, and set $h:=Mz+h^\perp$, $e:=\frac1ka^\top\sigma(h)-g(z)$, $w:=a\circ\sigma'(h)$, $v:=e\,w$ and $D_v:=\partial v/\partial h=\frac1kww^\top+\mathrm{diag}(\gamma)$, $\gamma:=e\,a\circ\sigma''(h)$. Then $J=\E D_v$, $K=\E[vv^\top]$ and, by Stein's identity, $J\Sigma=\E[vh^{\perp\top}]$. With $c_a:=\eta_an/k$ and $c_f:=\eta_fn/k$,
\begin{equation}
nb(x)=\Big(-c_a\big(\E[e\,\sigma(h)]+\lambda a\big),\ -c_f\E[vz^\top],\ -c_f(J\Sigma+\Sigma J)+c_f\tfrac{\eta_f}{k}J\Sigma J+c_f^2\tfrac{n_\perp}{n}K\Big).
\label{eq:S-drift}
\end{equation}
On $\mathcal R_n$, pointwise in $(z,h^\perp)$: $|e|\le C_e:=s_0\bar A_\tau+\|g\|_\infty$, $|w_i|\le s_1\bar A_\tau$, $|v_i|\le C_v:=C_es_1\bar A_\tau$ and $\|D_v\|_2\le C_D:=s_1^2\bar A_\tau^2+C_es_2\bar A_\tau$, so $\|J\|_2\le C_D$; moreover $\|\Sigma\|_2\le\Tr\Sigma\le k\bar Q_\tau$.

\emph{(S1).} $\mathcal R_n$ is convex. Mollifying $\sigma$ does not increase any $s_j$, and $b,\Phi$ depend continuously on $(\sigma,\sigma',\sigma'')$, so we may assume $\sigma\in C^\infty$. Let $x,y\in\mathcal R_n$, $\delta=(\delta a,\delta M,\delta\Sigma):=y-x$ and $x_\theta:=x+\theta\delta$. Gaussian interpolation \cite[Lemma 1.3.1]{talagrand2010mean} gives $\frac{d}{d\theta}\E_{x_\theta}[F]=\E[\partial_{(a,M)}F[\delta a,\delta M]]+\frac12\E\langle\nabla^2_{h^\perp}F,\delta\Sigma\rangle$ for smooth $F(a,M;z,h^\perp)$ of polynomial growth. Applying it to $u_i$, $v_iz$, $v_ih_j^\perp$ and $v_iv_j$, with $\nabla^2_{h^\perp}(v_ih_j^\perp)=h_j^\perp\nabla^2v_i+\nabla v_i\mathbf e_j^\top+\mathbf e_j\nabla v_i^\top$ ($\mathbf e_j$ the $j$-th standard basis vector), gives for $u\in\{e\,\sigma(h),v\}$ and the random vector $\hat\delta u:=\partial_{(a,M)}u[\delta a,\delta M]+\frac12\big(\langle\nabla_h^2u_i,\delta\Sigma\rangle\big)_{i\le k}$
\begin{equation}
\begin{aligned}
&\tfrac{d}{d\theta}\E u=\E\hat\delta u,\qquad
\tfrac{d}{d\theta}\E[vz^\top]=\E[\hat\delta v\,z^\top],\qquad
\tfrac{d}{d\theta}(J\Sigma)=\E[\hat\delta v\,h^{\perp\top}]+J\delta\Sigma,\\
&\tfrac{d}{d\theta}K=\E[\hat\delta v\,v^\top+v\,\hat\delta v^\top]+\E[D_v\,\delta\Sigma\,D_v].
\end{aligned}
\label{eq:S-deriv}
\end{equation}
Explicitly, with $\delta e:=\frac1k\big(\sigma(h)^\top\delta a+w^\top\delta M z\big)$,
\begin{align*}
\hat\delta v_i={}&\delta e\,w_i+e\sigma'(h_i)\delta a_i+e\,a_i\sigma''(h_i)\,\delta m_i^\top z\\
&+\tfrac1{2k}w_i\langle a\circ\sigma''(h),\mathrm{diag}\,\delta\Sigma\rangle+\tfrac1ka_i\sigma''(h_i)(\delta\Sigma\,w)_i+\tfrac12e\,a_i\sigma'''(h_i)\,\delta\Sigma_{ii},
\end{align*}
and $\hat\delta(e\,\sigma)_i$ has the same terms with $a_i\sigma^{(j)}(h_i)$ replaced by $\sigma^{(j-1)}(h_i)$ and without the $\delta a_i$ term; $\sigma'''$ enters only through the last term of $\hat\delta v_i$. Using $k\,\delta e^2\le2s_0^2\|\delta a\|_2^2+2s_1^2\bar A_\tau^2\|\delta Mz\|_2^2$, $\|w\|_2^2\le ks_1^2\bar A_\tau^2$, $\langle a\circ\sigma''(h),\mathrm{diag}\,\delta\Sigma\rangle^2\le ks_2^2\bar A_\tau^2\|\mathrm{diag}\,\delta\Sigma\|_2^2$, $\|\delta\Sigma\,w\|_2\le\|\delta\Sigma\|_2\|w\|_2$ and $\E\|\delta Mz\|_2^2=\|\delta M\|_F^2$, we obtain
\begin{equation}
\E\|\hat\delta u\|_2^2\le C\big(\|\delta a\|_2^2+\|\delta M\|_F^2+\|\mathrm{diag}\,\delta\Sigma\|_2^2+k^{-1}\|\delta\Sigma\|_2^2\big)\le Ck\|\delta\|_\star^2 .
\label{eq:S-key}
\end{equation}
For random $\psi\in\R^k$ and $\varphi\in\R^p$, Cauchy--Schwarz gives $\|\E[\psi\varphi^\top]\|_2\le(\E\|\psi\|_2^2)^{1/2}\|\E[\varphi\varphi^\top]\|_2^{1/2}$ and, if $p=k$, $\|\mathrm{diag}\,\E[\psi\varphi^\top]\|_2\le(\E\|\psi\|_2^2)^{1/2}\max_i(\E\varphi_i^2)^{1/2}$. Here $\|\E[\varphi\varphi^\top]\|_2=1$ for $\varphi\in\{1,z\}$, while $\|\E[\varphi\varphi^\top]\|_2\le k\max(\bar Q_\tau,C_v^2)$ and $\max_i\E\varphi_i^2\le\max(\bar Q_\tau,C_v^2)$ for $\varphi\in\{h^\perp,v\}$. With $\|\cdot\|_F\le\sqrt{k_\star}\|\cdot\|_2$ on $\R^{k\times k_\star}$, \eqref{eq:S-key} therefore bounds every expectation of the form $\E[\hat\delta u\,\varphi^\top]$ in \eqref{eq:S-deriv} by $C\|\delta\|_\star$ in each part of $\|\cdot\|_\star$ ($k^{-1/2}$ times the Euclidean, Frobenius or diagonal norm, and $k^{-1}$ times the operator norm). The diagonal bound with $\varphi=h^\perp$ is the only place in (S1) where the per-neuron bound $\Sigma_{ii}\le\bar Q_\tau$, rather than its average, is used. For the remaining terms, $\|J\delta\Sigma\|_2\le C_D\|\delta\Sigma\|_2$ and $\|\E[D_v\delta\Sigma D_v]\|_2\le C_D^2\|\delta\Sigma\|_2$, while
$(J\delta\Sigma)_{ii}=\frac1k\E[w_i(\delta\Sigma w)_i]+\E[\gamma_i]\,\delta\Sigma_{ii}$ and $(D_v\delta\Sigma D_v)_{ii}=\frac{w_i^2}{k^2}w^\top\delta\Sigma w+\frac2kw_i\gamma_i(\delta\Sigma w)_i+\gamma_i^2\,\delta\Sigma_{ii}$ give diagonals of Euclidean norm at most $C(\|\mathrm{diag}\,\delta\Sigma\|_2+k^{-1/2}\|\delta\Sigma\|_2)$, by $|w_i|\le s_1\bar A_\tau$ and $\|w\|_2\le\sqrt k\,s_1\bar A_\tau$. Integrating over $\theta\in[0,1]$, the changes of $\E[e\,\sigma(h)]$, $\E[vz^\top]$, $P:=J\Sigma$ and $K$ between $x$ and $y$ are at most $C\|\delta\|_\star$ in the corresponding parts of $\|\cdot\|_\star$. Finally, with primes denoting evaluation at $y$, $J\Sigma J-J'\Sigma'J'=(P-P')J+J'(P-P')^\top+J'(\Sigma'-\Sigma)J$, and $\|\mathrm{diag}(\cdot)\|_2\le\sqrt k\|\cdot\|_2$, so the term $c_f\frac{\eta_f}{k}J\Sigma J$ inherits the bound on $P$. Summing the blocks of \eqref{eq:S-drift} proves the bound on $b$. For $\Phi=\frac12\E[e^2]$,
$\frac{d}{d\theta}\Phi(x_\theta)=\E[e\,\delta e]+\frac1{2k^2}\E[w^\top\delta\Sigma\,w]+\frac1{2k}\E\big[e\langle a\circ\sigma''(h),\mathrm{diag}\,\delta\Sigma\rangle\big]$, which is at most $L_\Phi\|\delta\|_\star$ by the same bounds (without $\sigma'''$).

\emph{(S2).} For both procedures $|e_t|\le s_0\max_i|a_{i,t}|+\|g\|_\infty$, so their common $a$-update gives $\max_i|a_{i,t+1}|\le(1+C_a/k)\max_i|a_{i,t}|+C_a/k$ (as in Appendix~\ref{app:stability}, Step~2) and hence, since $|a_{i,0}|\le A$, deterministically $\max_i|a_{i,t}|\le e^{C_at/k}(A+1)\le\bar A_\tau$ for $t\le\tau n$; thus $|v_{i,t}|\le C_v$. Let $q_{i,t}:=\|m_{i,t}\|_2^2+(\Sigma_t)_{ii}$, $T:=\min\{t:\max_iq_{i,t}>\bar Q_\tau\}\wedge\lfloor\tau n\rfloor$, and split $q_{i,t+1}-q_{i,t}=\beta_{i,t}+\xi_{i,t}$ into its conditional mean and a martingale increment. For SGD, $q_{i,t+1}-q_{i,t}=-\frac{2\eta_f}{k}v_ih_i+\frac{\eta_f^2}{k^2}v_i^2\|x_t\|_2^2$; for Decomposed Dynamics it is $-\frac{2\eta_f}{k}v_im_i^\top z+\frac{\eta_f^2}{k^2}v_i^2\|z\|_2^2-\frac{2\eta_f}{k}(J\Sigma)_{ii}+\frac{\eta_f^2}{k^2}(J\Sigma J)_{ii}+2\langle\tilde f_i,\zeta_i\rangle+\|\zeta_i\|_2^2$, with $\tilde f_i$ the $i$-th row of $\tilde F:=(I-\widetilde\Lambda_t)F_t^\perp$ and $\zeta_i\sim\mathcal N(0,\Gamma_{ii}P_\perp)$, $\Gamma_{ii}\le(\eta_f/k)^2C_v^2$. Since $|\E_t[v_ih_i]|,|(J\Sigma)_{ii}|=|\E_t[v_ih_i^\perp]|$ and $|\E_t[v_im_i^\top z]|$ are at most $C_v\sqrt{q_{i,t}}$, and, for $t<T$, $(J\Sigma J)_{ii}\le(\sum_l|J_{il}|\Sigma_{ll}^{1/2})^2\le C\bar Q_\tau$ (as $|J_{il}|\le s_1^2\bar A_\tau^2/k$ for $l\ne i$), we get for $t<T$ and $n$ large: $\beta_{i,t}\le\frac{C_0}{k}(1+\sqrt{q_{i,t}})$ with $C_0$ not depending on $\bar Q_\tau$, and $\E_t[\xi_{i,t}^4]\le C(1+\bar Q_\tau)^2k^{-4}$. By Burkholder's inequality, $\E\max_{t\le\tau n}|\sum_{s<t\wedge T}\xi_{i,s}|^4\le C(\tau n)^2(1+\bar Q_\tau)^2k^{-4}=O(k^{-2})$, so with probability $1-O(1/k)$ these $k$ martingales all stay in $[-1,1]$; moreover $\max_iq_{i,0}\le2$ with probability $1-ke^{-n/8}$, since $nq_{i,0}\sim\chi^2_n$. On these events, $q_{i,t}\le3+c_\tau(1+\sqrt{\bar Q_\tau})$ for all $i$ and $t\le T$, with $c_\tau:=2C_0\tau/\alpha$, and this is $<\bar Q_\tau$ for $\bar Q_\tau:=4(1+c_\tau)^2$; hence $T=\lfloor\tau n\rfloor$. Finally $\Sigma_t\succeq0$ for both procedures.

\emph{(S3).} For $t<T$, the increments $\xi_t:=X_{t+1}-X_t-\E_t[X_{t+1}-X_t]$ satisfy $\E_t\|\xi^a_t\|_2^2+\E_t\|\xi^M_t\|_F^2+\E_t\|\mathrm{diag}\,\xi^\Sigma_t\|_2^2\le C/k$ and $\E_t\|\xi^\Sigma_t\|_F^2\le C$: for SGD, from $\Delta\Sigma_t=-\frac{\eta_f}{k}(v_th_t^{\perp\top}+h_t^\perp v_t^\top)+\frac{\eta_f^2}{k^2}\|g_t\|_2^2v_tv_t^\top$ and $\E_t\|h_t^\perp\|_2^2=\Tr\Sigma_t\le k\bar Q_\tau$; for Decomposed Dynamics, from $(\xi^\Sigma_t)_{ii}=2\langle\tilde f_i,\zeta_i\rangle+\|\zeta_i\|_2^2-\E_t\|\zeta_i\|_2^2$ with $\sum_i\Gamma_{ii}\|\tilde f_i\|_2^2\le\max_i\Gamma_{ii}\,\Tr(\tilde F\tilde F^\top)=O(1/k)$ and $\mathrm{Var}_t\|\zeta_i\|_2^2=2n_\perp\Gamma_{ii}^2$, and from $\E_t\|\tilde F\zeta_t^\top\|_F^2=\Tr\Gamma_t\,\Tr(\tilde F\tilde F^\top)$ and $\E_t\|\zeta_t\|_F^4\le3(n_\perp\Tr\Gamma_t)^2$. Doob's $L^2$ maximal inequality, applied to each block of the martingales stopped at $T$, gives $\E\sup_{t\le\tau n}\|\mathcal M_{t\wedge T}\|_\star^2\le C\tau n/k^2=O(1/n)$, and $\P(T<\lfloor\tau n\rfloor)=o(1)$ by (S2).
\end{proof}

\noindent The constants grow with $\tau$ through $\bar A_\tau$ and $\bar Q_\tau$. The bounded initialization and the bound on $\sigma'''$ in \textbf{A3} make $L$ finite and independent of $n$ in this proof: for Gaussian $a_{i,0}$, $\max_ia_i^2\asymp\log k$, and the term $a_ia_j\E[e^2\sigma''(h_i)\sigma''(h_j)]$ of the $\Sigma_{ij}$-derivative of $K$ generically makes the constant in (S1) at least of order $\log k$; if $\sigma''$ has a jump at $0$, the $m$-drift of a neuron with $m_i=0$ is generically only $\frac12$-H\"older in $\Sigma_{ii}$ near $\Sigma_{ii}=0$.

\paragraph{Proof of Theorem~\ref{thm:equivalence-spike}.}
\emph{(1) Drift mismatch.} By (S2), SGD lies in $\mathcal R_n$ for all $t\le\tau n$ with probability $1-o(1)$. By Steps~2--3, its conditional mean increment at $x\in\mathcal R_n$ is $b(x)+\Delta b(x)$, where the $(a,M)$ components of $\Delta b$ vanish and its $\Sigma$ component is $n_\perp E_t-(\eta_f/k)^2J_t\Sigma J_t$, with $E_t$ from Lemma~\ref{lem:gram-onestep}. By Appendix~\ref{app:eq-step3}, Step~(iii), $|(n_\perp E_t)_{ij}|\le C(\eta_f/k)^2\sqrt{n}\,|a_ia_j|$, so $k^{-1}\|n_\perp E_t\|_2\le k^{-1}\|n_\perp E_t\|_F=O(n^{-3/2})$ and $k^{-1/2}\|\mathrm{diag}(n_\perp E_t)\|_2\le Ck^{-5/2}\sqrt n\,\bar A_\tau\|a\|_2=O(n^{-3/2})$. Since $\|J_t\|_2\le C_D$ and $\|\Sigma\|_2\le\Tr\Sigma\le k\bar Q_\tau$, the term $(\eta_f/k)^2J_t\Sigma J_t$ contributes $O(n^{-3/2})$ to $\|\Delta b(x)\|_\star$ (using $\|\mathrm{diag}(\cdot)\|_2\le\|\cdot\|_F\le\|J_t\|_2^2\Tr\Sigma$). Hence $\sum_{s<\tau n}\|\Delta b(X_s^{\mathrm{SGD}})\|_\star=O(n^{-1/2})$.
\emph{(2) Gr\"onwall.} Let $D_t:=X_t^{\mathrm{SGD}}-X_t^{\mathrm{DD}}$, so $D_0=0$ and
$D_t=\sum_{s<t}\big(b(X^{\mathrm{SGD}}_s)-b(X^{\mathrm{DD}}_s)\big)+\sum_{s<t}\Delta b(X^{\mathrm{SGD}}_s)+\mathcal M^{\mathrm{SGD}}_t-\mathcal M^{\mathrm{DD}}_t$.
On the event that both chains stay in $\mathcal R_n$, (S1) gives $\|D_t\|_\star\le\frac{L}{n}\sum_{s<t}\|D_s\|_\star+\vartheta_n$ with $\vartheta_n:=\sum_{s<\tau n}\|\Delta b(X_s^{\mathrm{SGD}})\|_\star+2\sup_{t\le\tau n}\max\big(\|\mathcal M^{\mathrm{SGD}}_t\|_\star,\|\mathcal M^{\mathrm{DD}}_t\|_\star\big)$, and the discrete Gr\"onwall inequality gives $\sup_{t\le\tau n}\|D_t\|_\star\le\vartheta_n e^{L\tau}=O_{\P}(n^{-1/2})$ by (1), (S2) and (S3).
\emph{(3) Test loss.} By (S1), $\sup_{t\le\tau n}|\mathcal L^{\mathrm{SGD}}_t-\mathcal L^{\mathrm{DD}}_t|\le L_\Phi\sup_{t\le\tau n}\|D_t\|_\star=O_{\P}(n^{-1/2})$.\qed

\noindent The argument uses only conditional mean increments; the unmatched fluctuations of (iv) enter only through (S3). The proof applies to any procedure with the same drift map $b$ whose increments obey the per-step bounds used in Lemma~\ref{lem:S} (so that (S2)--(S3) hold).

\section{Proof of Proposition~\ref{prop:variance-inflation} (variance inflation by teacher-induced structure)}
\label{app:variance-inflation}

We prove Proposition~\ref{prop:variance-inflation} by iterating the law of total variance for the effective predictor \eqref{eq:yhat-eff}. Throughout we condition on the sigma field $\mathcal F_t$ by using the notation $\E_t[\cdot] = \E[\cdot \mid \mathcal F_t]$, so that $(a_{i,t},m_{i,t})$ and
$(\hat c_t,B_t^{\mathrm{acc}})$ are measurable (deterministic).

\paragraph{Setup.}
Let $z_\star\sim\mathcal N(0,I_{k_\star})$, $\varepsilon\sim\mathcal N(0, I_k)$, $s\sim\mathcal N(0,I_{tR})$ be independent of each other and of $\mathcal F_t$,
and define $\hat y_t^{\mathrm{eff}}$ as in \eqref{eq:yhat-eff}. Define
\[
    y^{\mathrm{eff}}:=g(z_\star),\qquad
    2\mathcal L_t^{\mathrm{eff}}=\E\!\left[(\hat y_t^{\mathrm{eff}}-y^{\mathrm{eff}})^2 \,\middle|\, \mathcal F_t\right].
\]

\paragraph{Step 1: decompose by conditioning on $z_\star$.}
Conditioning first on $\sigma(z_\star,\mathcal F_t)$ and applying the law of total variance yields
\begin{align}
    \E\!\left[(\hat y_t^{\mathrm{eff}}-y^{\mathrm{eff}})^2 \,\middle|\, \mathcal F_t\right]
    &=
    \E_{z_\star}\!\left[
        \Big(\E[\hat y_t^{\mathrm{eff}}\mid z_\star,\mathcal F_t]-y^{\mathrm{eff}}\Big)^2
        \;+\;
        \Var\!\big(\hat y_t^{\mathrm{eff}}\mid z_\star,\mathcal F_t\big)
    \right].
    \label{eq:app-var-step1}
\end{align}
The first term in \eqref{eq:app-var-step1} is exactly $\mathcal L_{\mathrm{bias},t}$ from \eqref{eq:Leff-bias-res}. It remains to
decompose the conditional variance term.

\paragraph{Step 2: decompose the conditional variance by conditioning on $s$.}
Given $(z_\star,\mathcal F_t)$, the only randomness in $\hat y_t^{\mathrm{eff}}$ is from the independent pair $(\varepsilon,s)$.
Apply the law of total variance with conditioning $\sigma$-field $\sigma(z_\star,s,\mathcal F_t)$:
\begin{equation}
    \Var\!\big(\hat y_t^{\mathrm{eff}}\mid z_\star,\mathcal F_t\big)
    \;=\;
    \E_{s}\!\left[
        \Var_{\varepsilon}\!\big(\hat y_t^{\mathrm{eff}}\mid z_\star,s,\mathcal F_t\big)
    \right]
    \;+\;
    \Var_{s}\!\left(
        \E_{\varepsilon}\!\big[\hat y_t^{\mathrm{eff}}\mid z_\star,s,\mathcal F_t\big]
    \right).
    \label{eq:app-var-step2}
\end{equation}
Taking $\E_{z_\star}[\cdot]$ of both sides identifies the first term as $\mathcal L_{\mathrm{res},t}$ from \eqref{eq:Leff-bias-res}
and the second as $\mathcal L_{\mathrm{str},t}$ from \eqref{eq:Leff-str}.

\paragraph{Step 3: combine and conclude nonnegativity / strictness.}
Substituting \eqref{eq:app-var-step2} into \eqref{eq:app-var-step1} gives the decomposition \eqref{eq:Leff-decomp}.
Nonnegativity of $\mathcal L_{\mathrm{str},t}$ is immediate since it is an expectation of a variance.
Moreover, $\mathcal L_{\mathrm{str},t}=0$ whenever $B_t^{\mathrm{acc}}=0$ since then $\hat y_t^{\mathrm{eff}}$ is independent of $s$ and the
inner variance vanishes almost surely. Conversely, if $\E_{\varepsilon}[\hat y_t^{\mathrm{eff}}\mid z_\star,s,\mathcal F_t]$ is not almost surely constant in $s$ on a set of $z_\star$ of positive probability, then the variance in \eqref{eq:Leff-str} is positive on that set and
hence $\mathcal L_{\mathrm{str},t}>0$.
The condition of Proposition~\ref{prop:variance-inflation} ensures this. Indeed, differentiating $\E_\varepsilon[\hat y_t^{\mathrm{eff}}\mid z_\star, s, \mathcal F_t] = \frac1k\sum_i a_i\,\tilde\sigma_{\hat c_t}(m_i^\top z_\star + (B_t^{\mathrm{acc}} s)_i)$ with $\tilde\sigma_{\hat c_t}(u) := \E_{\varepsilon_1}[\sigma(u + \sqrt{\hat c_t}\varepsilon_1)]$, the $s$-derivative at $s=0$ has $j$-th component $\frac1k\sum_i a_i\,\tilde\sigma_{\hat c_t}'(m_i^\top z_\star)\,(B_t^{\mathrm{acc}})_{ij}=\frac1k\big((B_t^{\mathrm{acc}})^\top\omega_t(z_\star)\big)_j$, since $\tilde\sigma_{\hat c_t}'(u)=\E_{\varepsilon_1}[\sigma'(u+\sqrt{\hat c_t}\varepsilon_1)]$. By that condition, this derivative is non-zero at some $z_\star$, hence, by continuity in $z_\star$, on an open set of $z_\star$, which has positive probability. On that set the conditional mean (of class $C^1$ in $s$) is not almost surely constant in $s$, since $s$ has full support; hence $\mathcal L_{\mathrm{str},t}>0$.
\qed

\paragraph{Remark (the cross-neuron term).}
For the network itself, with $h_i:=m_{i,t}^\top z_\star+h_i^\perp$ and $h^\perp\sim\mathcal N(0,\Sigma_t)$ independent of $z_\star$, write $2\mathcal L_t=\mathrm{Bias}^2+\mathrm{VarDiag}+\mathrm{Cross}$ with $\mathrm{Bias}^2:=\E_{z_\star}(\E[\hat y\mid z_\star]-g(z_\star))^2$ for $\hat y:=\frac1k\sum_ia_{i,t}\sigma(h_i)$, $\mathrm{VarDiag}:=\frac{1}{k^2}\E_{z_\star}\sum_{i}a_{i,t}^2\Var(\sigma(h_i)\mid z_\star)$ and $\mathrm{Cross}:=\frac{1}{k^2}\E_{z_\star}\sum_{i\ne j}a_{i,t}a_{j,t}\Cov(\sigma(h_i),\sigma(h_j)\mid z_\star)$. The first two terms depend on $\Sigma_t$ only through its diagonal, so, relative to a network with the same per-neuron variances and no cross-neuron structure, the structure changes $2\mathcal L_t$ by exactly $\mathrm{Cross}$. To first order (Mehler), $\mathrm{Cross}\approx\frac1{k^2}\E_{z_\star}[w^\top(\Sigma_t-\mathrm{diag}\,\Sigma_t)w]$ with $w_i:=a_{i,t}\E[\sigma'(h_i)\mid z_\star]$: it is positive for spikes aligned with $w$ and can be negative for structure orthogonal to $w$. Since $\Var(\hat y\mid z_\star)\ge0$, we have $\mathrm{Cross}\ge-\mathrm{VarDiag}$, and $\mathrm{VarDiag}\le\|\sigma\|_\infty^2B_\tau/k$ on \textbf{A6}; hence negative values of $\mathrm{Cross}$ are $O(1/k)$, while, to first order, spikes of $\Sigma_t$ with eigenvalue of order $k$ aligned with $w$ give $\mathrm{Cross}=\Theta(1)$ when $\|w\|_2^2$ is of order $k$. Likewise, as the $\varepsilon_i$ are independent across neurons, $\mathcal L_{\mathrm{res},t}\le\|\sigma\|_\infty^2B_\tau/k$, and $\mathcal L_{\mathrm{str},t}$ differs from the cross-neuron term of $\hat y_t^{\mathrm{eff}}$ (its excess over the same per-neuron variances, as above) by at most $\|\sigma\|_\infty^2B_\tau/k$.

\section{Proof of Proposition~\ref{prop:fast-alignment} (fast alignment of $m_t$ and $a_t$)}
\label{app:fast-alignment}

This appendix formalizes the oracle no-spike dynamics and proves
Proposition~\ref{prop:fast-alignment}. We work throughout in the single-index case $k_\star=1$
(and hence $m_{i,t}\in\R$).

\subsection{Additional conditions and oracle no-spike dynamics}

The proposition imposes the following additional conditions; \textbf{B3} is an idealization.

\begin{enumerate}
\item[\textbf{B1.}] \textbf{Odd teacher and odd activation, with nontrivial Hermite-1 coefficients.}
The teacher $g:\R\to\R$ is bounded, Lipschitz, and odd, and the activation $\sigma$ is odd (recall that
$\sigma\in W^{3,\infty}$ from \textbf{A3}). Furthermore,
\begin{equation}
\kappa_g \;:=\; \E[g(Z)Z]\ \neq\ 0, \qquad Z\sim\mathcal N(0,1).
\label{eq:kappa-g}
\end{equation}

\item[\textbf{B2.}] \textbf{Non-degenerate average slope.}
\begin{equation}
\kappa_\sigma \;:=\; \E[\sigma'(Z)]\ \neq\ 0, \qquad Z\sim\mathcal N(0,1).
\label{eq:kappa-sigma}
\end{equation}

\item[\textbf{B3.}] \textbf{Ideal isotropic complement channel.}
The oracle dynamics below replace the complement projections by synthetic variables
$z_t=(z_{1,t},\dots,z_{k,t})\sim\mathcal N(0,I_k)$, independent across $t$, of $z_{0,t}$ and of $\mathcal F_t$, with $(a_0,m_0)=(a_0,F_0\xi)$ initialized as in \textbf{A5}. This idealizes the complement channel (unit variances, no cross-neuron covariance); directions $U_0$ frozen at initialization would instead give covariance $U_0U_0^\top$.
\end{enumerate}

\noindent Like Theorem~\ref{thm:equivalence-spike}, the proposition also assumes bounded second-layer initialization: $|a_{i,0}|\le A$ almost surely for a constant $A<\infty$. This holds, e.g., for Rademacher or uniform $a_{i,0}$, and is standard in rigorous analyses of gradient-based training of two-layer networks \cite{ChizatBach2018OT,GoldtEtAl2020TeacherStudentSGD,AbbeBoixAdseraMisiakiewicz2022}. It is used only to bound $\max_i|a_{i,t}|$ by a constant (Appendix~\ref{app:stability}, Step~2, and Step~4 below); under \textbf{A5} alone this maximum can grow like $\sqrt{\log k}$ (e.g.,\ for Gaussian $a_{i,0}$). Our simulations use $a_{i,0}\sim\mathcal N(0,1)$; rerunning them with bounded $a_{i,0}$ of the same variance gives the same results.

\paragraph{Oracle no-spike dynamics.}
Let $\Xi=\xi\in\R^n$ and $z_{0,t}:=\xi^\top x_t\sim\mathcal N(0,1)$. In place of $f_{i,t}^\top x_t$, the oracle uses the synthetic preactivations of \textbf{B3},
\[
h_{i,t}:=m_{i,t} z_{0,t} + z_{i,t},
\qquad z_{i,t}\sim\mathcal N(0,1) \text{ i.i.d. and independent of } z_{0,t}.
\]
Define
\[
\hat y_t=\frac1k\sum_{\ell=1}^k a_{\ell,t}\sigma(h_{\ell,t}),\qquad
e_t=\hat y_t-g(z_{0,t}).
\]
The oracle no-spike dynamics evolves $(a_t,m_t)$ using the conditional drift
of the single-sample SGD updates (with ridge term $\lambda\ge 0$ on $a$ consistent with $\frac{\lambda}{2k}\|a\|_2^2$):
\begin{align}
a_{i,t+1} &= a_{i,t} - \frac{\eta_a}{k}\,\E\!\left[e_t\,\sigma(h_{i,t}) \,\middle|\, \mathcal F_t\right]
            - \frac{\eta_a\lambda}{k}\,a_{i,t},
\label{eq:dd0-a-drift}
\\
m_{i,t+1} &= m_{i,t} - \frac{\eta_f}{k}\,\E\!\left[e_t\,a_{i,t}\,\sigma'(h_{i,t})\,z_{0,t} \,\middle|\, \mathcal F_t\right].
\label{eq:dd0-m-drift}
\end{align}

\subsection{Proof of Proposition~\ref{prop:fast-alignment}}

Let $\kappa:=\kappa_g\kappa_\sigma$ and recall $\cos_t=a_t^\top m_t/(\|a_t\|_2\|m_t\|_2)$.

\paragraph{Step 1: linearization of the teacher contribution.}
Fix any $m\in\R$. Define the scalar functions
\begin{equation}
\Phi(m) \;:=\; \E\!\left[g(Z_0)\,\sigma(mZ_0+Z)\right],
\qquad
\Psi(m) \;:=\; \E\!\left[g(Z_0)\,Z_0\,\sigma'(mZ_0+Z)\right],
\label{eq:Phi-Psi-def}
\end{equation}
where $(Z_0,Z)$ are i.i.d. $\mathcal N(0,1)$. Under \textbf{A3} and \textbf{B1}, $\Phi$ is $C^1$ with
$\Phi'(m)=\Psi(m)$ by dominated convergence (bounded $\sigma'$ and bounded $g$). Moreover, by independence of
$Z_0$ and $Z$ and the fact that $g$ is odd (\textbf{B1}, so $\E[g(Z_0)]=0$),
\[
\Phi(0)=\E[g(Z_0)]\,\E[\sigma(Z)]=0,
\qquad
\Psi(0)=\E[g(Z_0)Z_0]\,\E[\sigma'(Z)]=\kappa_g\kappa_\sigma=\kappa.
\]
Since $\sigma'$ is Lipschitz (\textbf{A3}) and $g(Z_0)Z_0$ has finite moments (bounded $g$), $\Psi$ is Lipschitz and hence,
for any $|m|\le \bar m$,
\begin{equation}
\Phi(m)=\kappa\,m + R_\Phi(m),\qquad
\Psi(m)=\kappa + R_\Psi(m),
\qquad
|R_\Phi(m)|\le C\bar m\,|m|,\ \ |R_\Psi(m)|\le C|m|.
\label{eq:Phi-Psi-linear}
\end{equation}
Here $C<\infty$ depends only on the constants in \textbf{A3} and the bound on $g$.

\paragraph{Step 2: conditional drift is a $2\times 2$ linear system plus a small remainder.}
Fix $t$ and condition on $\mathcal F_t$. Under \textbf{B3}, $(z_{0,t},z_t)$ are independent standard Gaussians.
Write the conditional drift terms in \eqref{eq:dd0-a-drift}--\eqref{eq:dd0-m-drift} as
\begin{align}
\E[e_t\sigma(h_{i,t})\mid \mathcal F_t]
&=\E[\hat y_t\,\sigma(h_{i,t})\mid\mathcal F_t]-\E[g(z_{0,t})\sigma(h_{i,t})\mid\mathcal F_t],
\label{eq:split-a-drift}
\\
\E[e_ta_{i,t}\sigma'(h_{i,t})z_{0,t}\mid \mathcal F_t]
&=\E[\hat y_t\,a_{i,t}\sigma'(h_{i,t})z_{0,t}\mid\mathcal F_t]
 -\E[g(z_{0,t})\,a_{i,t}\sigma'(h_{i,t})z_{0,t}\mid\mathcal F_t].
\label{eq:split-m-drift}
\end{align}

\emph{Teacher contributions.}
Since $h_{i,t}=m_{i,t}z_{0,t}+z_{i,t}$ and $(z_{0,t},z_{i,t})$ are i.i.d. standard Gaussians,
\[
\E[g(z_{0,t})\sigma(h_{i,t})\mid\mathcal F_t]=\Phi(m_{i,t}),
\qquad
\E[g(z_{0,t})\,\sigma'(h_{i,t})z_{0,t}\mid\mathcal F_t]=\Psi(m_{i,t}).
\]
Thus on the event $\mathcal E_{\bar m}(t):=\{\max_{i\le k}|m_{i,t}|\le \bar m\}$, \eqref{eq:Phi-Psi-linear} yields
\begin{equation}
\E[g(z_{0,t})\sigma(h_{i,t})\mid\mathcal F_t]=\kappa\,m_{i,t} + r_{i,t}^{(m)},
\qquad
\E[g(z_{0,t})\,a_{i,t}\sigma'(h_{i,t})z_{0,t}\mid\mathcal F_t]=\kappa\,a_{i,t} + r_{i,t}^{(a)},
\label{eq:teacher-linear}
\end{equation}
with $|r_{i,t}^{(m)}|\le C\bar m\,|m_{i,t}|$ and $|r_{i,t}^{(a)}|\le C\bar m\,|a_{i,t}|$.

\emph{Self-interaction contributions.}
On $\mathcal E_{\bar m}(t)$, expanding $\hat y_t = \frac1k\sum_\ell a_\ell\sigma(h_\ell)$ with
$h_\ell = m_\ell z_{0,t} + z_{\ell,t}$ and using \textbf{B1} ($\sigma$ odd, so that
$\tilde\sigma(u) := \E_z[\sigma(u+z)]$ is also odd, with $\tilde\sigma(0)=0$ and Taylor expansion
$\tilde\sigma(u) = \kappa_\sigma u + O(u^3)$), we obtain by independence of $z_{\ell,t}$ across $\ell$,
\[
    \E[\hat y_t^2\mid\mathcal F_t]
    \;\le\;
    \frac{\|\sigma\|_\infty^2}{k^2}\sum_\ell a_{\ell,t}^2
    + \frac{\kappa_\sigma^2}{k^2}\,(a_t^\top m_t)^2 + O(\bar m^4)
    \;=\; O\!\Big(\tfrac{1}{k} + \bar m^2\Big),
\]
since on $\mathcal E_{\bar m}$ together with $\max_i|a_{i,t}|\le\bar A$ from Step~4, $|a_t^\top m_t|^2 \le \|a_t\|^2 \|m_t\|^2 \le k\bar m^2 \cdot O(k)$.
By Cauchy--Schwarz and bounded $\sigma,\sigma'$,
\begin{equation}
\big|\E[\hat y_t\,\sigma(h_{i,t})\mid\mathcal F_t]\big|
\;\le\; C\,(\bar m + k^{-1/2}),
\qquad
\big|\E[\hat y_t\,a_{i,t}\sigma'(h_{i,t})z_{0,t}\mid\mathcal F_t]\big|
\;\le\; C\,|a_{i,t}|\,(\bar m + k^{-1/2}).
\label{eq:self-bounds}
\end{equation}
(In the proportional regime the $k^{-1/2}$ term vanishes, so we treat the right-hand sides as $O(\bar m)$ in what follows.)

Combining \eqref{eq:split-a-drift}--\eqref{eq:self-bounds} with \eqref{eq:teacher-linear} gives, on
$\mathcal E_{\bar m}(t)$,
\begin{align}
\E[e_t\sigma(h_{i,t})\mid \mathcal F_t]
&= -\kappa\,m_{i,t} + \delta^{(a)}_{i,t},
\label{eq:drift-a-linear}
\\
\E[e_ta_{i,t}\sigma'(h_{i,t})z_{0,t}\mid \mathcal F_t]
&= -\kappa\,a_{i,t} + \delta^{(m)}_{i,t},
\label{eq:drift-m-linear}
\end{align}
with bounds
\begin{equation}
|\delta^{(a)}_{i,t}|\le C\,\bar m\,(|m_{i,t}| + 1),
\qquad
|\delta^{(m)}_{i,t}|\le C\,\bar m\,|a_{i,t}|.
\label{eq:delta-bounds}
\end{equation}

Insert \eqref{eq:drift-a-linear}--\eqref{eq:drift-m-linear} into \eqref{eq:dd0-a-drift}--\eqref{eq:dd0-m-drift}.
In vector form, for $t$ such that $\mathcal E_{\bar m}(t)$ holds,
\begin{equation}
\begin{bmatrix} a_{t+1}\\ m_{t+1}\end{bmatrix}
=
\begin{bmatrix} a_t\\ m_t\end{bmatrix}
+
\frac{1}{k}\,
\underbrace{
\begin{bmatrix}
-\eta_a\lambda & \eta_a\kappa\\
\eta_f\kappa & 0
\end{bmatrix}
}_{=:G}
\begin{bmatrix} a_t\\ m_t\end{bmatrix}
-
\frac{1}{k}
\begin{bmatrix}
\eta_a\,\delta^{(a)}_t\\
\eta_f\,\delta^{(m)}_t
\end{bmatrix}.
\label{eq:linear-plus-remainder}
\end{equation}

\paragraph{Step 3: alignment for the limiting linear flow.}
Ignore the remainder in \eqref{eq:linear-plus-remainder} and consider the linear ODE on macroscopic time $s=t/k$:
\begin{equation}
\frac{d}{ds}
\begin{bmatrix} a(s)\\ m(s)\end{bmatrix}
=
G\begin{bmatrix} a(s)\\ m(s)\end{bmatrix},
\qquad (a(0),m(0))=(a_0,m_0).
\label{eq:linear-ode}
\end{equation}
The matrix $G$ has eigenvalues
\begin{equation}
\mu_\pm \;=\; \frac{-\eta_a\lambda \pm \sqrt{\eta_a^2\lambda^2+4\eta_a\eta_f\kappa^2}}{2},
\qquad
\mathrm{gap}:=\mu_+-\mu_-=\sqrt{\eta_a^2\lambda^2+4\eta_a\eta_f\kappa^2}\;>\;0,
\label{eq:eigs-gap}
\end{equation}
since $\kappa\neq 0$ by \textbf{B1--B2}.
Let $X(s):=[a(s),m(s)]\in\R^{k\times2}$. Then $X(s)=X(0)e^{sG^\top}$, hence $\det(X(s)^\top X(s))=e^{2s\,\mathrm{tr}G}\det(X(0)^\top X(0))$ with $\mathrm{tr}\,G=-\eta_a\lambda$. Since $\det(X^\top X)=\|a\|_2^2\|m\|_2^2(1-\cos^2)$,
\begin{equation}
1-\cos_s^2=(1-\cos_0^2)\,e^{-2\eta_a\lambda s}\,\frac{\|a_0\|_2^2\,\|m_0\|_2^2}{\|a(s)\|_2^2\,\|m(s)\|_2^2}.
\label{eq:mode-separation}
\end{equation}
Under \textbf{A5}, with probability $1-o(1)$: $\|a_0\|_2^2\asymp k$, $\|m_0\|_2^2=\alpha(1+o(1))$ and $|a_0^\top m_0|=O(\sqrt{\log n})$. Expanding $m(s)=m_0+s\eta_f\kappa a_0+O(s^2)(\|a_0\|_2+\|m_0\|_2)$ and using $\|a(s)\|_2\ge\|a_0\|_2e^{-\|G\|s}$ gives, for $s\le\tau_0'$, with $\tau_0'>0$ small, $\|m(s)\|_2^2\ge\|m_0\|_2^2(1-Cs\sqrt{\log n})+c_0\,s^2k$ with $c_0>0$ (from the term $s^2\eta_f^2\kappa^2\|a_0\|_2^2$, as $\kappa\neq0$), hence, with $x_+^{-1}:=\infty$ for $x\le0$,
\[
1-\cos_s^2\le(1-\cos_0^2)\,e^{Cs}\min\Big\{\big(1-Cs\sqrt{\log n}\big)_+^{-1},\ \frac{C\alpha}{s^2k}\Big\}.
\]
For $s\le k^{-1/3}$ the first term gives $1-\cos_s^2\le(1-\cos_0^2)e^{-cs}+O(k^{-1/3}\sqrt{\log n})$ for any fixed $c>0$; for $s\ge k^{-1/3}$ the second gives $1-\cos_s^2=O(k^{-1/3})$. Hence, with $c:=2\,\mathrm{gap}$,
\begin{equation}
1-\cos_s^2\le(1-\cos_0^2)\,e^{-cs}+o(1)\qquad\text{uniformly on }[0,\tau_0'],
\label{eq:ode-alignment}
\end{equation}
and $\|a(s)\|_2\ge c_1\sqrt k$, $\|m(s)\|_2\ge c_1\max(\sqrt\alpha,s\sqrt k)$ on $[0,\tau_0']$ for a constant $c_1>0$.

\paragraph{Step 4: bootstrap and stability of the Euler scheme.}
We return to the discrete recursion \eqref{eq:linear-plus-remainder}; fix $\varepsilon>0$.

\emph{Bootstrap.} The oracle dynamics are deterministic given $(a_0,m_0)$. As in Steps~1--2 of Appendix~\ref{app:stability}, $\frac1k\|a_t\|_2^2$ stays bounded and $|e_t|\le C$, so $|a_{i,t+1}|\le(1+C/k)|a_{i,t}|+C/k$ and, since $|a_{i,0}|\le A$, $\max_i|a_{i,t}|\le\bar A:=e^{C\tau_0'}(A+1)$ for $t\le\tau_0'k$. The drift \eqref{eq:dd0-m-drift} is $a_{i,t}$ times a bounded factor, so $|m_{i,t}-m_{i,0}|\le C\eta_f\bar A\,t/k$. Since $\max_i|m_{i,0}|=O(\sqrt{\log n/n})$ with probability $1-o(1)$, $\mathcal E_{\bar m}(t)$ holds for all $t\le\tau_0k$ with $\tau_0:=\min\{\tau_0',\,\bar m/(2C\eta_f\bar A)\}$ ($\tau_0'$ from Step~3 above), for every $\bar m>0$ and $n$ large.

\emph{Relative error.} On this window, let $X^{(k)}(s)$ interpolate the discrete trajectory, let $X(s)$ solve \eqref{eq:linear-ode}, and let $D:=X^{(k)}-X$. By \eqref{eq:linear-plus-remainder}, $D$ solves the linear equation driven by the remainders \eqref{eq:delta-bounds}, up to an Euler error $O(s\|X\|_2/k)$. Since $\|\delta^{(m)}\|_2\le C\bar m\|a\|_2$ and $\|\delta^{(a)}\|_2\le C\bar m(\|m\|_2+\sqrt k)$, Gr\"onwall gives $\|D_a(s)\|_2+\|D_m(s)\|_2\le C\bar m\,s\sqrt k$ on $[0,\tau_0]$. With the lower bounds on $\|a(s)\|_2$ and $\|m(s)\|_2$ from Step~3, the relative errors of both directions are $O(\bar m)$, and since $1-\cos^2$ is Lipschitz in the two unit directions, $|(1-\cos_t^2)-(1-\cos^2_{s})|_{s=t/k}\le C'\bar m$. Choosing $\bar m:=\varepsilon/(2C')$ and using \eqref{eq:ode-alignment}, whose $o(1)$ term is below $\varepsilon/2$ for $n$ large,
\[
1-\cos_t^2\;\le\;(1-\cos_0^2)\exp\!\Big(-c\,\frac{t}{k}\Big)+\varepsilon,
\qquad \forall\,0\le t\le \tau_0(\varepsilon)\, k,
\]
which is the discrete analogue of \eqref{eq:ode-alignment}. Moreover, for $s\ge k^{-1/3}$ Step~3 gives $1-\cos_s^2=O(k^{-1/3})$ for the linear flow, hence $1-\cos_t^2\le C'\bar m+O(k^{-1/3})\le\varepsilon$ for all $k^{2/3}\le t\le\tau_0k$ and $n$ large, which is \eqref{eq:fast-alignment-bound-main}.\qed

\section{Moment stability under the mean-field scaling}
\label{app:stability}

\begin{lemma}[Moment stability: \textbf{A6} holds under \textbf{A1}--\textbf{A5}]
\label{lem:stability-sufficient}
Assume \textbf{A1--A5}. Consider SGD with loss
$\ell=\frac12(\hat y-y)^2+\frac{\lambda}{2k}\|a\|_2^2$ and updates \eqref{eq:sgd-update-f}--\eqref{eq:sgd-update-a}.
Fix $\tau>0$ and suppose $\eta_a,\eta_f=O(1)$ and $\lambda\ge 0$ are fixed constants.
Then there exists $B_\tau<\infty$ --- depending only on
$\tau,\,\alpha,\,\eta_a,\,\eta_f,\,\lambda,\,k_\star,\,L_g,\,\|g\|_\infty,\,
\|\sigma\|_\infty,\,\|\sigma'\|_\infty,\,\|\sigma''\|_\infty$, and $K_{\mathrm{sg}}$ --- such that \textbf{A6}
holds, i.e.,
\[
        \sup_{t\le \tau n}\frac{1}{k}\sum_{i=1}^k\Big(a_{i,t}^2+\|m_{i,t}\|_2^2+r_{i,t}^2\Big)\ \le\ B_\tau
\]
with probability $1-o(1)$ as $n\to\infty$. Moreover, with probability $1-o(1)$, $\max_{i\le k}\sup_{t\le\tau n}a_{i,t}^2\le A_n^2:=C_\tau'\log k$ (the event $\mathcal E^{\sup}_a$ of \eqref{eq:app-stab-Esup-a}).
\end{lemma}

\begin{proof}
Fix $\tau>0$ and let $T_{\max}:=\lfloor \tau n\rfloor$. Write
$\bar a_t:=(1/k)\|a_t\|_2^2$, $w_t:=(1/k)\sum_i\|m_{i,t}\|_2^2$, $u_t:=(1/k)\sum_i r_{i,t}^2$.

\paragraph{Step 1 (control of $\bar a_t$ and $|e_t|$).}
Expanding $\|a_{t+1}\|_2^2$ from \eqref{eq:sgd-update-a} and using bounded $\sigma$ ($\|\sigma_t\|_2\le\|\sigma\|_\infty\sqrt{k}$), bounded teacher ($|y_t|\le\|g\|_\infty$), and Young's inequality on the cross term, one obtains the deterministic recursion
\begin{equation}
\bar a_{t+1}\;\le\;(1+C_1/k)\,\bar a_t \;+\; C_1/k,
\qquad C_1=C(\eta_a,\lambda,\|\sigma\|_\infty,\|g\|_\infty).
\label{eq:app-stab-a-recursion}
\end{equation}
By the law of large numbers and \textbf{A5} (sub-Gaussian, hence finite second moment), $\bar a_0=O(1)$ with probability $1-o(1)$. Discrete Grönwall on \eqref{eq:app-stab-a-recursion} then yields
\begin{equation}
    \sup_{t\le T_{\max}}\bar a_t\;\le\;C_\tau,
    \qquad
    \sup_{t\le T_{\max}}|e_t|\;\le\;\|\sigma\|_\infty\sqrt{C_\tau}+\|g\|_\infty,
    \label{eq:app-stab-step1}
\end{equation}
with probability $1-o(1)$, with $C_\tau$ depending only on $\tau,\alpha,\eta_a,\lambda,\|\sigma\|_\infty,\|g\|_\infty,\E[a_{i,0}^2]$.

\paragraph{Step 2 (per-neuron sup-norm event $\mathcal E^{\sup}_a$).}
Let $A_t:=\max_{i\le k}|a_{i,t}|$. Since $|e_t|\le\|\sigma\|_\infty A_t+\|g\|_\infty$, the $a$-update gives, for $n$ large enough that $\eta_a\lambda\le k$ and $n/k\le2/\alpha$, the deterministic recursion $A_{t+1}\le(1+C_a/k)A_t+C_a/k$ with $C_a:=\eta_a\|\sigma\|_\infty\max(\|\sigma\|_\infty,\|g\|_\infty)$, and Grönwall yields
\[
    \max_{i\le k}\sup_{t\le T_{\max}}|a_{i,t}|\;\le\;e^{C_aT_{\max}/k}\big(\max_i|a_{i,0}|+1\big)\;\le\;e^{2C_a\tau/\alpha}\big(\max_i|a_{i,0}|+1\big).
\]
By the sub-Gaussian maximal inequality, $\max_i|a_{i,0}|\le K_{\mathrm{sg}}\sqrt{2\log k}\,(1+o(1))$ with probability $1-o(1)$, defining the event
\begin{equation}
    \mathcal E^{\sup}_a\;:=\;\Big\{\max_{i\le k}\sup_{t\le T_{\max}}a_{i,t}^2\le A_n^2\Big\},
    \qquad A_n^2:=C_\tau'\log k,
    \label{eq:app-stab-Esup-a}
\end{equation}
with $\P(\mathcal E^{\sup}_a)\ge 1-o(1)$.

\paragraph{Step 3 (control of $w_t$ and $u_t$ via drift+martingale).}
The recursions
$m_{i,t+1}=m_{i,t}-(\eta_f/k)v_{i,t}z_{\star,t}$ and
$f^\perp_{i,t+1}=f^\perp_{i,t}-(\eta_f/k)v_{i,t}g_t$,
with $v_{i,t}=e_t a_{i,t}\sigma'(f_{i,t}^\top x_t)$, $z_{\star,t}\sim\mathcal N(0,I_{k_\star})$ and $g_t\sim\mathcal N(0,P_\perp)$, expand into per-neuron increments
\[
    \|m_{i,t+1}\|^2-\|m_{i,t}\|^2=-\tfrac{2\eta_f}{k}v_i(m_i^\top z_\star)+\tfrac{\eta_f^2}{k^2}v_i^2\|z_\star\|^2,
\]
and analogously for $r_{i,t}^2$ with $g_t$ and $h^\perp_i:=\langle f^\perp_i,g_t\rangle$ in place of $z_\star$ and $m_i^\top z_\star$. Note $|v_{i,t}|\le C|a_{i,t}|$ pathwise on Step~1's event.

\smallskip\emph{Drift.} By conditional Cauchy--Schwarz ($\E_t[v_i(m_i^\top z_\star)]\le\sqrt{\E_t[v_i^2]}\sqrt{\E_t[(m_i^\top z_\star)^2]}\le C|a_i|\|m_i\|$, and analogously $\E_t[v_ih^\perp_i]\le C|a_i|r_i$), Cauchy in $i$, and Step~1's bound on $\bar a_t$,
\begin{equation}
    |\E_t[w_{t+1}-w_t]|\,\le\,\tfrac{1}{k}K_\tau(\sqrt{w_t}+1),\qquad
    |\E_t[u_{t+1}-u_t]|\,\le\,\tfrac{1}{k}K_\tau(\sqrt{u_t}+1).
    \label{eq:app-stab-drift}
\end{equation}
For the stopping times defined below, $w_s\le B^{(m)}$ for $s<T^{(m)}$ (resp.\ $u_s\le B^*$ for $s<T^{(r)}$), so the drift accumulated over the steps $s<T^{(m)}\wedge t$ (resp.\ $s<T^{(r)}\wedge t$) is at most $\tau K_\tau(\sqrt{B}+1)$ for every $t\le T_{\max}$, with $B=B^{(m)}$ (resp.\ $B^*$) and $K_\tau$ enlarged to absorb the factor $n/k\to1/\alpha$.

\smallskip\emph{Martingale variance via $\mathcal E^{\sup}_a$.} On $\mathcal E^{\sup}_a$ and Step~1's event, $|v_i|\le C|a_i|\le CA_n$ pointwise, so $\sum_iv_i^4\le CA_n^2\sum_ia_i^2\le CA_n^2k$. Using $\E_t[(m_i^\top z_\star)^2]=\|m_i\|^2$, $\E_t[(h^\perp_i)^2]=r_i^2$, $\E[\|z_\star\|^4]=O(1)$, and $\E[\|g_t\|^4]=O(n^2)$, expanding $(\Delta\|m_i\|^2)^2$ and $(\Delta r_i^2)^2$ and summing in $i$:
\[
    \sum_{i}\E_t[(\Delta\|m_i\|^2)^2]\le\frac{C\,A_n^2}{k}(w_t+1),\qquad
    \sum_{i}\E_t[(\Delta r_i^2)^2]\le\frac{C\,A_n^2}{k}(u_t+1),
\]
hence, by Cauchy--Schwarz in $i$, the centered martingale increments
$\xi^{(w)}_t:=(w_{t+1}-w_t)-\E_t[w_{t+1}-w_t]$ and
$\xi^{(u)}_t:=(u_{t+1}-u_t)-\E_t[u_{t+1}-u_t]$ satisfy
\begin{equation}
    \E_t[\xi^{(w)}_t{}^2]\le\frac{C\,A_n^2(w_t+1)}{k^2},
    \qquad
    \E_t[\xi^{(u)}_t{}^2]\le\frac{C\,A_n^2(u_t+1)}{k^2}.
    \label{eq:app-stab-xi-var}
\end{equation}
Let $M^{(w)}_t:=\sum_{s<t}\xi^{(w)}_s$ and $M^{(u)}_t:=\sum_{s<t}\xi^{(u)}_s$.

\smallskip\emph{Doob localization and bootstrap.} Define stopping times $T^{(m)}:=\inf\{t\le T_{\max}:w_t>B^{(m)}\}$, $T^{(r)}:=\inf\{t\le T_{\max}:u_t>B^*\}$, and $T^\circ:=\inf\{t\le T_{\max}:\text{Step~1's event or }\mathcal E^{\sup}_a\text{ fails at }t\}$, all $\wedge T_{\max}$, and let $T^{\ast,m}:=T^{(m)}\wedge T^\circ$, $T^{\ast,r}:=T^{(r)}\wedge T^\circ$. On $\{s<T^{\ast,m}\}$ (resp.\ $\{s<T^{\ast,r}\}$), \eqref{eq:app-stab-xi-var} holds with $w_s\le B^{(m)}$ (resp.\ $u_s\le B^*$). Doob's $L^2$ maximal inequality applied to the stopped martingales yields
\[
    \E\!\big[\sup_t(M^{(w)}_{T^{\ast,m}\wedge t})^2\big],\;\E\!\big[\sup_t(M^{(u)}_{T^{\ast,r}\wedge t})^2\big]\;=\;O\!\Big(\frac{(B+1)\,\tau\log n}{n}\Big)\;\to\;0,
\]
where $B$ stands for $B^{(m)}$ or $B^*$. By Markov at fixed $\delta>0$, $\sup_t|M^{(w)}_{T^{\ast,m}\wedge t}|=o_p(1)$ and $\sup_t|M^{(u)}_{T^{\ast,r}\wedge t}|=o_p(1)$, and a union bound with $\P(T^\circ<T_{\max})=o(1)$ gives the same conclusion with $T^\circ$-stopping removed.

The self-consistency
\begin{equation}
    1+\tau K_\tau(\sqrt{B^{(m)}}+1)+\delta\le B^{(m)},
    \qquad
    2+\tau K_\tau(\sqrt{B^*}+1)+\delta\le B^*,
    \label{eq:app-stab-self-consistent}
\end{equation}
is solvable at constants $B^{(m)},B^*=O((\tau K_\tau)^2+\delta^2+1)$ (LHS grows only as $\sqrt{B}$). Telescoping the $w$- and $u$-recursions and bounding the martingale by $\delta$ on the high-probability event yields $w_{T^{(m)}\wedge t}\le B^{(m)}$ and $u_{T^{(r)}\wedge t}\le B^*$ for all $t\le T_{\max}$. Since $w_{T^{(m)}}>B^{(m)}$ if $T^{(m)}<T_{\max}$, and likewise for $u$, this gives $T^{(m)}=T^{(r)}=T_{\max}$. With $w_0=O(1/n)$ from $\|m_{i,0}\|^2\sim\chi^2_{k_\star}/n$ and $u_0=1+o(1)$ from $\|f^\perp_{i,0}\|^2\sim\chi^2_{n-k_\star}/n$, $w_0\le1$ and $u_0\le2$ with probability $1-o(1)$.

\paragraph{Conclusion.} Setting $B_\tau:=C_\tau+B^{(m)}+B^*$ — a constant depending only on $(\tau,\alpha,\eta_a,\eta_f,\lambda,k_\star,L_g,\|g\|_\infty,\|\sigma\|_\infty,\|\sigma'\|_\infty,\|\sigma''\|_\infty,K_{\mathrm{sg}})$ — proves the lemma, establishing the moment-stability event \textbf{A6} under \textbf{A1}--\textbf{A5}.
\end{proof}
\subsection{Coupled second-moment bounds}
\label{app:stability-4th}

The Sobolev bound in Lemma~\ref{lem:app-mu-sobolev} of Appendix~\ref{app:teacher-finite-rank} controls
$\sum_i \E_{z_\star}[\|\nabla_{z_\star}\mu_{i,t}(z_\star)\|_2^2\mid\mathcal F_t]$, which decomposes (via the
chain rule) into terms of order $\sum_i a_{i,t}^2$ and the coupled sum $\sum_i a_{i,t}^2\|m_{i,t}\|_2^2$.
The first is bounded by Lemma~\ref{lem:stability-sufficient}; we now bound the coupled sum directly using the
per-neuron sup-norm event $\mathcal E^{\sup}_a$ of \eqref{eq:app-stab-Esup-a}.

\begin{lemma}[Coupled second-moment bounds]
\label{lem:stability-sufficient-4th}
Assume \textbf{A1}--\textbf{A5}. Fix $\tau>0$. There exists a sequence $B'_\tau(n)=O(\log n)$, with constants
depending only on the same parameters as $B_\tau$ in Lemma~\ref{lem:stability-sufficient}, such that with
probability $1-o(1)$ as $n\to\infty$,
\begin{equation}
    \sup_{t\le\tau n}\,\tfrac{1}{k}\sum_{i=1}^k a_{i,t}^2\|m_{i,t}\|_2^2 \;\le\; B'_\tau(n),\qquad
    \sup_{t\le\tau n}\,\tfrac{1}{k}\sum_{i=1}^k a_{i,t}^2 r_{i,t}^2 \;\le\; B'_\tau(n).
    \label{eq:app-stab-4th-coupled}
\end{equation}
One may take $B'_\tau(n):=A_n^2B_\tau=O(\log n)$; \eqref{eq:app-stab-4th-coupled} then holds on the intersection of $\mathcal E^{\sup}_a$ with the event \textbf{A6}.
\end{lemma}

\begin{proof}
On $\mathcal E^{\sup}_a$ (eq.~\eqref{eq:app-stab-Esup-a}), $\sup_{i,t}a_{i,t}^2\le A_n^2 = O(\log n)$. Hence
\[
    \tfrac{1}{k}\sum_i a_{i,t}^2\|m_{i,t}\|_2^2
    \;\le\; A_n^2\cdot \tfrac{1}{k}\sum_i\|m_{i,t}\|_2^2
    \;\le\; A_n^2 \cdot B_\tau,
\]
on the event \textbf{A6} of Lemma~\ref{lem:stability-sufficient}, and similarly for
$\sum_i a_{i,t}^2 r_{i,t}^2$. This event has probability $1-o(1)$. 
\end{proof}

\paragraph{Consequence for the bulk-plus-spike approximation.}
Combining \eqref{eq:app-stab-4th-coupled} with the Sobolev bound of
Lemma~\ref{lem:app-mu-sobolev} (which is established in Appendix~\ref{app:teacher-finite-rank}),
Theorems~\ref{thm:app-finite-rank} and~\ref{thm:app-accumulated-finite} hold with rate $G^2(n)/(D+1)$ where
$G^2(n)=O(\log n)$ via Lemma~\ref{lem:stability-sufficient-4th}. The Hermite truncation of $S_t$ and of the undamped sum $\sum_{s<t}S_s$ underlying the bulk-plus-spike approximation
$n_\perp G_t \approx \hat c_t I + B_t^{\mathrm{acc}}(B_t^{\mathrm{acc}})^\top$ used in Proposition~\ref{prop:variance-inflation} is therefore
established at rate $O(\log n/(D+1))$ in normalized Frobenius norm; choosing the truncation degree
$D=D(n)$ with $\log n/(D+1)\to0$ (e.g.,\ $D=\lceil(\log n)^2\rceil$) makes this $o(1)$. The same rate holds for the damped accumulation $\sum_s\Pi_{s,t}S_s\Pi_{s,t}^\top$ behind $B_t^{\mathrm{acc}}$ whenever the propagators are uniformly bounded, $\sup_{s<t\le\tau n}\|\Pi_{s,t}\|_2=O(1)$.

\section{Extensions: multi-index teacher and beyond smooth activations}
\label{app:extensions}

\subsection{From single-index ($k_\star=1$) to multi-index ($k_\star>1$)}
\label{app:multi-index}

Replace $z_0=\xi^\top x\in\mathbb R$ by $z_\star=\Xi^\top x\in\mathbb R^{k_\star}$ and
$m_{i,t}\in\mathbb R$ by $m_{i,t}=\Xi^\top f_{i,t}\in\mathbb R^{k_\star}$.
All arguments (except for Proposition~\ref{prop:fast-alignment}) extend with the same structure: $g_t=P_\perp x_t$ remains Gaussian and independent of
$z_{\star,t}$, and the conditional Gaussianity of test-time projections $F^{\perp}_t g$ holds with
$P_\perp=I-\Xi\Xi^\top$. Result by result:
\begin{itemize}[topsep=0pt,itemsep=-0.5ex,partopsep=1ex,parsep=1ex,leftmargin=*]
    \item \textbf{Complement update.} \eqref{eq:complement-update-main} holds verbatim, with $n_\perp=\Tr(P_\perp)=n-k_\star$.
    \item \textbf{Lemma~\ref{lem:complement-geometry}.} Set $\mu_s:=\E_s[v_s\mid z_{\star,s}]$ and $\mathcal G_s:=\sigma(\mathcal F_s,z_{\star,s})$. Given $\mathcal G_s$, $h_{i,s}=m_{i,s}^\top z_{\star,s}+h_{i,s}^\perp$ with $h_s^\perp\sim\mathcal N(0,\Sigma_s)$, as for $k_\star=1$, so the decoupling bound of Appendix~\ref{app:decoupling} (with $m_{i,t}z_{0,t}$ replaced by $m_{i,t}^\top z_{\star,t}$) and the martingale arguments of Appendix~\ref{app:complement-geometry-proof} carry over unchanged.
    \item \textbf{Lemma~\ref{lem:gram-onestep}.} Stein's identity is applied over $h_t^\perp$ given $(\mathcal F_t,z_{\star,t})$ (Appendix~\ref{app:gram-onestep-proof}); $J_t$, $K_t$ and the error bound are unchanged.
    \item \textbf{Finite-mode structure.} Appendix~\ref{app:teacher-finite-rank} is written for general $k_\star$. The truncation keeps the $N_D=\binom{k_\star+D}{D}$ Hermite modes of total degree at most $D$, rather than $D+1$, so a multi-index teacher can inject a richer set of coherent modes.
    \item \textbf{Theorem~\ref{thm:equivalence-spike}.} Steps~1--3 of Appendix~\ref{app:equivalence-proof} are written in terms of $z_{\star,t}$ and $M_t\in\R^{k\times k_\star}$, and the test loss still depends on the complement only through $\Sigma_t$. Lemma~\ref{lem:S} is proved for $M_t\in\R^{k\times k_\star}$ (the norm $\|\cdot\|_\star$ uses $\|M\|_F$, and $k_\star$ enters its constants only through moments of $\|z_\star\|_2$), and Step~4 is unchanged.
    \item \textbf{Proposition~\ref{prop:variance-inflation}} is stated for $z_\star\in\R^{k_\star}$ and needs no change; neither do the moment bounds of Appendix~\ref{app:stability}.
    \item \textbf{Proposition~\ref{prop:fast-alignment}} does not extend verbatim. For $k_\star>1$ the signal state is the matrix $M_t\in\R^{k\times k_\star}$ rather than a vector, and its coupling to $a_t$ depends on the direction-dependent Hermite coefficients of $g$; a multi-index alignment result requires a separate matrix-valued analysis.
\end{itemize} 

\subsection{Beyond smooth activations (e.g., ReLU)}

Assumption~\textbf{A3} ($\sigma\in W^{3,\infty}$) is dictated by the proof technique rather than by the underlying
phenomena. Boundedness of $\sigma''$ enters our analysis via three ingredients: (i) Stein's identity for $J_t$ in
Lemma~\ref{lem:gram-onestep} and Step~3 of Appendix~\ref{app:equivalence-proof}, and Lemma~\ref{lem:S} in its Step~4, which also uses bounded $\sigma'''$; (ii) the chain rule for $\sigma'(h_{i,t})$ in Lemma~\ref{lem:app-mu-sobolev} and the Lipschitz bound on $\Psi$ in Appendix~\ref{app:fast-alignment}; and (iii) the Gaussian-interpolation covariance bound of Lemma~\ref{lem:app-decouple-centered}, used for Lemma~\ref{lem:complement-geometry}. (Proposition~\ref{prop:fast-alignment} additionally requires an odd $\sigma$, \textbf{B1}, which ReLU violates.) In Figure~\ref{fig:equivalence} (ReLU student in panels (a)--(b) and in one curve of (c)), Decomposed Dynamics closely tracks SGD, suggesting that the smoothness
requirement is technical rather than essential.

Several standard routes can extend the present analysis to non-smooth activations such as ReLU:
\begin{enumerate}[topsep=0pt,itemsep=-0.5ex,partopsep=1ex,parsep=1ex,leftmargin=*]
    \item \textbf{Mollification.} Replace $\sigma$ by a smooth surrogate $\sigma^{(\delta)} := \sigma_M * \phi_\delta$
    for a smooth mollifier $\phi_\delta$ with bandwidth $\delta>0$, with $\sigma_M$ the truncation of item~2 for unbounded $\sigma$ (and $\sigma_M:=\sigma$ otherwise), so that $\sigma^{(\delta)}\in W^{3,\infty}$ and
    the present analysis applies for each $\delta>0$. Quantitative tracking of the resulting bounds as
    $\delta\to 0$ and, for unbounded $\sigma$, $M\to\infty$ (with $\delta=\delta(n)\to 0$ and $M=M(n)\to\infty$ slowly) recovers the result for $\sigma$ in the limit, provided the $\sigma$- and $\sigma^{(\delta)}$-trajectories
    can be compared over $t\le\tau n$. For Theorem~\ref{thm:equivalence-spike}, $\delta(n)$ must also decay slowly enough for the bound of Step~4 of Appendix~\ref{app:equivalence-proof} to vanish, since the constants of Lemma~\ref{lem:S} depend on $\|\sigma^{(\delta)}\|_{W^{3,\infty}}$, which for ReLU grows like $\delta^{-2}$.
    \item \textbf{Truncation and uniform integrability.} For unbounded $\sigma$ (e.g.,\ ReLU) replace it by
    $\sigma_M := \psi_M\circ\sigma$ at level $M\to\infty$, where $\psi_M\in C^\infty$ satisfies $\psi_M(u)=u$ for $|u|\le M$ and is constant on $(-\infty,-M-1]$ and on $[M+1,\infty)$, then propagate uniform moment bounds on the
    preactivations through the truncation error.
    \item \textbf{Distributional Stein's identity.} For Lipschitz $\sigma$ with weak derivative $\sigma'\in L^\infty$,
    the Gaussian integration-by-parts identity $\E[Z\,\phi(Z)]=\E[\phi'(Z)]$ holds in the distributional sense, and
    extends to integrals of $\sigma'(h)$ via Stein's lemma without requiring pointwise $\sigma''$. The chain-rule
    expansions in Lemmas~\ref{lem:gram-onestep} and~\ref{lem:app-mu-sobolev} can be replaced by their
    Sobolev/Stein--Malliavin counterparts \cite{Nourdin2012malliavin}, at the cost of additional bookkeeping. For ReLU, Lemma~\ref{lem:app-mu-sobolev} then needs a lower bound on the complement norms $r_{i,t}$, and Lemmas~\ref{lem:app-decouple-centered} and~\ref{lem:S} need separate arguments, since they use $L^2$ bounds on $\partial_iF_i$ and bounded $\sigma'''$, respectively. The bound on $\|J_t\|_2$ likewise requires a lower bound on $\min_i\|f_{i,t}\|_2$, and for unbounded $\sigma$ this route must be combined with item~2.
    \item \textbf{Hermite truncation.} Decompose $\sigma$ into its Hermite expansion $\sigma = \sum_{j\ge 0}
    \sigma_j H_j$ on $L^2(\gamma)$ and truncate at finite degree $N$. The truncated activation $\sigma^{(\le N)}$
    is a polynomial (hence smooth, but unbounded, so it must also be combined with item~2), and for Lipschitz $\sigma$ the truncation error vanishes as $N\to\infty$ at
    rate $\sigma - \sigma^{(\le N)} = O(1/\sqrt N)$ in $L^2(\gamma)$; since the preactivation variances $q_{i,t}:=\|f_{i,t}\|_2^2$ differ from $1$ and vary, the expansion must be taken in $L^2(\mathcal N(0,\bar q))$ with $\bar q\ge\max_{i,t}q_{i,t}$, which calls for uniform upper and lower bounds on $q_{i,t}$.
\end{enumerate}
Carrying out any of these routes for ReLU, with full quantitative tracking through every proof step, falls
outside the scope of the present paper. The agreement observed for ReLU
(Figure~\ref{fig:equivalence}) suggests that the equivalence of Theorem~\ref{thm:equivalence-spike} extends, and we leave a complete
extension as an open direction.

\end{document}